\documentclass[10pt,twocolumn,letterpaper]{article}

\usepackage{cvpr}              

\usepackage{fancyhdr}

\fancypagestyle{puffin}{
    \fancyhf{}
\fancyhead[R]{\raisebox{30pt}{\footnotesize \thepage}}

}

\makeatletter

\let\oldmaketitle\@maketitle

\renewcommand{\@maketitle}{%
    \oldmaketitle

    \vspace{-8mm}

    \begin{center}
        \includegraphics[width=\textwidth]{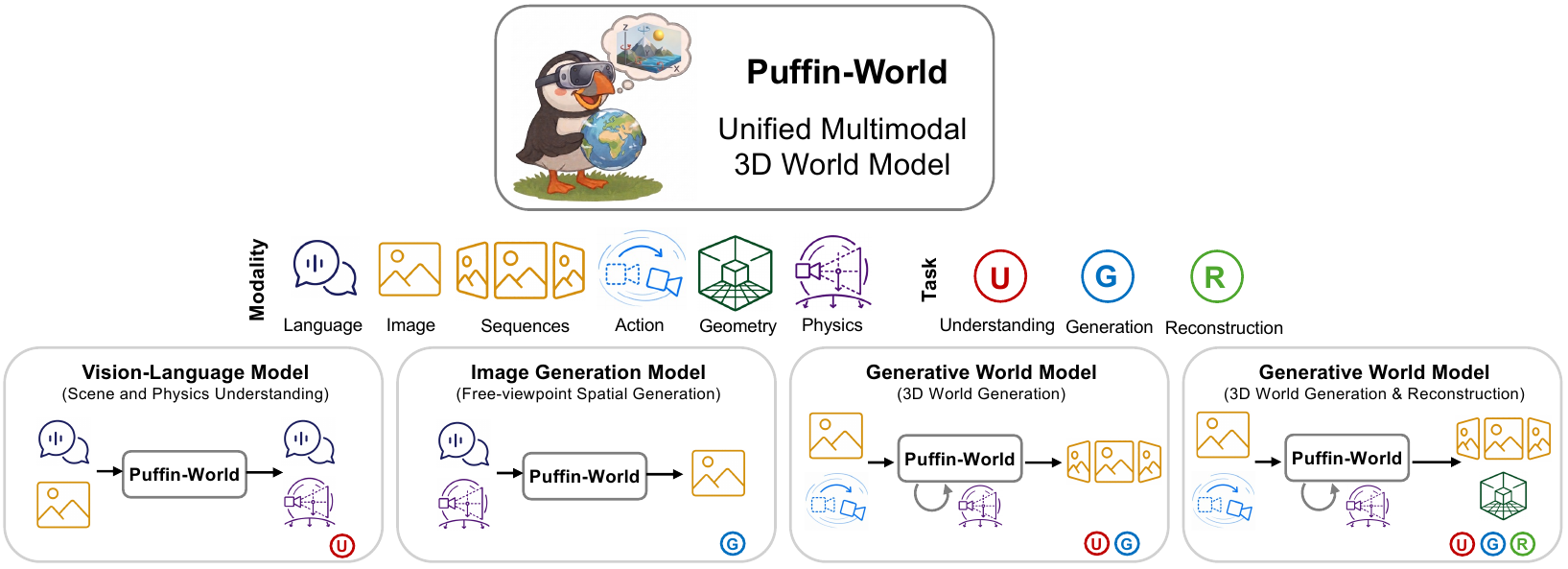}

        \vspace{1mm}

        \parbox{\textwidth}{
            \small
            \refstepcounter{figure}
            \textbf{Figure~\thefigure.}
            \textbf{Puffin-World is a unified multimodal model using native 3D world states
            (Appearance-Geometry-Physics) for spatial intelligence and physical AI.}
            It unifies camera physics understanding, free-viewpoint spatial simulation,
            and 3D world generation and reconstruction within a single framework.
            \label{fig:teaser_overall}
        }
    \end{center}

    \vspace{3mm}
}

\makeatother

\usepackage{amsmath}
\usepackage{amsfonts}
\usepackage{amssymb}
\usepackage{bbm}
\usepackage{bm}

\usepackage{array}
\usepackage{booktabs}
\usepackage{multirow}
\usepackage{tabularx}
\usepackage{makecell}
\usepackage{adjustbox}
\usepackage{colortbl}

\usepackage{graphicx}
\usepackage{subcaption}
\usepackage{wrapfig}
\usepackage{rotating}

\usepackage{algorithm}
\usepackage{algorithmic}

\usepackage{pifont}
\usepackage{gensymb}
\usepackage{siunitx}

\usepackage{nicefrac}
\usepackage{microtype}
\usepackage[normalem]{ulem}
\usepackage{enumitem}
\usepackage{etoc}

\usepackage{cuted}

\usepackage{svg}

\definecolor{crimson}{rgb}{0.86,0.08,0.24}
\definecolor{lightblue}{rgb}{0.90,0.95,1.00}
\definecolor{lightred}{rgb}{1.00,0.90,0.95}
\definecolor{darkred}{RGB}{159,0,0}

\definecolor{tabfirst}{rgb}{0.68,0.85,0.68}
\definecolor{tabsecond}{rgb}{0.86,0.95,0.86}
\definecolor{tabthird}{rgb}{1.00,1.00,1.00}

\newcommand{\cthird}{}
\newcommand{\cfirst}{\cellcolor{tabfirst}\bfseries}
\newcommand{\csecond}{\cellcolor{tabsecond}}

\newcommand{\greencheck}{\textcolor{green}{\ding{52}}}
\newcommand{\redcheck}{\textcolor{red}{\ding{55}}}

\newcommand{\0}{\phantom{0}}

\renewcommand{\paragraph}[1]{%
    \vskip4pt
    \noindent
    \textbf{#1}%
}

\makeatletter
\renewcommand\@fnsymbol[1]{}
\makeatother

\definecolor{cvprblue}{rgb}{0.21,0.49,0.74}

\definecolor{urlpink}{RGB}{219, 70, 140}

\usepackage[
    pagebackref,
    breaklinks,
    colorlinks,
    linkcolor=cvprblue,
    citecolor=cvprblue,
    urlcolor=urlpink
]{hyperref}

\title{
Puffin-World: Scaling a Unified Multimodal  Model with \\ Native 3D World States
}

\author{
Kang Liao\textsuperscript{1}\qquad
Yihang Luo\textsuperscript{1}\qquad
Xiao-Ming Wu\textsuperscript{1}\qquad
Linyi Jin\textsuperscript{2}\qquad
Size Wu\textsuperscript{1}
\\
Chunyu Lin\textsuperscript{3}\qquad
Yao Zhao\textsuperscript{3}\qquad
Fei Wang\textsuperscript{4}\qquad
Wei Li\textsuperscript{1}\qquad
Chen Change Loy\textsuperscript{1,4}
\\[2pt]
\textsuperscript{1}S-Lab, Nanyang Technological University
\qquad
\textsuperscript{2}University of Michigan
\\
\textsuperscript{3}Beijing Jiaotong University
\qquad
\textsuperscript{4}ACE Robotics
\\[3pt]
\url{https://kangliao929.github.io/projects/puffin-world/}
}

\begin{document}
\pagestyle{puffin}

\maketitle
\thispagestyle{puffin}

\markboth{Puffin-World}%
{}

\begin{abstract}
We propose Puffin-World, a unified multimodal architecture that integrates physical understanding, spatial simulation, and 3D world generation and reconstruction without relying on external offline modules. To reliably construct and interact with 3D worlds, our framework jointly models three native world states: physics (gravity field and latitude), geometry (depth), and appearance (image), together with a unified Omni-Camera representation that supports diverse tasks and flexible motions. Beyond modeling these states, we introduce a strategy for propagating physical dynamics across future frames. By grounding absolute camera properties in the real world, Puffin-World enables physically consistent and visually stable world generation. We further couple appearance and geometry within a single generative process, jointly synthesizing each future view and reconstructing its underlying geometry. This unified paradigm enables interleaved closed-loop applications requiring synergy across multiple tasks, including mimic and self-calibrated world exploration. To scale Puffin-World to complex scenarios, we construct Puffin-16M, comprising 15 million vision–language–camera triplets and 1 million trajectories featuring various and challenging motions. To foster further research in this area, we released the code, models, and datasets.
\end{abstract}

\begin{figure*}[!htbp]
    \centering

    \includegraphics[width=\linewidth]{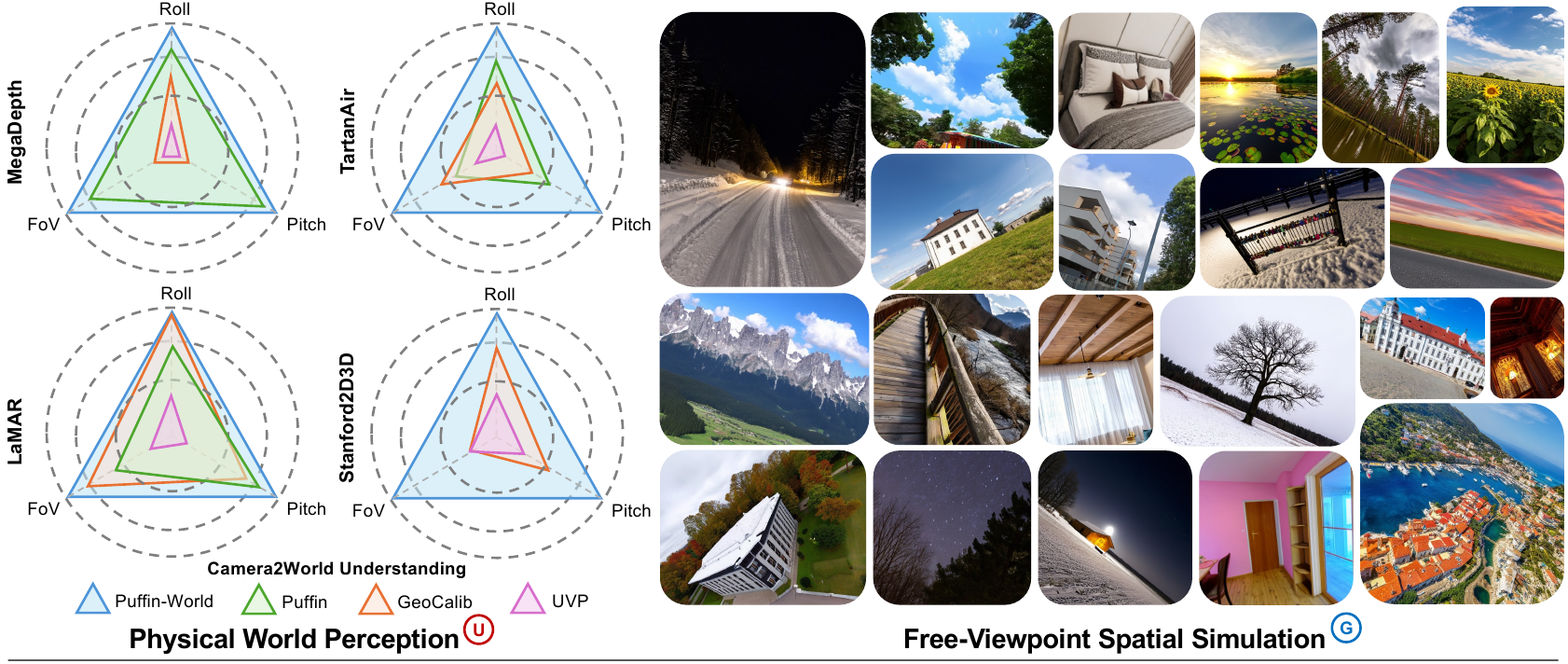}


    \includegraphics[width=\linewidth]{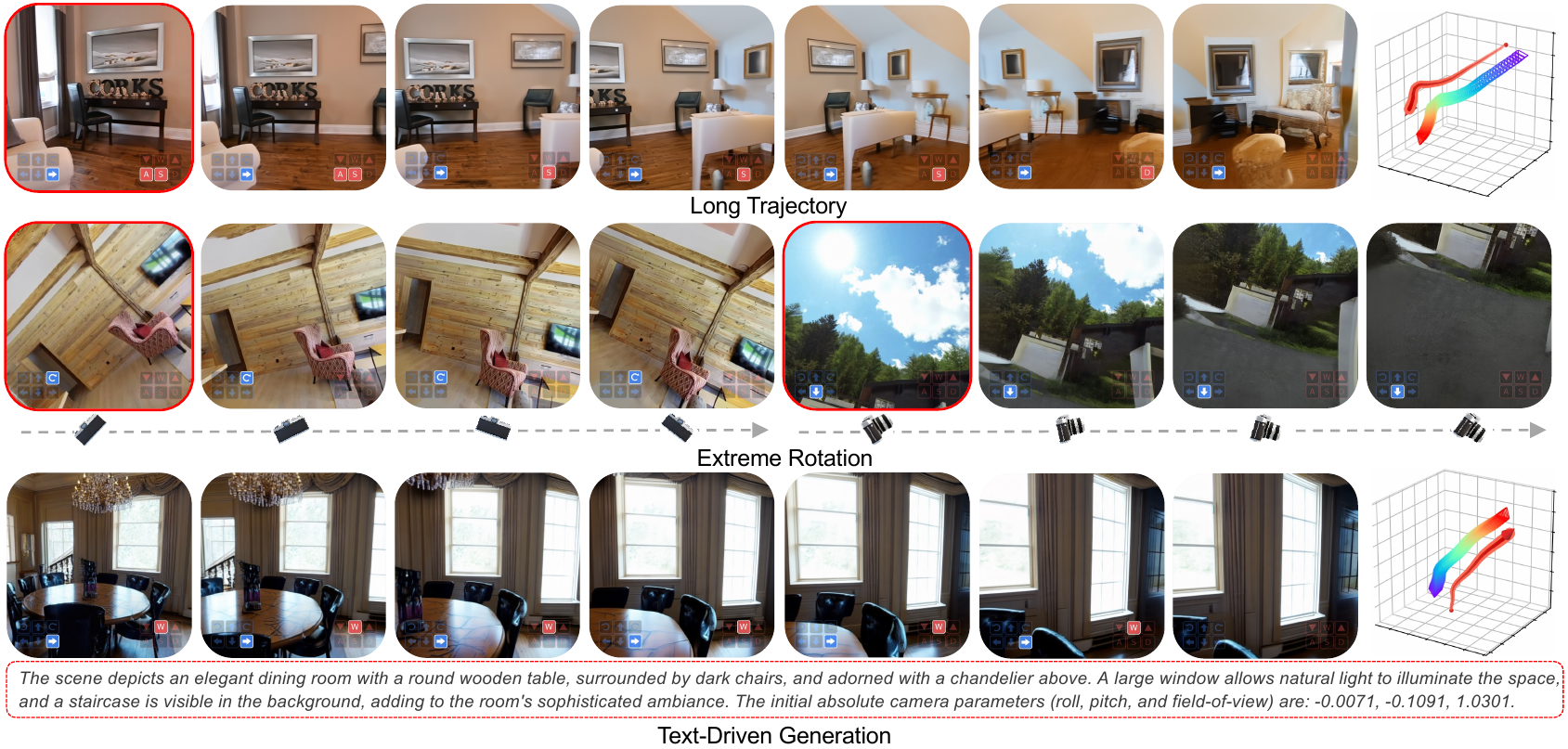}


    \includegraphics[width=\linewidth]{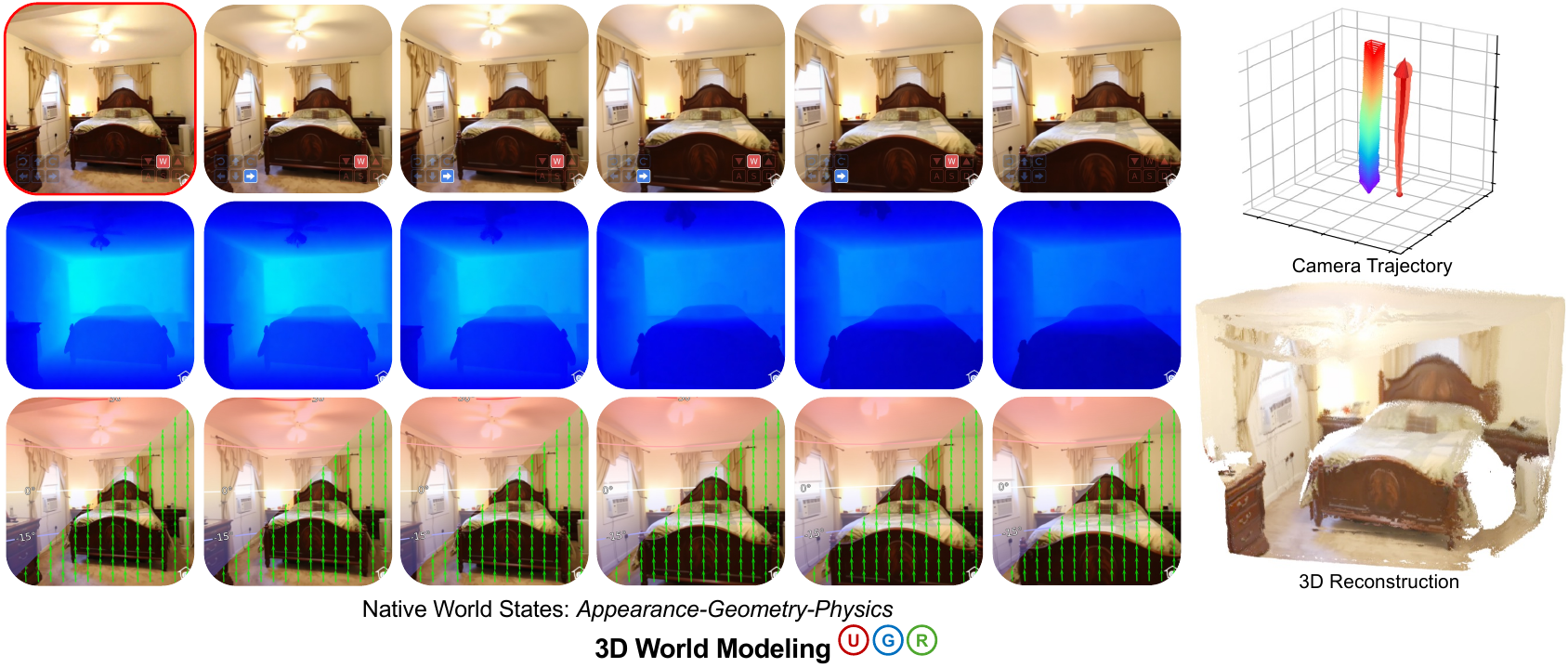}

    \caption{
    \textbf{Illustration of the versatile capabilities of our model.}
    It shows the best performance (AUC\,$\triangleright$\,5\degree) on camera-to-world understanding for physical world perception; flexible camera-controllable abilities for free-viewpoint spatial simulation; 3D world generation and reconstruction with flexible-DoF actions and native world states. The red box marks the input to the model.
    }
    \label{fig:teaser}
\end{figure*}

\section{Introduction}

Building a model that can perceive, generate, and reconstruct the world from arbitrary visual observations is a central goal of multimodal spatial intelligence and physical AI. A capable world model should not merely generate plausible pixels; it should understand where the camera or viewpoint is in the real world, reason about the geometry of the scene, and synthesize what the world looks like from new viewpoints in a way that stays consistent with physical law. The physical world therefore cannot be adequately represented as a collection of 2D images. Instead, effective interaction with the real world calls for a unified framework that jointly models its multiple interconnected states.

Progress so far has advanced along separate fronts that each capture only part of this picture. Generative world models~\cite{agarwal2025cosmos, genie3, he2024cameractrl, wang2024motionctrl, gao2024cat3d, zhu2026astra, huang2025vid2world} have made striking progress, but they predict the world almost exclusively at the appearance level, treating each frame as RGB content with no explicit notion of the camera's physical orientation or the underlying scene geometry. In parallel, unified multimodal models~\cite{team2024chameleon, wang2024emu3, zhou2024transfusion, wu2025janus, chen2025janus} couple understanding and generation within a single network, but do so for only 2D semantics. As a result, the field still lacks a single model that unifies holistic modalities and tasks for 3D world modeling.

Unifying these capabilities is fundamentally non-trivial. A simple combination of existing perception, generation, and reconstruction components is insufficient because a unified world model faces three tightly coupled challenges: (i) establishing a unified action representation that supports both continuous single- and cross-view control while remaining grounded in absolute physical concepts such as gravity, uprightness, and orientation; (ii) maintaining a physically persistent frame that allows knowledge inferred from observed views to consistently guide 3D world modeling at unseen viewpoints; and (iii) scaling these capabilities with data that provides both absolute camera grounding and diverse, challenging motion. Existing relative camera representations, such as Pl\"ucker embeddings~\cite{sitzmann2021light, he2024cameractrl, gao2024cat3d}, offer effective control but lack a global physical anchor, while prevailing 3D datasets~\cite{ling2024dl3dv, RealEstate10K, dai2017scannet} contain limited rotational diversity and rarely provide absolute camera orientation.

In this work, we present \textbf{Puffin-World}, a unified multimodal model that scales 3D world perception, generation, and reconstruction as shown in Figure~\ref{fig:teaser_overall}, without relying on any external offline modules. In particular, we formulate three native world states to jointly model the 3D world in an end-to-end manner: the \emph{physics} that grounds an observation in the absolute physical world, the \emph{geometry} that describes underlying 3D spatial structure, and the \emph{appearance} that we finally see from the visual content manifested in images and sequences.
To address the aforementioned challenges, we first introduce the Omni-Camera representation, a dense action signal that combines a gravity-anchored absolute field with a relative ray field, which enables physics-grounded text-to-image spatial simulation (single-view) and image-to-3D scene generation (cross-view). Subsequently, we propose physics propagation: by anchoring the absolute spatial knowledge derived from world perception and propagating it across the relative control signals of future frames, Puffin-World generates appearance that remains stable and gravity-consistent over complex motions. Furthermore, we scale the whole framework with the constructed \textbf{Puffin-16M}, a new dataset of $15$M vision--language--camera triplets and $1$M trajectories featuring diverse and challenging camera motions with precise labels.

Thanks to this unified paradigm, Puffin-World supports diverse multimodal tasks as illustrated in Fig.~\ref{fig:teaser}. For physical world perception, it attains state-of-the-art camera-to-world understanding, achieving the best median error and the best AUC at $5^\circ$ across four public benchmarks. For free-viewpoint spatial simulation, it generates camera-controllable images with more faithful distribution than strong general-purpose generators~\cite{GPT-Image2, NanoBanana2, Qwen-Image-2, flux-2-2025}. For 3D world modeling, it performs high-DoF action-conditioned, text/image-to-3D generation while jointly reconstructing per-view geometry, enabling consistent and physically grounded world modeling. Furthermore, Puffin-World enables complicated, closed-loop applications that require multi-task synergy, such as mimic and self-calibrated world exploration. We summarize our contributions as follows:

\begin{itemize}[noitemsep,topsep=2pt,leftmargin=1.2em]
    \item We propose Puffin-World, a unified multimodal model that formulates native 3D  world states and jointly perceives, generates, and reconstructs the 3D world within a single framework.
    \item We introduce the Omni-Camera representation, a unified dense camera condition that integrates gravity-aware absolute orientation with ray-based relative geometry. A physics propagation mechanism is further proposed to anchor absolute spatial knowledge from perception and propagate it across future control signals.
    \item We construct \emph{Puffin-16M}, comprising Puffin-Cam-15M ($15$M vision--language--camera triplets) and Puffin-Traj-1M ($1$M challenging-motion trajectories). We release it at \url{https://kangliao929.github.io/projects/puffin-16m}.
    \item To promote the development of the research community, we have fully open-sourced our code, models, and datasets. Moreover, leveraging Puffin-World’s accurate understanding of absolute camera physics, we annotated 28 widely used public datasets. These annotations cover approximately 44.5 million images across diverse data distributions.
\end{itemize}
\section{Related Work}
\label{sec:related_work}

\noindent\textbf{Camera-to-World Understanding.}
Recovering physical camera parameters from images, including camera calibration and pose estimation, is a long-standing problem in 3D vision~\citep{pollefeys1999self, hartley2003multiple, liao2023deep, veicht2024geocalib, jin2023perspective, lin2025towards}. Early learning-based methods directly regress camera parameters from a single image~\citep{hold2018perceptual, workman2015deepfocal, bogdan2018deepcalib, zhai2016detecting, kendall2015posenet}, whereas more recent approaches narrow the prediction gap by incorporating intermediate geometric structures or semantic cues~\citep{lee2020neural, lee2021ctrl, Song2024MSCC, janampa2024sofi, yin2018fisheyerecnet}. A particularly successful direction learns dense, pixel-wise geometric representations, including distortion maps~\citep{liao2020model, liao2021deep}, pixel displacement fields~\citep{li2019blind, liao2025mowa, xie2025aligndiff}, camera rays~\citep{zhang2024cameras}, perspective fields~\citep{jin2023perspective, veicht2024geocalib, tirado2025anycalib}, and incidence fields~\citep{zhu2023tame, he2025diffcalib, deng2024boost}. These representations provide spatially grounded supervision and generally offer greater robustness than direct global regression. More recently, Puffin~\citep{puffin} reframes this task as a language-modeling problem by representing the camera as text and predicting its parameters through spatial reasoning. However, its limited model scale and training data constrain its performance and scalability, causing it to remain inferior to specialized vision-based methods such as GeoCalib~\citep{veicht2024geocalib} on several benchmarks~\cite{sarlin2022lamar}.

\noindent\textbf{Unified Multimodal Models.}
Building upon conventional large multimodal models, unified multimodal models~\citep{team2024chameleon, wang2024emu3, tang2025ugen, wu2025harmonizing, lin2025toklip, wu2024vila, tong2024metamorph, zhou2024transfusion, deng2025emerging, zhang2025unified, chen2025janus, song2026joyai, agarwal2026cosmos, su2026generation, han2026vision} integrate visual understanding and generation within a single network, either through autoregressive modeling over discrete~\citep{team2024chameleon, wu2024vila, wang2024emu3, wu2025janus} or continuous~\citep{fan2025unified, qu2025tokenflow} visual tokens, or by connecting pretrained multimodal models with diffusion decoders~\citep{pan2025transfer, chen2025blip3, wu2025openuni, lin2025uniworld, ma2025janusflow, huang2025illume+, zhang2025nexus, xie2024show, xie2025show}. Despite strong progress in general image understanding and generation, these models largely treat images as 2D appearance and semantics under simplified camera assumptions, without explicitly modeling camera physics, scene geometry, or other essential states. Puffin~\citep{puffin} takes an initial step toward camera-centric unification, but still focuses on isolated-view perception and lacks a persistent world frame across views. This motivates extending multimodal unification from 2D semantics to 3D world states.

\noindent\textbf{3D World Models.}
Generative world models~\cite{lee2026unified} aim to simulate the world by predicting future observations, with recent video- and multi-view diffusion frameworks achieving impressive visual fidelity~\citep{ha2018world, genie3, agarwal2025cosmos, assran2025v, bar2025navigation, yuan2025generative, hansen2023td, zhu2026astra, huang2025vid2world, wu2026geometry, dai2026fantasyworld, zhang2026worldstereo, zheng2026versecrafter}. To enable controllable simulation, extensive research conditions generation on camera information: dense camera-pose or Pl\"ucker-ray embeddings guide camera-controlled video generation~\citep{he2024cameractrl, wang2024motionctrl, bahmani2024vd3d, xu2024camco, yang2024direct}, while multi-view diffusion models synthesize novel views or complete scenes from one or a few reference images~\citep{gao2024cat3d, cao2024mvgenmaster, ren2025gen3c, bernal2025precisecam}, sometimes coupled with feed-forward 3D reconstruction models to recover geometry~\citep{wang2024dust3r, wang2025vggt, wang2025cut3r, stream3r2025}. Despite this progress, existing 3D world models exhibit three major limitations. First, they predominantly model the world at the \emph{appearance} level, leaving geometry and physical grounding implicit. Second, they rely primarily on \emph{relative} camera motion, which lacks a global physical reference frame. Consequently, the same relative trajectory may correspond to different absolute world orientations, leading to orientation drift and degraded performance under challenging motions, long horizons, and single-view settings. Third, generation is typically treated as an isolated task, decoupled from physical camera understanding and geometry reconstruction, and is often trained on data with limited rotational diversity. These limitations motivate a unified world model that jointly represents physics, geometry, and appearance, anchors generation to an absolute physical frame, and integrates perception, simulation, and reconstruction within a single framework.

\section{Method}
\label{sec:method}

\subsection{Preliminary}

\subsubsection{3D Native World States}

Most existing generative world models focus primarily on appearance-level prediction, where future states are represented as RGB images or videos. However, the physical world is natively organized by multiple levels of states, including physics, geometry, and appearance. Physics-level states provide global physical grounding, geometry-level states describe the 3D spatial structure, and appearance-level states represent the observable visual content. Jointly modeling these complementary states is crucial for stable generation, spatially consistent simulation, and grounded world interaction.

In Puffin-World, we formulate a hierarchy of native 3D world states. Specifically, the \emph{physics} state captures absolute physical cues, including the gravity field and latitude map; the \emph{geometry} state represents scene depth; and the \emph{appearance} state corresponds to RGB observations. Since some states, particularly physics-level cues, are difficult to infer directly from visual observations, and their interactions across different levels remain underexplored, Puffin-World jointly perceives, propagates, and generates these states within a unified framework. This design allows physics-aware signals to guide future-view generation toward geometrically consistent structure and realistic appearance.

\subsubsection{Unifying Camera Representations}
The camera serves as a fundamental interface between the 3D world and 2D visual data modalities. 
In visual perception and reconstruction, camera models explicitly encode the physical rules of perspective projection, thereby supporting tasks such as single-view camera calibration, multi-view geometric reasoning, and joint spatial reconstruction. Beyond perception, cameras also provide effective and practical control signals for generative models, enabling spatially controllable synthesis, continuous world generation, and consistent physical simulation.

\begin{figure*}[!htbp]
    \centering

    \includegraphics[width=\linewidth]{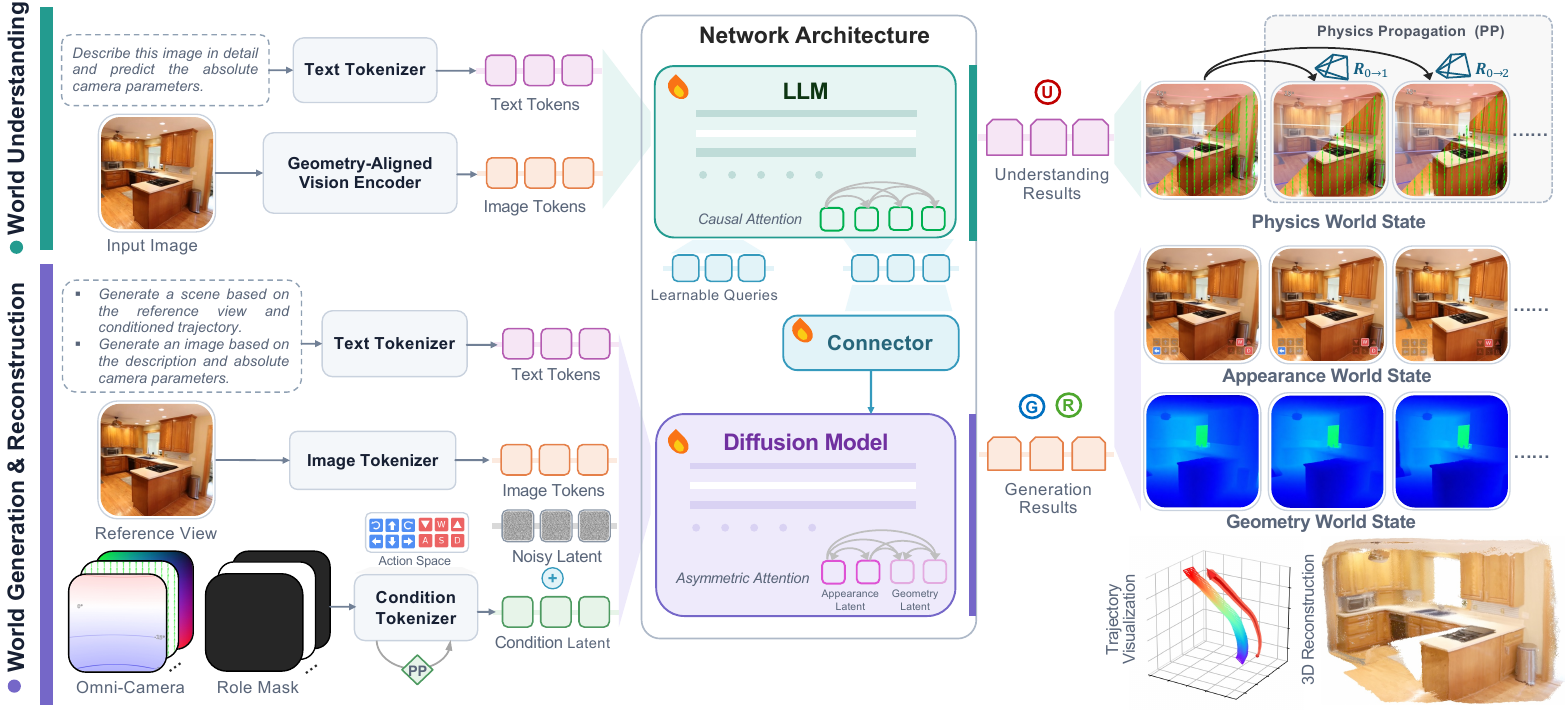}

    \caption{
    \textbf{The network architecture of our Puffin-World.} It mainly comprises 3D world understanding, generation, and reconstruction, formulating native 3D world states within one framework. Both Omni-Camera representation and role mask are conditioned on the reference view and the target views. For clarity, we omit the camera-controllable text-to-image generation and the mathematical mapping from the understanding results to the physics world state.
    }
    \label{fig:framework}
\end{figure*}

Existing camera representations can be broadly categorized as relative or absolute. \textbf{Relative camera representations} capture spatial relationships across viewpoints and are widely used for multi-view reconstruction and continuous scene generation. Although readily obtained from structure-from-motion and effective for cross-view geometry, they lack a global physical anchor and cannot encode absolute orientation with respect to the real world. \textbf{Absolute camera representations}, in contrast, ground individual observations to physical cues such as the horizon and scene uprightness, but are less suited to continuous multi-view transitions and are substantially harder to obtain from monocular images. These complementary properties motivate a unified camera representation that combines absolute physical grounding with continuous spatial modeling. To this end, we propose the Omni-Camera representation, a simple but effective unified camera representation for versatile applications. For each pixel $\mathbf{x}=(u,v)$, Omni-Camera representation combines an absolute camera field and a relative ray field along the channel dimension:
\begin{equation}
\begin{aligned}
\mathbf{c}_{\mathbf{x}}
&=
\mathrm{Concat}\left(\mathbf{a}_{\mathbf{x}}, \mathbf{p}_{\mathbf{x}}\right)
\in \mathbb{R}^{9},
\\
\mathbf{a}_{\mathbf{x}}
&=
\left(
\mathbf{u}_{\mathbf{x}},
\varphi_{\mathbf{x}}
\right)
\in \mathbb{R}^{3},
\quad
\mathbf{p}_{\mathbf{x}}
\in \mathbb{R}^{6}.
\end{aligned}
\label{eq:omni_camera_representation}
\end{equation}
Here, $\mathbf{a}_{\mathbf{x}}$ denotes the absolute camera representation consisting of the pixel-wise up-vector $\mathbf{u}_{\mathbf{x}}$ and latitude angle $\varphi_{\mathbf{x}}$, while $\mathbf{p}_{\mathbf{x}}$ denotes the relative camera representation given by the ray map composed of the ray origin and direction.
Let $\tilde{\mathbf{x}}=[u,v,1]^\top$ be the homogeneous coordinate of pixel $\mathbf{x}$. Given the camera intrinsic matrix $\mathbf{K}$ and the world-to-camera extrinsic parameters $(\mathbf{R},\mathbf{t})$, where $\mathbf{X}_c=\mathbf{R}\mathbf{X}_w+\mathbf{t}$, the camera center in the world coordinate system is $\mathbf{o}=-\mathbf{R}^{\top}\mathbf{t}$. The viewing ray direction from the camera center to pixel $\mathbf{x}$ is
\begin{equation}
\mathbf{d}_{\mathbf{x}}
=
\frac{
\mathbf{R}^{\top}\mathbf{K}^{-1}\tilde{\mathbf{x}}
}{
\left\|\mathbf{R}^{\top}\mathbf{K}^{-1}\tilde{\mathbf{x}}\right\|_2
}.
\label{eq:ray_direction}
\end{equation}
The relative ray representation~\citep{gao2024cat3d} of this ray is then formed by concatenating the ray origin and the ray direction,
\begin{equation}
\mathbf{p}_{\mathbf{x}}
=
\left(
\mathbf{o},
\mathbf{d}_{\mathbf{x}}
\right)
\in \mathbb{R}^{6},
\label{eq:ray_map}
\end{equation}
where $\mathbf{o}=-\mathbf{R}^{\top}\mathbf{t}$ is the ray origin shared by all pixels of the view and $\mathbf{d}_{\mathbf{x}}$ is the per-pixel ray direction. Relative camera motion between two views is thereby encoded by the change of the ray origin (translation) and the ray direction (rotation).

For the absolute component, we follow the Perspective Field~\citep{jin2023perspective} to represent the physical orientation of each pixel. Let $\mathbf{g}\in\mathbb{R}^{3}$ be the unit gravity direction and $\Pi(\cdot)$ be the projection function. For any 3D point $\mathbf{X}$ on the ray of pixel $\mathbf{x}$, satisfying $\Pi(\mathbf{X})=\mathbf{x}$, the up-vector and latitude angle are defined as
\begin{equation}
\begin{aligned}
\mathbf{u}_{\mathbf{x}}
&=
\lim_{\epsilon \rightarrow 0}
\frac{
\Pi(\mathbf{X}-\epsilon\mathbf{g})-\Pi(\mathbf{X})
}{
\left\|\Pi(\mathbf{X}-\epsilon\mathbf{g})-\Pi(\mathbf{X})\right\|_2
},
\\
\varphi_{\mathbf{x}}
&=
\arcsin
\left(
\frac{
(-\mathbf{d}_{\mathbf{x}})^{\top}\mathbf{g}
}{
\left\|\mathbf{d}_{\mathbf{x}}\right\|_2
}
\right).
\end{aligned}
\label{eq:perspective_field}
\end{equation}
Here, $-\mathbf{d}_{\mathbf{x}}$ corresponds to the incoming light ray direction. The up-vector $\mathbf{u}_{\mathbf{x}}$ describes the image-plane projection of the direction opposite to gravity, and the latitude angle $\varphi_{\mathbf{x}}$ measures the elevation of the incoming ray with respect to the horizontal plane.

By integrating gravity-aware absolute orientation with ray-based relative geometry, the Omni-Camera representation provides both global physical grounding and continuous spatial modeling. Although conceptually simple, it effectively serves as a unified and flexible camera condition for physical-world understanding, camera-controllable image generation, and continuous world generation, supporting both single-view and cross-view world modeling within a common representation.

\subsection{Puffin-World}
Puffin-World unifies physical-world perception, free-viewpoint spatial simulation, and 3D world generation and reconstruction within a single multimodal framework. Its versatility arises from three levels of unification rather than task-specific subnetworks. \emph{(i) Representation unification.} Appearance and geometry share a common latent space, with RGB images and depth maps encoded by the same VAE, while all camera configurations and motions, ranging from in-place rotation to large translation, are represented by the unified Omni-Camera representation $\mathbf{C}$ defined in Eq.~\ref{eq:omni_camera_representation}. A compact four-channel role mask $\mathbf{m}\in\mathbb{R}^{4\times H\times W}$ is paired with $\mathbf{C}$ to indicate whether each view serves as a generation target, a conditioning reference, an image-conditioned input, or a geometry view, and both are processed by the same condition fusion module. \emph{(ii) Modality unification.} A single backbone both perceives and generates the world: the geometry-aligned vision encoder and LLM yield autoregressive understanding outputs, while the same LLM hidden states are transformed by learnable queries and a lightweight connector into conditioning signals for the diffusion generator, enabling perception and generation as two complementary outputs of one model. \emph{(iii) Task unification.} In Puffin-World, each task is determined by the available inputs, including text, target cameras, reference views, together with the role mask $\mathbf{m}$. By varying this input composition, the same framework supports camera-to-world understanding, camera-controllable text-to-image generation, 3D world generation, and joint appearance-geometry reconstruction. The overview of Puffin-World's framework is illustrated in Figure~\ref{fig:framework}. We detail these capabilities below.

\subsubsection{Physics Perception}
\label{sec:physics_perception}
Given a single image, Puffin-World estimates the camera's absolute physical state, including its gravity-relative orientation specified by roll and pitch, its intrinsic vertical field-of-view (FoV), and a semantic scene description. Following Puffin~\citep{puffin}, we formulate this task as autoregressive multimodal sequence modeling rather than direct regression. The geometry-aligned vision encoder~\cite{puffin} extracts visual features from images, which are projected into the LLM embedding space through an MLP projector. The LLM then generates a structured scene and spatial analysis, followed by the numerical camera parameters, and is trained using a next-token cross-entropy loss.

This formulation offers two key advantages for physical-state perception. First, it leverages the LLM's sequence modeling capability and favorable scaling behavior, recasting fine-grained camera understanding as language modeling. Second, because the camera parameters are predicted after reasoning about scene-level cues such as the horizon, vertical structures, and foreground composition, the estimation relies on holistic scene understanding rather than low-level visual cues alone. This design enables robust gravity-aware perception across diverse scenes.

\subsubsection{Free-Viewpoint Spatial Simulation}
\label{sec:free_viewpoint}
Puffin-World enables camera-controllable image generation: given a text prompt and a desired Omni-Camera map $\mathbf{C}$, it synthesizes an image whose realized viewpoint and intrinsics strictly adhere to $\mathbf{C}$. Because this part focuses on single-view generation, the ray map within the Omni-Camera representation is held constant. The resulting query hidden states pass through a lightweight connector, yielding a pooled conditioning vector and a sequence of joint-attention conditioning embeddings for the multimodal diffusion transformer (MMDiT)~\citep{esser2024scaling}. The MMDiT then denoises a target latent under a flow-matching objective, after which a VAE decoder maps the denoised latent back into pixel space.

Our approach to integrating camera condition is the primary departure from Puffin~\citep{puffin}. Puffin encodes a 3-channel Perspective Field into a continuous latent via a VAE and injects it into the image latent via cross-attention, limiting its extension to broader tasks and camera motions. Moreover, because a holistic camera representation cannot be formulated with only 3 channels, it cannot be straightforwardly encoded and injected in this manner. Instead, we fuse the Omni-Camera representation directly within the diffusion latent space. A lightweight condition fusion module $\mathcal{F}$ maps $\mathbf{C}$ and the role mask $\mathbf{m}$ into the input latent space of MMDiT. This output is added to the noisy image latent $\mathbf{z}$ prior to the patch embedding operation $\mathcal{P}$:
\begin{equation}
\mathbf{h}^{0} = \mathcal{P}\!\big(\mathbf{z} + \mathcal{F}([\mathbf{C};\mathbf{m}])\big).
\label{eq:cam_inject}
\end{equation}
To preserve the pretrained generator's behavior at the start of training, $\mathcal{F}$ is initialized such that its contribution starts near zero. Furthermore, to ensure the camera signal remains effective across deeper layers, we re-inject the patchified camera features at a sparse set of transformer blocks $\mathcal{L}$:
\begin{equation}
\mathbf{h}^{l} \leftarrow \mathbf{h}^{l} + \mathbf{W}_{l}\,\mathcal{P}\!\big(\mathcal{F}([\mathbf{C};\mathbf{m}])\big),\quad l\in\mathcal{L}.
\label{eq:cam_reinject}
\end{equation}
Compared to using a channel-limited mechanism over a separately VAE-encoded camera branch, this pixel-aligned additive injection supports versatile tasks. It grounds every latent token in its precise camera geometry across multiple layers, providing superior and stable spatial control while introducing negligible parameter overhead.

\subsubsection{3D World Modeling}
\label{sec:world_modeling}
Puffin-World then extends the single-view generator to multi-view trajectories while jointly supporting geometry reconstruction. Instead of denoising a single target image, the model processes a set of $T$ views within a unified joint-attention sequence. Among them, $K$ reference views are provided as clean latents $\{\mathbf{z}^{\mathrm{ref}}_{v}\}$ and are also encoded by the geometry-aligned vision encoder and the LLM to provide semantic conditioning. The remaining $T-K$ target views, represented by $\{\mathbf{z}^{\mathrm{tgt}}_{v}\}$, are corrupted with noise and jointly denoised under the flow-matching objective. Each view, whether reference or target, is associated with its own Omni-Camera condition, which is fused into the corresponding image latent through the additive injection mechanism in Eq.~\ref{eq:cam_inject} and Eq.~\ref{eq:cam_reinject}. To explicitly encode view identity within the shared attention sequence, we assign each token a \emph{view-axis} index and apply a one-dimensional rotary positional embedding along this axis. This allows the relative organization of views to be modeled directly within a single attention operation, enabling cross-view consistency to emerge without requiring an explicit inter-view consistency loss.

\noindent\textbf{Trajectory Anchoring via Physics Propagation.}
A trajectory introduces a challenge specific to absolute grounding: relative motion across views can be readily obtained from camera poses, whereas the absolute orientation that anchors the trajectory to the real world is generally unavailable. The Omni-Camera representation addresses this through its complementary components: the relative ray map captures inter-view motion, while the absolute Perspective Field encodes gravity-aligned orientation. We recover the latter for future frames through \emph{physics propagation}. Specifically, the model first perceives the absolute state $(\phi_0,\theta_0,\alpha_0,\dots)$ of the reference view through the Physics Perception pathway (Sec.~\ref{sec:physics_perception}), where $\phi_0$, $\theta_0$, and $\alpha_0$ denote its roll, pitch, and vertical FoV, respectively, yielding the gravity direction $\mathbf{g}_0$ in the reference frame. Given the relative rotation $\mathbf{R}^{\mathrm{rel}}_{t\leftarrow0}$ that transforms vectors from the reference camera coordinate frame to the camera coordinate frame of view $t$, the gravity direction in each view is propagated as:

\begin{equation}
\mathcal{G}
=
\left\{
\mathbf{g}_t
\right\}_{t=0}^{T-1}
=
\left\{
\mathbf{R}^{\mathrm{rel}}_{t\leftarrow0}\mathbf{g}_0
\right\}_{t=0}^{T-1}.
\label{eq:phys_prop}
\end{equation}
The propagated gravity direction of each view is then used to render its corresponding absolute camera representation following Eq.~\ref{eq:perspective_field}, which is concatenated with the relative ray map to form the complete Omni-Camera condition. In this way, all views along the trajectory share a coherent gravity-anchored world frame while retaining their individual relative motions, enabling physically and spatially stable generation under diverse and challenging camera trajectories.

\noindent\textbf{Joint Appearance--Geometry Modeling.}
Puffin-World predicts not only the appearance but also the geometry of each generated view, casting reconstruction as part of the same generative process. 
In particular, we encode depth as an RGB image using an invertible color mapping inspired by the 3D Hilbert curve in Vision Banana~\citep{visionbanana}, and encode it with the same frozen VAE used for appearance. Instead of directly regressing unbounded depth or disparity, this gives us a bounded three-channel representation that fits the pretrained VAE without changing its input or output interface. The mapping is deterministic and invertible, so the generated result can be decoded back to depth. 
Its nonlinear transform also allocates more of the color range to nearby surfaces, where geometric accuracy matters most.
This allows appearance and geometry to use the same latent representation without requiring a dedicated depth encoder or decoder.
For each modeled view, the depth latent is appended as an additional token block that shares the Omni-Camera condition and view-axis index of its RGB counterpart, distinguished only by the geometry channel of the role mask $\mathbf{m}$. A single denoising pass therefore jointly yields the appearance and geometry latents of all target views under the shared flow-matching objective:
\begin{equation}
\mathcal{L}_{\mathrm{fm}}(\mathcal{V}) = \mathbb{E}_{t,\,\bm{\epsilon}}\!\left[\frac{1}{|\mathcal{V}|}\sum_{v\in\mathcal{V}}\big\| \mathbf{v}_{\theta}(\mathbf{z}^{v}_{t}, t, \mathbf{c}^{v}) - (\bm{\epsilon}^{v}-\mathbf{z}^{v}_{0}) \big\|_{2}^{2}\right],
\label{eq:flow_loss}
\end{equation}
where $\mathbf{z}^{v}_{t}=(1-\sigma_{t})\mathbf{z}^{v}_{0}+\sigma_{t}\bm{\epsilon}^{v}$, $\mathbf{c}^{v}$ collects the per-view camera and connector conditioning, and $\mathcal{V}$ is a set of target views. Because the saturated colors of the geometry latents could otherwise corrupt the converged appearance model, we do not simply pool appearance and geometry targets into one objective. Instead, we evaluate Eq.~\ref{eq:flow_loss} separately over the appearance target views $\mathcal{V}_{\mathrm{rgb}}$ and the geometry target views $\mathcal{V}_{\mathrm{geo}}$, and combine them with a time-dependent weight:
\begin{equation}
\begin{aligned}
\mathcal{L} &= \mathcal{L}_{\mathrm{fm}}(\mathcal{V}_{\mathrm{rgb}}) + \omega(\tau)\,\mathcal{L}_{\mathrm{fm}}(\mathcal{V}_{\mathrm{geo}}),\\
\omega(\tau) &= \omega_{\max}\,\min\!\Big(1,\tfrac{\tau}{\tau_{0}}\Big),
\end{aligned}
\label{eq:joint_loss}
\end{equation}
where $\tau$ is the training iteration and $\omega(\tau)$ ramps linearly from $0$ to $\omega_{\max}$ over the first $\tau_{0}$ iterations. Normalizing each modality by its own view count keeps the appearance gradient magnitude stable, while the ramp starts the geometry weight at zero so that early depth gradients do not perturb the shared modulation and output layers.


\definecolor{stageblue}{RGB}{231,240,250}
\definecolor{sectionblue}{RGB}{243,247,252}
\definecolor{trainbg}{RGB}{238,247,240}
\definecolor{frozenbg}{RGB}{247,247,247}
\definecolor{rulegray}{RGB}{175,185,195}

\newcommand{\Trainable}{\cellcolor{trainbg}Trainable}
\newcommand{\Frozen}{\cellcolor{frozenbg}Frozen}

\begin{table*}[t]
\centering
\footnotesize
\setlength{\tabcolsep}{10.5pt}
\renewcommand{\arraystretch}{1.20}
\arrayrulecolor{rulegray}

\caption{\textbf{Training recipe of Puffin-World.} We report the per-stage hyperparameters, the trainable/frozen status of each module, and the data sampling ratio across tasks. \textit{Und.}, \textit{Gen.}, and \textit{Recon.} abbreviate understanding, generation, and reconstruction, respectively.}
\vspace{2pt}

\begin{tabularx}{\textwidth}{
    >{\raggedright\arraybackslash}X
    >{\centering\arraybackslash}m{0.125\textwidth}
    >{\centering\arraybackslash}m{0.125\textwidth}
    >{\centering\arraybackslash}m{0.125\textwidth}
    >{\centering\arraybackslash}m{0.125\textwidth}
}
\toprule[1.1pt]

&
\textbf{Stage I} &
\textbf{Stage II} &
\textbf{Stage III} &
\textbf{Stage IV} \\

\midrule[0.6pt]

\rowcolor{sectionblue}
\multicolumn{5}{l}{\hspace{2pt}\textbf{Hyperparameters}} \\[-1pt]

Learning rate
    & $1\times10^{-4}$
    & $2\times10^{-5}$
    & $5\times10^{-5}$
    & $2\times10^{-5}$ \\

LR Scheduler
    & \multicolumn{4}{c}{Cosine} \\

Weight Decay
    & \multicolumn{4}{c}{0.05} \\

Betas
    & \multicolumn{4}{c}{(0.9, 0.95)} \\

Optimizer
    & \multicolumn{4}{c}{AdamW} \\

Batch Size
    & 512
    & 512
    & 256
    & 128 \\

\addlinespace[4pt]

\rowcolor{sectionblue}
\multicolumn{5}{l}{\hspace{2pt}\textbf{Trainable Modules}} \\[-1pt]

MLP Projector
    & \Trainable
    & \Trainable
    & \Frozen
    & \Frozen \\

Connector
    & \Trainable
    & \Trainable
    & \Trainable
    & \Trainable \\

Condition Tokenizer
    & \Trainable
    & \Trainable
    & \Trainable
    & \Trainable \\

Vision Encoder
    & \Frozen
    & \Trainable
    & \Frozen
    & \Frozen \\

LLM
    & \Frozen
    & \Trainable
    & \Frozen
    & \Frozen \\

Diffusion Model
    & \Frozen
    & \Trainable
    & \Trainable
    & \Trainable \\

\addlinespace[4pt]

\rowcolor{sectionblue}
\multicolumn{5}{l}{\hspace{2pt}\textbf{Data Sampling Ratio}} \\[-1pt]

Image$\rightarrow$Text-Camera (single-view \textit{Und.})
    & 0.5
    & 0.5
    & \textendash
    & \textendash \\

Text-Camera$\rightarrow$Image (single-view \textit{Gen.})
    & 0.5
    & 0.5
    & 0.14
    & 0.09 \\

Image$\rightarrow$3D (cross-view \textit{Gen.})
    & \textendash
    & \textendash
    & 0.86
    & \textendash \\

Image$\rightarrow$3D (cross-view \textit{Gen.} \& \textit{Recon.})
    & \textendash
    & \textendash
    & \textendash
    & 0.91 \\

\bottomrule[1.1pt]
\end{tabularx}

\label{tab:training_recipe}
\end{table*}

To activate the geometry branch on a converged appearance model without degrading visual quality, we introduce two safeguards that preserve the appearance pathway. First, we adopt \emph{asymmetric} attention as shown in Figure~\ref{fig:framework}: appearance and text queries cannot attend to geometry tokens, while geometry tokens attend to all modalities. This prevents geometry features from leaking into the appearance stream while retaining full scene context for geometry prediction. Second, we add a zero-initialized learnable geometry-modality embedding after patch embedding. This keeps the model functionally identical to the original multi-view generator at initialization and gradually introduces geometry-specific features during training. At inference, appearance latents are decoded into images, whereas geometry latents are decoded and inverse-mapped from Hilbert color space to scalar depth.

\noindent\textbf{Long-horizon Exploration.}
The joint-attention sequence is trained with a fixed number of views, yet real exploration requires trajectories of arbitrary length. We therefore extend generation autoregressively in chunks: the model first generates a chunk of views, and the last generated view is then carried over as the reference view for the next chunk under a sliding context window, so that a trajectory of any length is produced by repeatedly advancing this window. The key design choice is that consecutive chunks are linked in the latent space: the next chunk conditions directly on the denoised target latent of the carried-over view rather than on a re-encoded pixel image, which avoids the compounding VAE encode--decode artifacts that would otherwise accumulate at every chunk boundary and drift over a long horizon. Physics propagation is applied across the whole sequence rather than within a single chunk, so all chunks share one gravity-aligned absolute frame, and the extended trajectory stays self-calibrated instead of accumulating orientation drift.

\subsection{Training Recipe}
\label{sec:method_training}
We adopt a four-stage training strategy to learn the whole framework. The first two stages perform cross-modal alignment and supervised fine-tuning (SFT) for single-view understanding and generation across the vision, language, and camera modalities. These stages cover scene and physical-state perception, as well as free-viewpoint spatial simulation. Building upon the acquired single-view capabilities, the final two stages focus on post-training for 3D world modeling, extending the model to sequential generation and reconstruction. The complete training recipe is summarized in Table~\ref{tab:training_recipe}, and each stage is detailed below.

\begin{itemize}
    \item \textbf{Stage I}. In this stage, we align the vision encoder with the LLM by training only the MLP projector to predict scene descriptions and camera parameters from input images. For generation, text and Omni-Camera representation are used to condition target image synthesis. We train the learnable queries and connector to map LLM hidden states into diffusion conditioning signals, while the vision encoder, LLM, and diffusion model remain frozen.

    \item \textbf{Stage II.} After cross-modal alignment, we unfreeze all modules except the VAE and fine-tune the full framework using the same data and objectives as in Stage I. To stabilize optimization and preserve the pretrained visual representations, we scale the gradients of the vision encoder by 0.1.
    
    \item \textbf{Stage III.} We then extend Puffin-World from single-view generation to cross-view 3D world modeling. Conditioned on one or more reference views and their corresponding Omni-Camera maps, the model jointly denoises all target views within a unified sequence. We train the model on trajectories with diverse and challenging camera motions, including large translations, in-place rotations, and full $360^\circ$ exploration. This stage is performed as post-training on the converged single-view model: the vision encoder and LLM are frozen, while the connector, diffusion backbone, and condition fusion module, remain trainable.

    \item \textbf{Stage IV.} Finally, we improve Puffin-World for geometry reconstruction by jointly modeling appearance and depth. We activate a geometry branch that generates the depth maps alongside the target RGB views. Both of them share the same flow-matching formulation but are combined through the separately weighted objective of Eq.~\ref{eq:joint_loss}. This stage keeps the same trainable modules as Stage~III and additionally trains the depth-modality embedding.
\end{itemize}

\subsection{Multi-Task Synergy}
\label{sec:multi_task_synergy}
Beyond the individual tasks above, Puffin-World exhibits promising multi-task synergy for complex, closed-loop applications that jointly involve perception, reasoning, and generation. In particular, \textit{mimic world exploration} supports 3D world generation from the same initial viewpoint while following a shared camera trajectory, whereas \textit{self-calibrated world exploration} automatically detects and corrects gravity misalignment through predicted actions and imagined target observations, resembling the closed-loop interaction paradigm of World-Action Models (WAMs)~\cite{DreamZero}. These capabilities highlight the potential of Puffin-World for virtual reality and embodied intelligence. Thanks to its unified multimodal framework, Puffin-World realizes these diverse capabilities within a single model without relying on external modules. We present the corresponding results in Section~\ref{sec:application}.
\section{Dataset}

\begin{figure*}[!t]
    \centering

    \includegraphics[width=\linewidth]{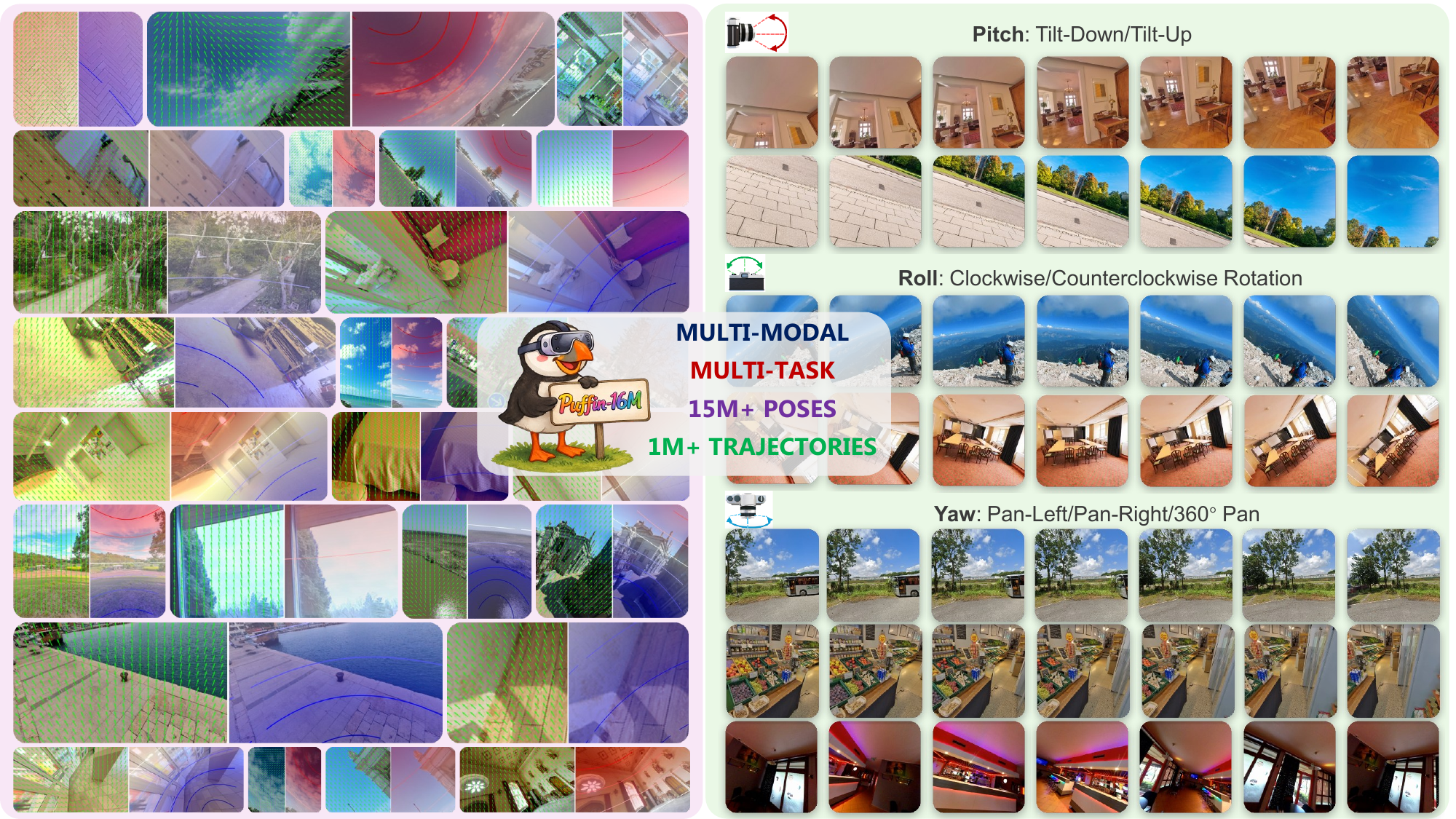}

    \caption{
    \textbf{An overview of the constructed Puffin-16M dataset.} It comprises 15 million vision-language-camera triplets with diverse resolutions and aspect ratios (left), along with 1 million trajectories featuring diverse and challenging rotational camera motions (right).
    }
    \label{fig:puffin_16m}
\end{figure*}

\definecolor{headerblue}{RGB}{205,224,245}      
\definecolor{subheaderblue}{RGB}{226,236,247}   
\definecolor{rowblue}{RGB}{242,246,250}         
\definecolor{puffinblue}{RGB}{226,236,247}      
\definecolor{puffinstrong}{RGB}{205,224,245}    
\definecolor{rulegray}{RGB}{145,155,165}        

\begin{table*}[!t]
\captionsetup{aboveskip=2pt}
\centering
\footnotesize
\setlength{\tabcolsep}{4.5pt}
\renewcommand{\arraystretch}{1.18}
\arrayrulecolor{rulegray}

\caption{\textbf{Dataset Comparisons.} 
Compared with previous datasets for single-task~\citep{lee2021ctrl, bogdan2018deepcalib, veicht2024geocalib, jin2023perspective, hold2018perceptual, bernal2025precisecam} and unified-task~\citep{puffin}. Our Puffin-16M dataset shows promising advantages across scale, versatility, spatial distribution, and motion diversity. For the camera parameters, we denote the intrinsic parameters: focal length ($f$), radial distortion coefficient ($\xi$); and the extrinsic parameters: roll ($\phi$), pitch ($\theta$), yaw ($\psi$), camera height ($h$).}

\resizebox{\textwidth}{!}{%
\begin{tabular}{lccccccccccc}

\toprule[1.2pt]

\multirow{2}{*}{\textbf{Dataset}}
&
\multirow{2}{*}{\textbf{Task Type}}
&
\multirow{2}{*}{\textbf{Intrinsics}}
&
\multirow{2}{*}{\textbf{Extrinsics}}
&
\multirow{2}{*}{\textbf{\# Frames}}
&
\multicolumn{7}{c}{\textbf{Details}}
\\

\cmidrule(lr){6-12}

&
&
&
&
&
Camera
&
Text
&
Reasoning
&
Single-View
&
Cross-View
&
Trajectory
&
Multi-Ratio
\\

\midrule[0.7pt]

GeoCalib~\citep{veicht2024geocalib}
& Understanding
& {$f, \xi$}
& {$\phi, \theta$}
& 37K
& \greencheck
& \redcheck
& \redcheck
& \greencheck
& \redcheck
& \redcheck
& \redcheck
\\

\rowcolor{rowblue}
CTRL-C~\citep{lee2021ctrl}
& Understanding
& $f$
& {$\phi, \theta$}
& 45K
& \greencheck
& \redcheck
& \redcheck
& \greencheck
& \redcheck
& \redcheck
& \redcheck
\\

Deepcalib~\citep{bogdan2018deepcalib}
& Understanding
& {$f, \xi$}
& -
& 67K
& \greencheck
& \redcheck
& \redcheck
& \greencheck
& \redcheck
& \redcheck
& \redcheck
\\

\rowcolor{rowblue}
ParamNet~\citep{jin2023perspective}
& Understanding
& $f$
& {$\phi, \theta$}
& 190K
& \greencheck
& \redcheck
& \redcheck
& \greencheck
& \redcheck
& \redcheck
& \redcheck
\\

Perceptual~\citep{hold2018perceptual}
& Understanding
& $f$
& {$\phi, \theta$}
& 390K
& \greencheck
& \redcheck
& \redcheck
& \greencheck
& \redcheck
& \redcheck
& \redcheck
\\

\rowcolor{rowblue}
PreciseCam~\citep{bernal2025precisecam}
& Generation
& {$f, \xi$}
& {$\phi, \theta$}
& 57K
& \greencheck
& \greencheck
& \redcheck
& \greencheck
& \redcheck
& \redcheck
& \redcheck
\\

\midrule[0.55pt]

\rowcolor{puffinblue}
Puffin-4M~\citep{puffin}
& Unified Multimodal
& $f$
& {$\phi, \theta, \psi$}
& 4M
& \greencheck
& \greencheck
& \greencheck
& \greencheck
& \greencheck
& \redcheck
& \redcheck
\\

\rowcolor{puffinblue}
\textbf{Puffin-16M}
& Unified Multimodal
& $f$
& {$\phi, \theta, \psi, h$}
& 16M
& \greencheck
& \greencheck
& \greencheck
& \greencheck
& \greencheck
& \greencheck
& \greencheck
\\

\bottomrule[1.2pt]

\end{tabular}%
}

\label{tab:dataset_cp}
\captionsetup{justification=raggedright,singlelinecheck=false}
\end{table*}

\subsection{Puffin-16M}
Datasets with precise, high-quality alignment among visual content, semantics, and physical properties remain scarce in the multimodal community. Puffin-4M~\cite{puffin} narrows this gap with 4 million vision-language-camera triplets, but still faces three limitations: (i) its scale remains insufficient for larger models; (ii) all images are fixed at $512\times512$, restricting multi-resolution generation and requiring non-square inputs to be centrally cropped and resized for camera understanding, which may discard useful content; and (iii) its spatial coverage is dominated by single-view data, with only limited cross-view pairs and no long-range camera trajectories. To overcome these limitations, we introduce Puffin-16M, consisting of 15 million vision-language-camera triplets with diverse resolutions and aspect ratios, together with 1 million trajectories featuring diverse and challenging rational camera motions. We denote these two subsets as Puffin-Cam-15M and Puffin-Traj-1M, respectively, and summarize the comparison with prior datasets in Tab.~\ref{tab:dataset_cp}.

\noindent\subsubsection{Puffin-Cam-15M}
The construction pipeline follows a procedure similar to that of Puffin-4M~\cite{puffin}, including panoramic data collection and preprocessing, perspective image generation, and scene and reasoning captioning. Specifically, we expand the source panoramic images used in Puffin-4M from 200K to 900K, covering diverse scenarios such as outdoor and indoor scenes, synthetic and real-world environments, and self-driving street views across different countries. Such diversity in the source panoramas provides a strong foundation for scaling large multimodal models, particularly for multimodal spatial intelligence. Due to variations in $360^\circ$ camera calibration and acquisition stability, some panoramas exhibit geometric distortions and misalignment, especially in consumer-grade capture and autonomous-driving onboard settings. We therefore apply geometric correction techniques based on line segmentation and vanishing point estimation, aligning the panoramas with the gravity direction and improving structural consistency. Compared with Puffin-4M, we exclude Stanford2D3D~\cite{Stanford2d3d} from the source panoramic data to ensure fair evaluation on this benchmark for camera-to-world understanding. 

Following the standard protocol adopted in recent studies~\citep{veicht2024geocalib, puffin}, we generate diverse perspective views from each panorama using a virtual pinhole camera with varying intrinsic and extrinsic parameters. Specifically, we uniformly sample roll and pitch from $[-45^\circ,45^\circ]$ and vertical FoV from $[20^\circ,105^\circ]$, while adaptively determining the number of crops according to the resolution of each source panorama. This process yields approximately 15M perspective images with precise camera parameters and diverse aspect ratios (\textit{e.g.}, 1:1, 2:3, 3:2, 3:4, 4:3, 9:16, and 16:9). We retain the full set for camera-to-world understanding, while further curating around 8M images based on the realism and aesthetic quality of the source panoramas for camera-controllable image generation. We then employ advanced multimodal large language models (\textit{e.g.}, Qwen3-VL-8B-Instruct and Qwen3-VL-32B-Instruct~\cite{bai2025qwen3}) to annotate each image with high-quality captions that include detailed scene semantics and structured chain-of-thought reasoning over spatial relationships, camera parameters, and underlying physical principles. In addition, we annotate the camera height of each panorama into five levels, namely, underwater shot, low-position shot, eye-level shot, high-position shot, and aerial shot, to facilitate more advanced spatial reasoning in future work. The detailed captioning prompts are provided in Figure~\ref{fig:camera_height_caption} of the Appendix.

\noindent\subsubsection{Puffin-Traj-1M}

As shown in Figure~\ref{fig:dataset_cam_dis}, most existing 3D world modeling datasets~\cite{ling2024dl3dv, RealEstate10K, dai2017scannet, MVS-Synth, Tartanair, roberts2021hypersim} are limited in terms of camera rotational diversity. Specifically, roll is typically restricted to $[-5^\circ, 5^\circ]$ and pitch to $[-10^\circ, 10^\circ]$. This limitation arises because most datasets are captured using handheld cameras, where the primary source of variation comes from camera translation rather than rotation. Consequently, models trained on such data often struggle to understand challenging camera motions and accurately simulate the corresponding future world states, particularly in real-world scenarios.

However, a spatially intelligent agent should be able to actively explore its environment from arbitrary viewpoints. To this end, we construct Puffin-Traj-1M, a large-scale dataset comprising one million trajectories with diverse and challenging camera motions. The dataset includes long-horizon trajectories featuring continuous look-down, look-up, clockwise rotation, counterclockwise rotation, and full $360^\circ$ surrounding-view exploration.

Specifically, we adopt the same data construction pipeline as Puffin-Cam-15M, while introducing an additional camera extrinsic parameter, namely yaw, to enable cross-view exploration and arbitrary-viewpoint camera rotations. All camera extrinsics are sampled from broad ranges: roll and pitch are uniformly sampled from $[-45^\circ, 45^\circ]$, while yaw is sampled from $[0^\circ, 360^\circ)$. For each trajectory, one extrinsic parameter is randomly selected to define the camera motion pattern, and frames are generated either in a single-pass or recursive-pass manner. The single-pass strategy provides long-range scene exploration trajectories, whereas the recursive-pass strategy encourages the emergence of spatial memory and facilitates the preservation of 3D consistency over extended horizons. The vertical field-of-view is randomly sampled from $[60^\circ, 100^\circ]$.

\subsection{Additional Captions for Public Datasets}
\label{sec:datasets_add_captions}
Beyond contributing our self-constructed dataset, we also enrich the commonly used public datasets~\cite{ling2024dl3dv, RealEstate10K, dai2017scannet, MVS-Synth, Tartanair, roberts2021hypersim} in terms of the physical and geometric cues.

\noindent\subsubsection{Physical Caption}
Although these datasets provide accurate per-frame camera poses, they primarily characterize relative motion across views and lack the absolute orientation required to anchor observations to the physical world. Models trained solely on such trajectories can therefore learn scene-centric motion patterns without establishing a persistent physical reference, making gravity direction, horizon location, and absolute camera orientation ambiguous. To address this limitation, we leverage Puffin-World to estimate absolute camera parameters for every frame in the public training datasets, from which gravity fields and latitude maps are derived as physical priors for our physics propagation mechanism. Beyond the training data, we further annotate a broad collection of widely used public datasets, including ImageNet~\cite{russakovsky2014imagenet}, GPIC~\cite{chandrasegaran2026gpic}, Objects365~\cite{shao2019objects365}, CC12M~\citep{changpinyo2021conceptual}, and Megalith-10M~\citep{matsubara2024megalith10m}. In total, our release covers 28 datasets and approximately 44.5M images with absolute camera annotations, providing large-scale supervision for gravity-aware perception, physically grounded world modeling, and future research on spatial intelligence. An overview of the annotated datasets is shown in Figure~\ref{fig:camera_dataset} and the details of each annotated dataset are listed in Table~\ref{tab:camera_datasets} of the Appendix.

\noindent\subsubsection{Geometric Caption}
For public synthetic datasets~\cite{MVS-Synth, Tartanair, roberts2021hypersim}, we directly use the provided dense geometric annotations. For real-world datasets, however, the available depth supervision is often incomplete. Although ScanNet~\cite{dai2017scannet} and DL3DV~\cite{ling2024dl3dv} provide camera poses, their depth maps may be sparse or missing. In ScanNet, depth measurements suffer from sensor holes and invalid observations. For DL3DV, the depth maps obtained through MVS reconstruction~\cite{schonberger2016colmap}, following CUT3R~\cite{wang2025cut3r} are incomplete.  To obtain dense geometric supervision, we run DA3~\cite{depthanything3} on every frame to predict dense depth maps. We then align these predictions with the available sparse depth annotations, preserving the reliable scale of the original measurements while filling in missing regions. This yields dense and geometrically consistent depth annotations for real-world videos. The comparison of the original sparse depth label, our re-annotated dense depth label, and our predicted depth is visualized in Figure~\ref{fig:depth_cp} of the Appendix. 
\section{Experiments}
\subsection{Implementation Details}
\subsubsection{Network Configuration}

We develop two architectural variants, namely Puffin-World-Base and Puffin-World-Pro, targeting different trade-offs between physical world understanding and world simulation capabilities. Specifically, Puffin-World-Base is initialized with the pretrained C-RADIOv3-H~\citep{heinrich2025radiov2}, Qwen2.5-7B-Instruct~\citep{qwen2024qwen2}, and SD3.5-Medium-2.5B~\citep{sd3p5} as its geometry-aware vision encoder, large language model, and diffusion backbone, respectively. To further enhance the generative fidelity, Puffin-World-Pro adopts the pretrained C-RADIOv4-H~\citep{radiov4}, Qwen2.5-1.5B-Instruct~\citep{qwen2024qwen2}, and SD3.5-Large-8.1B~\citep{sd3p5} for the corresponding components.
For both variants, a set of 64 learnable queries and a lightweight six-layer Transformer connector are used to project the LLM hidden states into diffusion conditioning embeddings.
In addition to the above two variants, we also release Puffin-World-Caption, an expert model dedicated to annotating physical world states, such as absolute camera rotation parameters (roll and pitch) and the intrinsic parameter (FoV). To support large-scale dataset construction, Puffin-World-Caption is built upon the pretrained C-RADIOv3-H~\citep{heinrich2025radiov2} and Qwen3.5-0.8B-Instruct~\citep{team2026qwen3p5}, providing an effective balance between annotation accuracy and inference efficiency for large-scale physical-world labeling. 

\begin{table*}[t]
\centering
\footnotesize
\renewcommand{\arraystretch}{1.0}%
\setlength\tabcolsep{7.9pt}%
\caption{\textbf{Evaluation results on camera-to-world understanding.} The comparison methods are evaluated on the public datasets: Stanford2D3D~\citep{Stanford2d3d}, MegaDepth~\citep{Megadepth}, TartanAir~\citep{Tartanair}, and LaMAR~\citep{sarlin2022lamar}. Our Puffin-World outperforms the previous methods on all median error metrics and most AUC metrics. Note that {AnyCalib\textsuperscript{\rm $\dagger$}}~\citep{tirado2025anycalib} is specifically designed for intrinsic parameter estimation (\textit{i.e.}, FoV estimation) and is included for reference. We color the \colorbox{tabfirst}{\textbf{best}} and \colorbox{tabsecond}{second best} results.}
\begin{tabular}{clcccccccccccc}
\toprule
&\multirow{2}{*}[-.4em]{\textbf{Approach}}
& \multicolumn{4}{c}{\textbf{Roll [degrees]}} 
& \multicolumn{4}{c}{\textbf{Pitch [degrees]}} 
& \multicolumn{4}{c}{\textbf{FoV [degrees]}}\\
\cmidrule(lr){3-6}
\cmidrule(lr){7-10}
\cmidrule(lr){11-14}
&& error\,$\downarrow$ & \multicolumn{3}{c}{AUC\,$\triangleright$\,1/5/10\degree\,$\uparrow$}
& error\,$\downarrow$ & \multicolumn{3}{c}{AUC\,$\triangleright$\,1/5/10\degree\,$\uparrow$}
& error\,$\downarrow$ & \multicolumn{3}{c}{AUC\,$\triangleright$\,1/5/10\degree\,$\uparrow$} \\
\midrule
\multirow{10}{*}{\begin{sideways}\textbf{Stanford2D3D~\citep{Stanford2d3d}}\end{sideways}}
&{DeepCalib}~\citep{lopez2019deep}      &          \01.59 &          33.8 &          63.9 & \cthird  79.2 &          \02.58 &          21.6 &          46.9 &          65.7 &          \06.67 &          \08.1 &           20.6 &          37.6 \\
&{Perceptual}~\citep{hold2018perceptual}              &          \02.08 &          26.8 &          53.8 &          70.7 &          \03.17 &          21.5 &          41.8 &          57.8 &           13.84 &          \02.8 &          \07.7 &          16.1 \\
&{CTRL-C}~\citep{lee2021ctrl}                       &          \03.04 &          23.2 &          43.0 &          56.9 &          \03.43 &          18.3 &          38.6 &          53.8 &          \08.50 &          \07.7 &           18.2 &          31.5 \\
&{MSCC}~\citep{Song2024MSCC}                  &          \03.43 &          13.5 &          36.8 &          57.3 &          \02.64 &          22.6 &          45.0 &          60.5 & \cthird  \05.81 & \cthird  \09.6 & \cthird   23.8 & \cthird  41.6 \\
&{ParamNet}~\citep{jin2023perspective} & \cthird  \01.14 & \cthird  44.6 & \cthird  73.9 & \cthird 84.8 & \cthird  \01.94 & \cthird  29.2 & \cthird  56.7 & \cthird 73.1 &          \09.01 &          \05.8 &           14.3 &          27.8 \\
&{SVA}~\citep{lochman2021minimal}                            &              -  &          21.7 &          24.6 &          25.8 &              -  &          15.4 &          19.9 &          22.4 &              -  &          \06.2 &           11.5 &          15.2 \\
&{UVP}~\citep{pautrat2023vanishing}    & \cthird \00.52 & \cthird 65.3 & \cthird 74.6 &          79.1 & \cthird \00.95 & \cthird 51.2 & \cthird 63.0 & \cthird  69.2 & \cthird \03.65 & \csecond   22.2 & \cthird  39.5 & \cthird 51.3 \\
&{GeoCalib}~\citep{veicht2024geocalib}                                   & \csecond  \00.40 & \csecond  83.1 & \csecond  91.8 & \csecond  94.8 & \csecond  \00.93 & \csecond  52.3 & \csecond  74.8 & \csecond  84.6 & \csecond  \03.21 & \cthird  17.4 & \csecond   40.0 & \csecond  59.4 \\
&{AnyCalib\textsuperscript{\rm $\dagger$}}~\citep{tirado2025anycalib}                               & - & - & - & - & - & - & - & - & \02.55 & 21.1 & 46.8 & 64.6 \\
&\textbf{Puffin-World}                                   & \cfirst  \00.29 & \cfirst  93.1 & \cfirst  97.4 & \cfirst  98.5 & \cfirst  \00.53 & \cfirst  73.3 & \cfirst  88.8 & \cfirst  94.0 & \cfirst  \01.62 & \cfirst  34.5 & \cfirst   61.1 & \cfirst  76.4 \\
\midrule
\multirow{11}{*}{\begin{sideways}\textbf{MegaDepth~\citep{Megadepth}}\end{sideways}}
&{DeepCalib}~\citep{lopez2019deep}      &          \01.41 &          34.6 &          65.4 &          79.4 &          \05.19 &          11.9 &          27.8 &          44.8 &           11.14 &          \05.6 &           12.1 &          22.9 \\
&{Perceptual}~\citep{hold2018perceptual}              &          \01.07 &          47.9 &          72.4 &          83.2 & \cthird \03.49 & \cthird  19.8 & \cthird 39.1 & \cthird  54.2 &           13.40 &          \02.9 &          \08.2 &          16.8 \\
&{CTRL-C}~\citep{lee2021ctrl}                       & \cthird  \00.88 & \cthird  54.5 & \cthird  75.0 & \cthird  84.2 &          \04.80 &          16.6 &          33.2 &          46.5 &           18.65 &          \02.0 &          \05.8 &          12.8 \\
&{MSCC}~\citep{Song2024MSCC}                  &          \00.90 &          53.1 &          72.8 &          82.1 &          \05.73 &          19.0 &          33.2 &          44.3 & \cthird  10.80 &          \06.0 &           14.6 & \cthird  26.2 \\
&{ParamNet}~\citep{jin2023perspective} &          \01.17 &          43.4 &          70.7 &          82.2 &          \03.99 &          15.4 &          34.5 &          53.3 &           11.01 &          \03.2 &           10.1 &          21.3 \\
&{SVA}~\citep{lochman2021minimal}                            &              -  &          31.9 &          35.0 &          36.2 &              -  &          13.6 &          20.6 &          24.9 &              -  & \cthird \09.4 & \cthird   16.1 &          21.1 \\
&{UVP}~\citep{pautrat2023vanishing}    & \cthird \00.51 & \cthird 69.2 & \cthird 81.6 & \cthird 86.9 &          \04.59 & \cthird 21.6 &          36.2 &          47.4 & \cthird   10.92 & \cthird  \08.2 & \cthird  18.7 & \cthird 29.8 \\
&{GeoCalib}~\citep{veicht2024geocalib}                                & \cthird  \00.36 & \cthird  82.6 & \cthird  90.6 & \cthird  94.0 & \cthird  \01.94 & \cthird  32.4 & \cthird  53.3 & \cthird  67.5 & \cthird  \04.46 & \cthird   13.6 & \cthird   31.7 & \cthird  48.2 \\
&{AnyCalib\textsuperscript{\rm $\dagger$}}~\citep{tirado2025anycalib}                                & - & - & - & - & - & - & - & - & \03.14 & 19.4 & 40.8 & 59.1 \\
&Puffin~\citep{puffin}                                  & \csecond  \00.32 & \csecond  84.9 & \csecond  93.4 & \csecond  96.2 & \csecond  \01.08 & \csecond  47.6 & \csecond  68.2 & \csecond  79.4 & \csecond  \02.42 & \csecond  23.9 & \csecond   47.8 & \csecond  64.1 \\
&\textbf{Puffin-World}                                   & \cfirst  \00.28 & \cfirst  87.7 & \cfirst  94.2 & \cfirst  96.7 & \cfirst  \01.01 & \cfirst  49.7 & \cfirst  69.3 & \cfirst  79.6 & \cfirst  \02.41 & \cfirst  24.8 & \cfirst   48.8 & \cfirst  65.5 \\
\midrule
\multirow{11}{*}{\begin{sideways}\textbf{TartanAir~\citep{Tartanair}}\end{sideways}}
&{DeepCalib}~\citep{lopez2019deep}      &          \01.95 &          24.7 &          55.4 &          71.5 &          \03.27 &          16.3 &          38.8 &          58.5 &          \08.07 &          \01.5 &          \08.8 &          27.2 \\
&{Perceptual}~\citep{hold2018perceptual}              &          \02.24 &          23.2 &          48.6 &          66.7 &          \02.86 &          23.5 &          44.6 &          61.5 &           15.06 &          \05.1 &          \08.9 &          17.1 \\
&{CTRL-C}~\citep{lee2021ctrl}                       &          \01.68 &          32.8 &          59.1 & \cthird 74.1 & \cthird \02.39 & \cthird  24.6 & \cthird  48.6 & \cthird 65.2 & \cthird \05.64 & \cthird   10.7 & \cthird   25.4 & \cthird 43.5 \\
&{MSCC}~\citep{Song2024MSCC}                  &          \03.50 &          15.0 &          37.2 &          57.7 &          \03.48 &          18.8 &          38.6 &          54.3 &           11.18 &          \04.4 &           11.8 &          23.0 \\
&{ParamNet}~\citep{jin2023perspective} & \cthird  \01.63 & \cthird  34.5 & \cthird  59.2 & \cthird  73.9 &          \03.05 &          19.4 &          42.0 &          60.3 &          \08.21 &          \06.0 &           16.8 &          31.6 \\
&{SVA}~\citep{lochman2021minimal}                            &          \09.48 &          32.4 &          39.6 &          44.1 &           18.46 &          21.2 &          28.8 &          34.5 &           43.01 &          \08.8 &           16.1 &          21.6 \\
&{UVP}~\citep{pautrat2023vanishing}    & \cthird \00.89 & \cthird 52.1 & \cthird 64.8 &          71.9 & \cthird  \02.48 & \cthird 36.2 & \cthird 48.8 &          58.6 &          \09.15 & \cthird   15.8 & \cthird  25.8 &          35.7 \\
&{GeoCalib}~\citep{veicht2024geocalib}                                   & \cthird  \00.43 & \cthird  71.3 & \cthird  83.8 & \cthird  89.8 & \cthird  \01.49 & \cthird  38.2 & \cthird  62.9 & \cthird  76.6 & \csecond  \04.90 & \cthird  14.1 & \csecond   30.4 & \csecond  47.6 \\
&{AnyCalib\textsuperscript{\rm $\dagger$}}~\citep{tirado2025anycalib}                                & - & - & - & - & - & - & - & - & \03.62 & 15.5 & 36.4 & 55.1 \\
& Puffin~\citep{puffin}                               & \csecond  \00.40 & \csecond  71.7 & \csecond  86.2 & \csecond  92.1 & \csecond  \00.95 & \csecond  51.0 & \csecond  68.2 & \csecond  79.3 & \cthird  \07.48 & \csecond  16.3 & \cthird   28.5 & \cthird  39.0 \\
&\textbf{Puffin-World}                                & \cfirst  \00.31 & \cfirst  80.1 & \cfirst  90.0 & \cfirst  94.2 & \cfirst  \00.67 & \cfirst  60.2 & \cfirst  78.1 & \cfirst  87.2 & \cfirst  \02.34 & \cfirst  26.6 & \cfirst   46.4 & \cfirst  59.5 \\
\midrule
\multirow{11}{*}{\begin{sideways}\textbf{LaMAR~\citep{sarlin2022lamar}}\end{sideways}}
&{DeepCalib}~\citep{lopez2019deep}      &          \01.15 &           44.1 &           73.9 &           84.8 &          \04.68 &           10.8 &           28.3 &           49.8 &           10.93 &          \00.7 &           13.0 &           24.0 \\
&{Perceptual}~\citep{hold2018perceptual}              &          \01.29 &           40.0 &           68.9 &           81.6 &          \02.83 &           21.2 &           44.7 &           62.6 &           17.78 &          \03.0 &          \05.3 &           10.7 \\
&{CTRL-C}~\citep{lee2021ctrl}                       &          \01.20 &           43.5 &           70.9 &           82.5 & \cthird  \01.94 & \cthird   27.6 & \cthird   54.7 & \cthird  70.2 & \cthird  \05.64 & \cthird  \09.8 & \cthird   24.6 & \cthird   43.2 \\
&{MSCC}~\citep{Song2024MSCC}                  &          \01.44 &           39.6 &           60.7 &           72.8 &          \03.02 &           20.9 &           41.8 &           55.7 &           14.78 &          \03.2 &          \08.3 &           16.8 \\
&{ParamNet}~\citep{jin2023perspective} & \cthird  \00.93 & \cthird   51.7 & \cthird   77.0 & \cthird  86.0 &          \02.15 &           27.0 &           52.7 & \cthird  70.2 &           14.71 &          \02.8 &          \06.8 &           14.3 \\
&{SVA}~\citep{lochman2021minimal}                            &              -  &          \08.6 &          \09.2 &          \09.7 &              -  &          \03.4 &          \05.7 &          \07.0 &              -  &          \01.2 &          \02.7 &          \04.1 \\
&{UVP}~\citep{pautrat2023vanishing}    & \cthird \00.38 & \cthird  72.7 & \cthird  81.8 & \cthird   85.7 & \cthird \01.34 & \cthird  42.3 & \cthird  59.9 & \cthird   69.4 & \cthird \05.57 & \cthird  15.6 & \cthird  30.6 & \cthird  43.5 \\
&{GeoCalib}~\citep{veicht2024geocalib}                                 & \csecond  \00.28 & \cfirst   86.4 & \cfirst   92.5 & \csecond   95.0 & \csecond  \00.87 & \cthird   55.0 & \cthird   76.9 & \cthird   86.2 & \csecond  \03.03 & \cfirst   19.1 & \csecond   41.5 & \cfirst   60.0 \\
&{AnyCalib\textsuperscript{\rm $\dagger$}}~\citep{tirado2025anycalib}                                & - & - & - & - & - & - & - & - & \02.25 & 24.6 & 51.6 & 70.5 \\
&Puffin~\citep{puffin}
& \cthird  \00.38 & \cthird  80.6 & \cthird  89.8 & \cthird  93.5 & \cfirst  \00.71 & \csecond  61.7 & \csecond  78.9 & \csecond  86.4 & \cthird  \03.62 & \cthird  17.0 & \cthird   37.3 & \cthird  53.1 \\
&\textbf{Puffin-World}
& \cfirst  \00.26 & \csecond  85.8 & \cfirst  92.5 & \cfirst  95.3 & \cfirst  \00.71 & \cfirst  63.5 & \cfirst  81.2 & \cfirst  88.2 & \cfirst  \02.73 & \csecond  19.0 & \cfirst   43.0 & \csecond  59.4 \\

\bottomrule
\end{tabular}
\label{tab:cam_und_evaluation}
\end{table*}

\subsubsection{Training/Inference Settings}
The whole training process comprises four stages, whose per-stage learning rates, batch sizes, data sampling ratios, and trainable modules are summarized in Table~\ref{tab:training_recipe}. Images and mapped depth are encoded by the frozen VAE into $16$-channel latents. For single-view camera-controllable generation, we train at the native resolutions and aspect ratios of Puffin-Cam-15M, whereas for 3D world modeling we crop all views to $640\times640$. Each sequence contains a total of $8$ views: the number of reference (initial) views is randomly sampled from $1$ to $3$ in Stage~III and fixed to $1$ in Stage~IV, with the remaining views serving as generation targets. Diffusion timesteps are drawn from a logit-normal distribution ($\mu{=}0$, $\sigma{=}1$) with a resolution-dependent shift, and Stage~III and Stage~IV adopt a larger shift to bias training toward higher-noise timesteps that benefit cross-view consistency. For the Omni-Camera representation, the absolute representation is placed before the relative representation in the channel order, resulting in a $9$-channel dense camera map $\mathbf{C}\in\mathbb{R}^{H\times W\times 9}$ with the same spatial resolution as the image.

For the multi-view post-training Stage~III, the model is trained on a mixture of trajectory data to learn diverse camera motions involving both translation (DL3DV~\citep{ling2024dl3dv} and RealEstate10K~\citep{RealEstate10K}) and rotation (Puffin-Traj-1M), together with single-view camera-controllable generation samples from Puffin-Cam-15M. The single-view batches serve as rehearsal data to preserve the camera controllability acquired in Stage~II. We adopt a multi-source sampling strategy, where homogeneous mini-batches are drawn from each data source according to fixed repeat factors of $3{:}2{:}1{:}1$. To reduce the training cost of the proposed physics propagation, we pre-compute the absolute camera fields for all training frames offline using the physics perception capability (at Stage-II) of Puffin-World. During training, each view is therefore directly conditioned on its corresponding Omni-Camera map. The condition fusion module is warm-started from the single-view model. Stage~IV then inherits the Stage-III weights and activates the joint appearance–geometry branch (Sec.~\ref{sec:world_modeling}), widening the multi-view mixture with Hypersim~\citep{roberts2021hypersim}, MVS-Synth~\citep{MVS-Synth}, TartanAir~\citep{Tartanair}, and ScanNet~\citep{dai2017scannet} while keeping the same source-balanced sampling and single-view rehearsal. The depth loss weight is ramped from zero to $\omega_{\max}{=}1$ over the first $3{,}000$ iterations (Eq.~\ref{eq:joint_loss}). 

We enable classifier-free guidance (CFG) through condition dropout during training. For single-view camera-controllable generation, we jointly replace the text prompt with an empty string and zero the Omni-Camera condition with probability $0.1$. For multi-view world generation, the same joint dropout is applied to the text and all per-view Omni-Camera maps. We additionally drop the absolute Perspective Field with probability $0.15$ while retaining the relative ray map, reducing over-reliance on absolute cues and enabling component-wise guidance. At inference, camera-to-world understanding uses greedy decoding, while generation employs $50$ sampling steps. We set the CFG scale to $4.5$ and $2.0$ for single-view camera-controllable image generation and 3D world modeling, respectively.

\subsection{Camera-to-World Understanding Results} 
Following prior works, we compare Puffin-World against a diverse set of camera-to-world understanding methods, including learning-based approaches such as DeepCalib~\citep{lopez2019deep}, Perceptual~\citep{hold2018perceptual}, CTRL-C~\citep{lee2021ctrl}, MSCC~\citep{Song2024MSCC}, ParamNet~\citep{jin2023perspective}, GeoCalib~\citep{veicht2024geocalib}, and Puffin~\cite{puffin}, as well as classical geometric methods including SVA~\citep{lochman2021minimal} and UVP~\citep{pautrat2023vanishing}. For each image, gravity estimation performance is measured by the angular errors of roll and pitch, while camera intrinsic estimation is evaluated using the error in vertical  (vFoV). Following standard practice, we report both the median error and the Area Under the Recall Curve (AUC) at error thresholds of $1^\circ$, $5^\circ$, and $10^\circ$. Evaluations are conducted on four widely adopted benchmarks: MegaDepth~\citep{Megadepth}, TartanAir~\citep{Tartanair}, LaMAR~\citep{sarlin2022lamar}, and Stanford2D3D~\cite{Stanford2d3d}.

\begin{figure*}[!htbp]
    \centering

    \includegraphics[width=\linewidth]{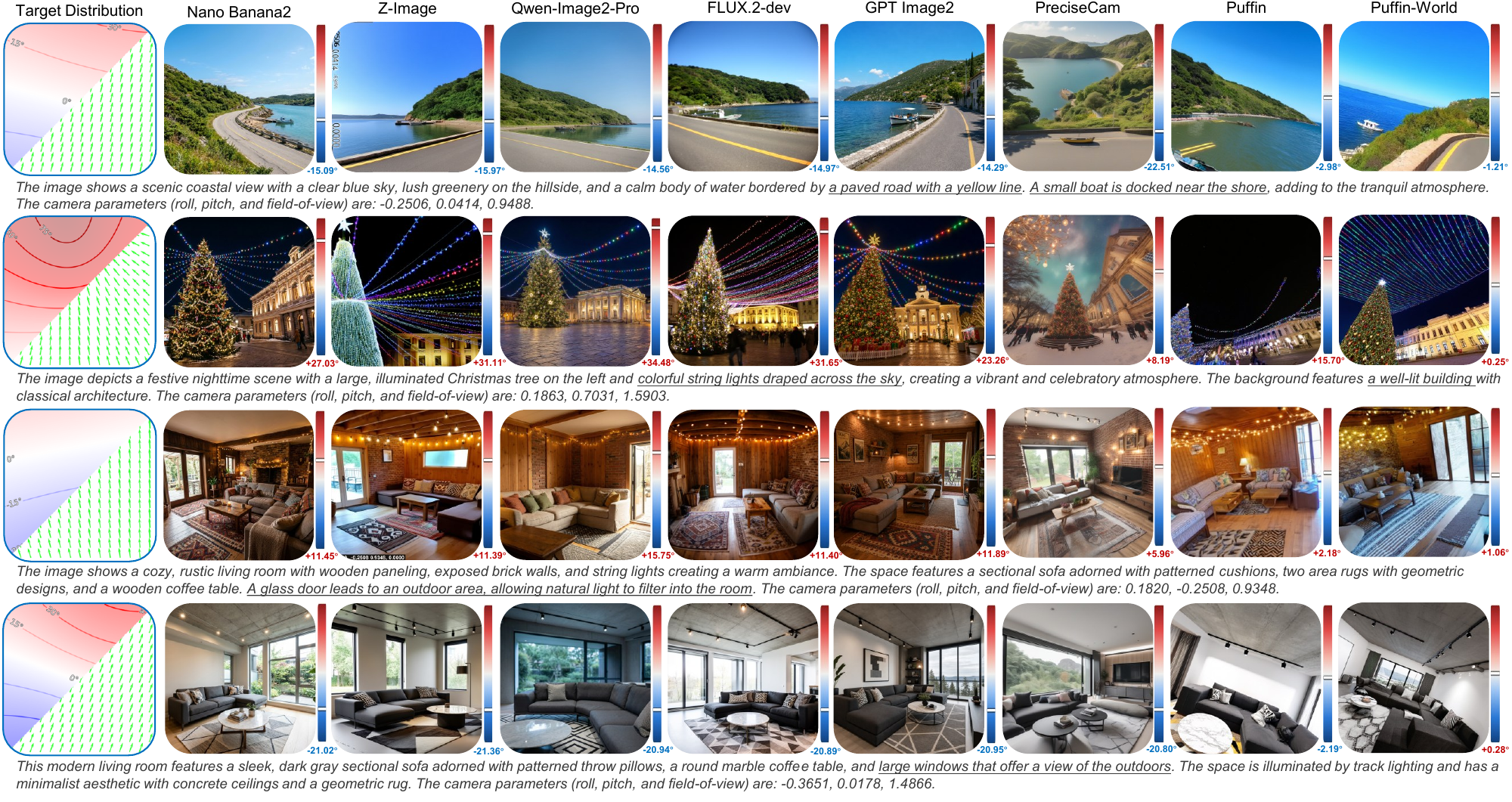}

    \caption{
    \textbf{Comparison results on camera-controllable text-to-image generation for free-viewpoint spatial simulation.} For each result, we visualize the angular error between the simulated gravity and the ground truth; values closer to $0^\circ$ indicate better spatial alignment. Essential elements are \uline{underlined} in the text prompt.
    }
    \label{fig:cam_gen_cp}
\end{figure*}

As shown in Table~\ref{tab:cam_und_evaluation}, Puffin-World consistently outperforms all competing methods in terms of median error and achieves the best performance on the majority of AUC metrics across all four benchmarks. These results demonstrate its strong capability to perceive and ground observations within the physical world. In particular, Puffin-World surpasses Puffin~\citep{puffin} with a noticeable margin, showing the meaningful and progressive benefits of scaling the model and data for this task. We note that AnyCalib~\cite{tirado2025anycalib} is specifically designed for camera intrinsic estimation, particularly FoV prediction, and is included only as a specialized reference baseline. Despite its task-specific design, Puffin-World still outperforms AnyCalib on most benchmarks, highlighting the effectiveness and generalizability of our unified framework.

\begin{table*}[t]
\centering
\footnotesize
\renewcommand{\arraystretch}{1.0}%
\setlength\tabcolsep{4pt}%
\caption{\textbf{Camera-controllable generation evaluation on Puffin-Cam-Bench.}}
\begin{tabular}{clcccccccc}
\toprule
& \multirow{2}{*}[-.4em]{\textbf{Approach}}
& \multicolumn{2}{c}{\textbf{Up Vector [degrees]}}
& \multicolumn{2}{c}{\textbf{Latitude [degrees]}}
& \multicolumn{2}{c}{\textbf{Gravity [degrees]}}
& \multicolumn{1}{c}{{\textbf{Visual Quality}}}\\
\cmidrule(lr){3-4}\cmidrule(lr){5-6}\cmidrule(lr){7-8}\cmidrule(lr){9-10}
&& mean error $\downarrow$ & median error $\downarrow$
& mean error $\downarrow$ & median error $\downarrow$
& mean error $\downarrow$ & median error $\downarrow$
& {FID $\downarrow$} & \\
\midrule
\multirow{1}{*}{\begin{sideways}\textbf{}\end{sideways}}
& GPT Image2~\citep{GPT-Image2}
& 24.13 & 23.44 & 16.42 & 15.98 &  28.83 &  29.38 &  99.36 &  \\
& Nano Banana2~\citep{NanoBanana2}
&  24.01 &  23.19 &  15.98 &  15.52 &  28.00 &  26.92 &  90.11 &  \\
& Qwen-Image2-Pro~\citep{Qwen-Image-2}
&  24.12 &  23.51 &  16.45 &  15.53 &  28.57 &  28.69 &  98.51 &  \\
& FLUX.2-dev~\citep{flux-2-2025}
&  24.28 &  23.62 &  15.75 &  15.24 &  28.31 &  28.36 &  97.06 &  \\
& Z-Image~\citep{zimage}
&  24.64 &  24.04 &  16.96 &  16.09 &  29.41 &  28.82 &  98.33 &  \\
& PreciseCam~\citep{bernal2025precisecam}
& 13.37 & 12.61 & 12.31 & 12.01 & 17.07 & 14.01 &  90.89 &  \\
& Puffin~\citep{puffin}
& \csecond \03.86 & \csecond \03.48 & \csecond \04.77 & \csecond \04.43 & \csecond \04.92 & \csecond \02.87 &  \csecond 80.29 & \\
& \textbf{Puffin-World}
& \cfirst \00.96 & \cfirst \00.84 & \cfirst \01.34 & \cfirst \01.26 & \cfirst \01.32 & \cfirst \00.79 &  \cfirst 75.93 & \\
\bottomrule
\end{tabular}
\label{tab:cam_gen_evaluation_fid}
\end{table*}

\subsection{Camera-Controllable Generation Results} 

We compare Puffin-World against recent state-of-the-art multimodal generation models, including GPT Image2~\citep{GPT-Image2}, Nano Banana 2~\cite{NanoBanana2}, Qwen-Image2-Pro~\cite{Qwen-Image-2}, FLUX.2-dev~\cite{flux-2-2025}, and Z-Image~\cite{zimage}, as well as specialized camera-controllable image generation methods~\citep{bernal2025precisecam, puffin}. All methods are evaluated using the same text prompts and camera specifications.

To quantitatively assess camera controllability, we leverage Puffin-World to estimate the camera parameters of each generated image and derive the corresponding pixel-wise perspective fields. We then compare these estimated fields against the GT fields and report both mean and median angular errors for the up vector, gravity direction, and latitude map, all measured in degrees. In addition, we report the Fréchet Inception Distance (FID) to evaluate the overall visual fidelity and realism of the generated images. Since no existing benchmark provides paired text descriptions and precise camera parameters across diverse camera configurations and aspect ratios, we introduce Puffin-Cam-Bench to fill this gap. It contains 600 text–camera specification pairs covering a wide range of scenes, viewpoints, camera poses, fields of view, and aspect ratios, enabling comprehensive evaluation of camera-controllable generation for free-viewpoint spatial simulation.

Table~\ref{tab:cam_gen_evaluation_fid} summarizes the quantitative results, while Figure~\ref{fig:cam_gen_cp} provides qualitative comparisons across different methods. Our method consistently outperforms existing multimodal generation models~\citep{GPT-Image2, NanoBanana2, Qwen-Image-2, flux-2-2025, zimage} by a substantial margin across all evaluation metrics. Although these general-purpose models can produce visually appealing and high-fidelity images, they often fail to preserve spatially consistent scene geometry under specified camera configurations, resulting in noticeable deviations from the desired viewpoints. In contrast, Puffin-World achieves accurate and robust camera control while maintaining high visual quality and scene diversity. Its strong generalization across a wide range of scenes, viewpoints, and camera configurations highlights its practicality for real-world camera-controllable image generation beyond prior specialized models~\cite{bernal2025precisecam, puffin}. Additional qualitative results are presented in Figure~\ref{fig:cam_gen_sp} in the Appendix.

\subsection{3D World Modeling Results}

\begin{figure*}[!htbp]
    \centering

    \includegraphics[width=\linewidth]{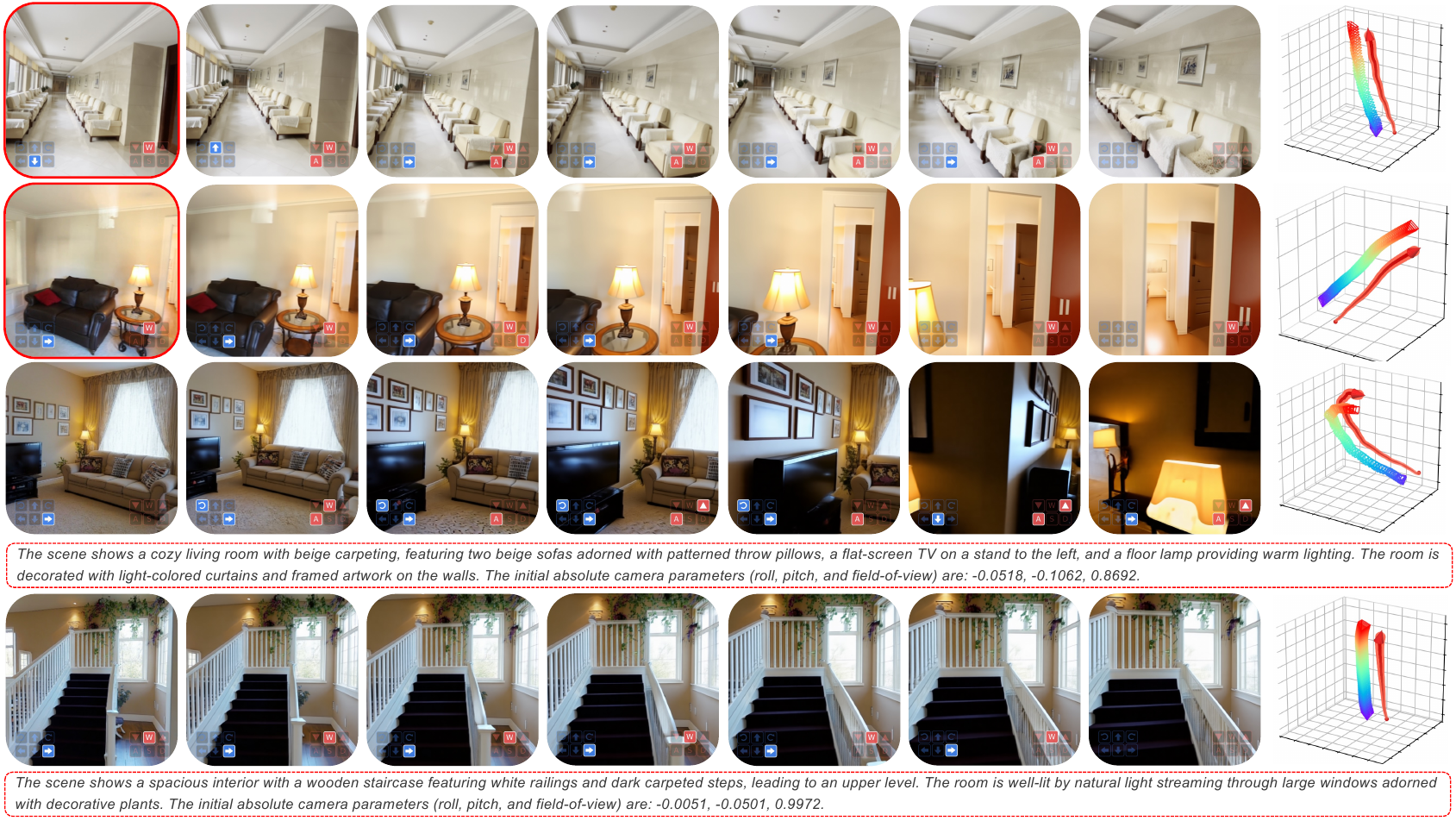}


    \includegraphics[width=\linewidth]{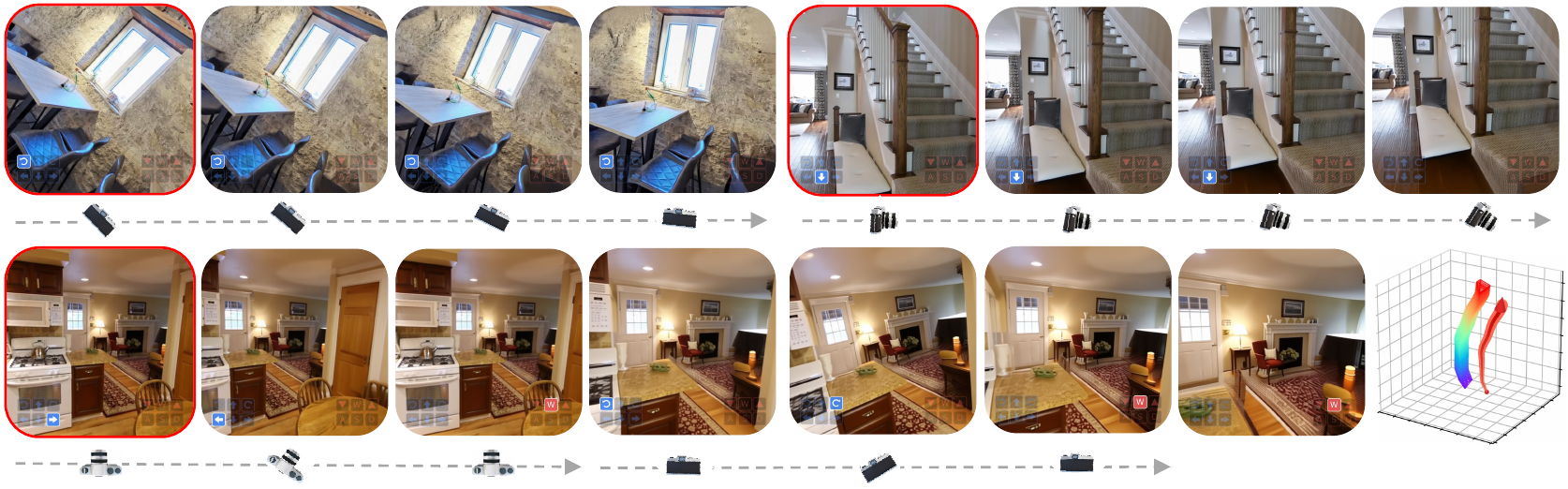}


    \includegraphics[width=\linewidth]{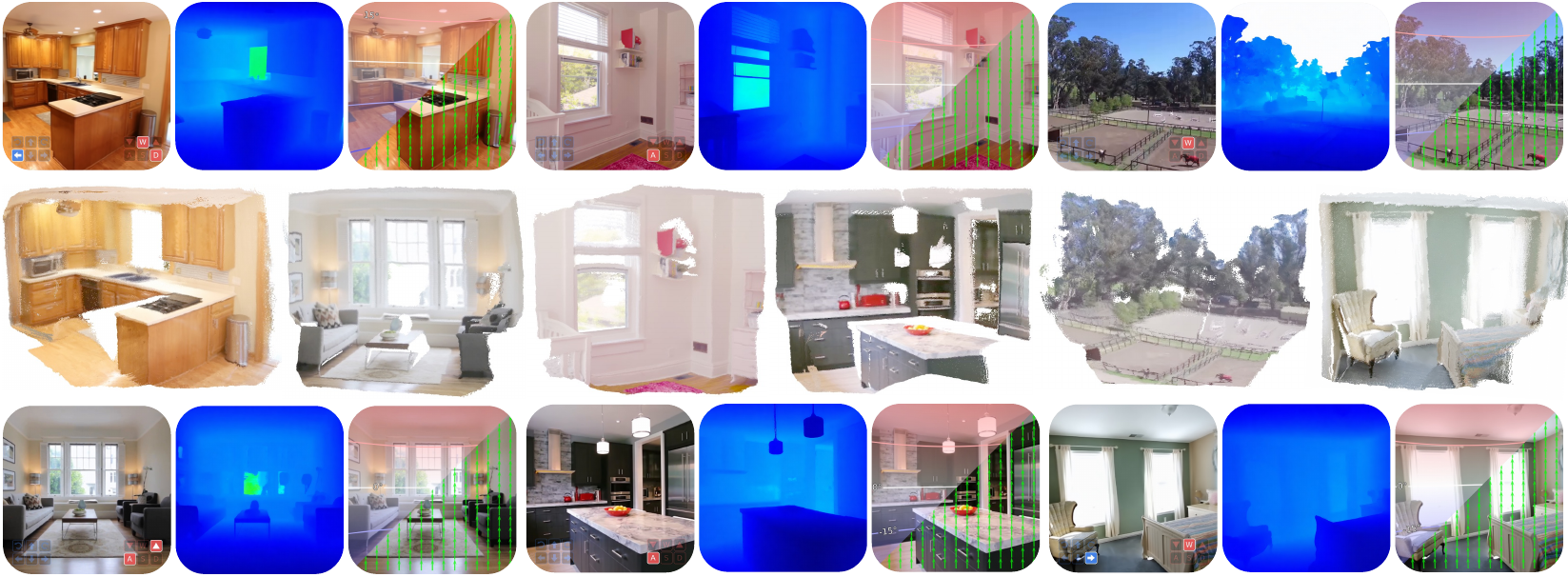}

    \caption{
    \textbf{Qualitative results of 3D world modeling.} Our Puffin-World supports various 3D modeling tasks: image-to-3D generation, text-to-3D generation, flexible rotation control, compound motion control, native 3D world states formulation, and 3D reconstruction. The red box marks the model's input, and others are generated.
    }
    \label{fig:3D_results}
\end{figure*}

As visualized in Figure~\ref{fig:3D_results}, Puffin-World demonstrates versatile capabilities for 3D world modeling, including image-to-3D generation, text-to-3D generation, flexible rotation control, compound motion control, native 3D world-state modeling, and 3D reconstruction. For text-to-3D generation, we first synthesize a reference view from the text prompt and initial absolute camera parameters through free-viewpoint spatial simulation, and then generate subsequent views conditioned on the reference image and target camera trajectory. For compound motion control, we sequentially combine translation trajectories from RealEstate10K~\citep{RealEstate10K} with rotational trajectories sampled from Puffin-Traj-1M, covering diverse roll, pitch, and yaw variations. For each predicted set of native 3D world states, we visualize the \emph{appearance} (image), \emph{geometry} (depth), and \emph{physics} (gravity field and latitude map), together with the resulting 3D reconstruction.

The results demonstrate that Puffin-World effectively handles diverse and challenging camera actions, including long trajectories, extreme rotations, and complex compound motions. The generated views faithfully preserve the semantic content of the reference image or text prompt while maintaining coherent spatial structure across viewpoints. Moreover, the jointly predicted world states exhibit physically and geometrically plausible distributions while preserving realistic visual appearance. The resulting 3D reconstructions further exhibit clear scene structure, coherent surface geometry, and strong cross-view consistency, precisely recovering the spatial layout and major geometric details of the generated world and demonstrating that the predicted multi-view observations can be consolidated into a consistent 3D representation. More results are presented in Figure~\ref{fig:rotation_traj_sp}, Figure~\ref{fig:3D_res_sp}, and Figure~\ref{fig:3D_res_recon_sp} in the Appendix.

\begin{table*}[t]
\centering
\footnotesize
\renewcommand{\arraystretch}{1.0}%
\setlength\tabcolsep{5.0pt}%

\caption{\textbf{Evaluation results on 3D world generation.}
We evaluate all methods on RealEstate10K and Puffin-Traj-Bench.
For Puffin-Traj-Bench, Roll, Pitch, and FoV report the median absolute errors of camera parameters re-estimated from generated frames, with Roll/Pitch AUC reported at $1^\circ/5^\circ/10^\circ$ thresholds.
Note that camera-control metrics are not reported on RealEstate10K due to the absence of corresponding absolute-camera ground truth.
$^\dagger$MVGenMaster's training data include samples from the evaluated RealEstate10K benchmark.
We color the \colorbox{tabfirst}{\textbf{best}} and
\colorbox{tabsecond}{second best} results within each benchmark.}

\begin{tabular}{clcccccccccccc}
\toprule

&
\multirow{2}{*}[-.4em]{\textbf{Approach}}
&
\multicolumn{3}{c}{\textbf{Visual Fidelity}}
&
\multicolumn{4}{c}{\textbf{Roll [degrees]}}
&
\multicolumn{4}{c}{\textbf{Pitch [degrees]}}
&
\textbf{FoV [degrees]}
\\

\cmidrule(lr){3-5}
\cmidrule(lr){6-9}
\cmidrule(lr){10-13}
\cmidrule(l){14-14}

&
&
PSNR$\uparrow$
&
SSIM$\uparrow$
&
LPIPS$\downarrow$
&
error$\downarrow$
&
\multicolumn{3}{c}{AUC\,$\triangleright$\,1/5/10$^\circ$\,$\uparrow$}
&
error$\downarrow$
&
\multicolumn{3}{c}{AUC\,$\triangleright$\,1/5/10$^\circ$\,$\uparrow$}
&
error$\downarrow$
\\

\midrule

\multirow{6}{*}[-0.15em]{%
    \rotatebox{90}{\textbf{RealEstate10K}}%
}
&
MotionCtrl~\cite{wang2024motionctrl}
& 13.38
& 0.482
& 0.507
& --
& --
& --
& --
& --
& --
& --
& --
& --
\\

&
CameraCtrl~\cite{he2024cameractrl}
& 16.93
& 0.589
& 0.352
& --
& --
& --
& --
& --
& --
& --
& --
& --
\\

&
ViewCrafter~\cite{ViewCrafter}
& 16.04
& 0.569
& \csecond 0.331
& --
& --
& --
& --
& --
& --
& --
& --
& --
\\

&
SEVA~\cite{SEVA}
& 14.84
& 0.529
& 0.421
& --
& --
& --
& --
& --
& --
& --
& --
& --
\\

&
MVGenMaster~\cite{cao2024mvgenmaster}$^\dagger$
& \csecond 17.11
& \csecond 0.591
& 0.348
& --
& --
& --
& --
& --
& --
& --
& --
& --
\\

&
\textbf{Puffin-World}
& \cfirst 17.22
& \cfirst 0.595
& \cfirst 0.318
& --
& --
& --
& --
& --
& --
& --
& --
& --
\\

\midrule

\multirow{6}{*}[-0.15em]{%
    \rotatebox{90}{%
        \scalebox{0.88}{\textbf{Puffin-Traj-Bench}}%
    }%
}
&
MotionCtrl~\cite{wang2024motionctrl}
& 12.53
& 0.438
& 0.604
& \03.20
& \00.25
& \00.43
& \00.54
& \02.15
& \00.28
& \00.49
& \00.59
& 10.04
\\

&
CameraCtrl~\cite{he2024cameractrl}
& 14.28
& 0.512
& 0.532
& 11.79
& \00.05
& \00.11
& \00.21
& 10.81
& \00.05
& \00.16
& \00.26
& 45.84
\\

&
ViewCrafter~\cite{ViewCrafter}
& 14.52
& 0.525
& 0.510
& \01.92
& \00.32
& \00.53
& \00.66
& \02.89
& \00.23
& \00.43
& \00.58
& \csecond \06.43
\\

&
SEVA~\cite{SEVA}
& \csecond 17.94
& \cfirst 0.614
& \csecond 0.307
& \csecond \01.35
& \csecond \00.38
& \csecond \00.63
& \csecond \00.75
& \csecond \01.76
& \csecond \00.29
& \csecond \00.56
& \csecond \00.71
& 12.59
\\

&
MVGenMaster~\cite{cao2024mvgenmaster}
& 13.64
& 0.496
& 0.589
& \09.24
& \00.06
& \00.14
& \00.28
& \09.43
& \00.08
& \00.17
& \00.30
& 26.98
\\

&
\textbf{Puffin-World}
& \cfirst 18.00
& \csecond 0.613
& \cfirst 0.288
& \cfirst \00.80
& \cfirst \00.57
& \cfirst \00.79
& \cfirst \00.88
& \cfirst \01.10
& \cfirst \00.45
& \cfirst \00.72
& \cfirst \00.83
& \cfirst \02.96
\\

\bottomrule
\end{tabular}

\label{tab:3D_gen_eva}
\end{table*}

We then quantitatively evaluate Puffin-World on RealEstate10K (50 held-out clips with no overlap with the training data) and Puffin-Traj-Bench (100 clips) following prior work~\cite{cao2024mvgenmaster}, with comparisons against MotionCtrl~\cite{wang2024motionctrl}, CameraCtrl~\cite{he2024cameractrl}, ViewCrafter~\cite{ViewCrafter}, SEVA~\cite{SEVA}, and MVGenMaster~\cite{cao2024mvgenmaster}. Visual fidelity is measured using PSNR, SSIM, and LPIPS on both benchmarks. Since RealEstate10K does not provide absolute-camera ground truth, camera-control accuracy is evaluated only on Puffin-Traj-Bench by re-estimating roll, pitch, and vertical FoV from the generated frames using the same Puffin-World camera-understanding branch for all methods, reporting median angular error and AUC@$1/5/10^\circ$. As listed in Table~\ref{tab:3D_gen_eva}, Puffin-World achieves the best PSNR, SSIM, and LPIPS on RealEstate10K. On the more challenging Puffin-Traj-Bench, which contains substantially larger rotations and intrinsic variations, Puffin-World achieves the best PSNR and LPIPS with competitive SSIM, while substantially outperforming all baselines in camera-control accuracy across roll, pitch, and FoV. These results demonstrate that Puffin-World maintains strong perceptual fidelity on conventional real-world trajectories while providing markedly more accurate and robust camera control under challenging spatial transformations.

\subsection{Ablation Study}
\label{sec:ablation_study}
\begin{table}[t]
\centering
\footnotesize
\renewcommand{\arraystretch}{1.0}%
\setlength\tabcolsep{2.9pt}%
\caption{\textbf{Ablation study on the proposed physics propagation (PP) for 3D world generation with challenging camera motions.}}
\begin{tabular}{llccccc}
\toprule
\multirow{2}{*}[-.4em]{\textbf{Approach}} &\multirow{2}{*}[-.4em]{\textbf{Motion}} 
& \multicolumn{3}{c}{\textbf{Image Metrics}} 
& \multicolumn{2}{c}{\textbf{Camera Metrics}} \\
\cmidrule(lr){3-5}
\cmidrule(lr){6-7}
&& PSNR\,$\uparrow$ & SSIM\,$\uparrow$ & LPIPS\,$\downarrow$
& Error\_R\,$\downarrow$ & Error\_P\,$\downarrow$ \\
\midrule
\multirow{4}{*}\textbf{Baseline}
&Roll     &          22.97 &          \00.69 &          \00.12 &          \02.35 &          \06.73           \\
&Pitch              &          18.53 &          \00.60 &          \00.26 &          \02.17 &  \05.60 \\
&Yaw                       &  18.12 & \00.59 &  \00.30 &   \01.31 &          \03.10 \\
&Average                 &          \csecond 19.87 &          \csecond \00.62 &          \csecond \00.23&          \csecond \01.94 &          \csecond \05.14 \\
\midrule
\multirow{4}{*}\textbf{Baseline \textit{w}/ PP}
&Roll      &          24.02 &           \00.74 &          \00.10 &          \02.04 &          \06.31\\
&Pitch              &          19.18 &          \00.65 &          \00.24 &          \02.03 &          \05.28\\
&Yaw                       &          18.42 &          \00.62 &          \00.30 & \01.22 & \02.67\\
&Average                  &          \cfirst \ 20.54 &          \cfirst \00.67 &         \cfirst \00.21 &          \cfirst \01.76 &          \cfirst \04.75 \\
\bottomrule
\end{tabular}
\label{tab:ablation_physical_prop}
\end{table}

We isolate the contribution of physics propagation, which anchors a generated trajectory to a consistent gravity-aligned frame (Sec.~\ref{sec:world_modeling}), and focus on rotation trajectories because they provide the most sensitive probe of absolute grounding: unlike translation, pure rotation directly changes the camera orientation with respect to gravity and the horizon, making inconsistencies in the physical frame immediately observable. We use Puffin-Traj-Bench (spanning roll, pitch, and yaw rotations), generate the corresponding views with and without physics propagation, and recover the camera parameters of each generated frame using Puffin-World's physics-perception capability. The calibrated roll and pitch errors (Error\_R and Error\_P) are compared against the GT cameras and reported in Table~\ref{tab:ablation_physical_prop}, together with PSNR, SSIM, and LPIPS against the target views. Physics propagation consistently improves both visual fidelity and physical world grounding, with the largest gains on roll and pitch trajectories that are directly coupled to gravity. Yaw motion shows smaller appearance improvements because it rotates around the gravity axis, although physics propagation still suppresses spurious horizon tilt and improves physical consistency. These results support our design rationale: propagating the absolute physical state from the reference view through the known relative motion maintains a coherent gravity-anchored representation across the trajectory, leading to more stable horizons, better scene uprightness, improved perceptual quality, and camera states that better follow the intended motion. Qualitative comparisons are provided in Figure~\ref{fig:ablation_pp_vis} of the Appendix.

\begin{figure*}[htbp]
    \centering

    \includegraphics[width=\linewidth]{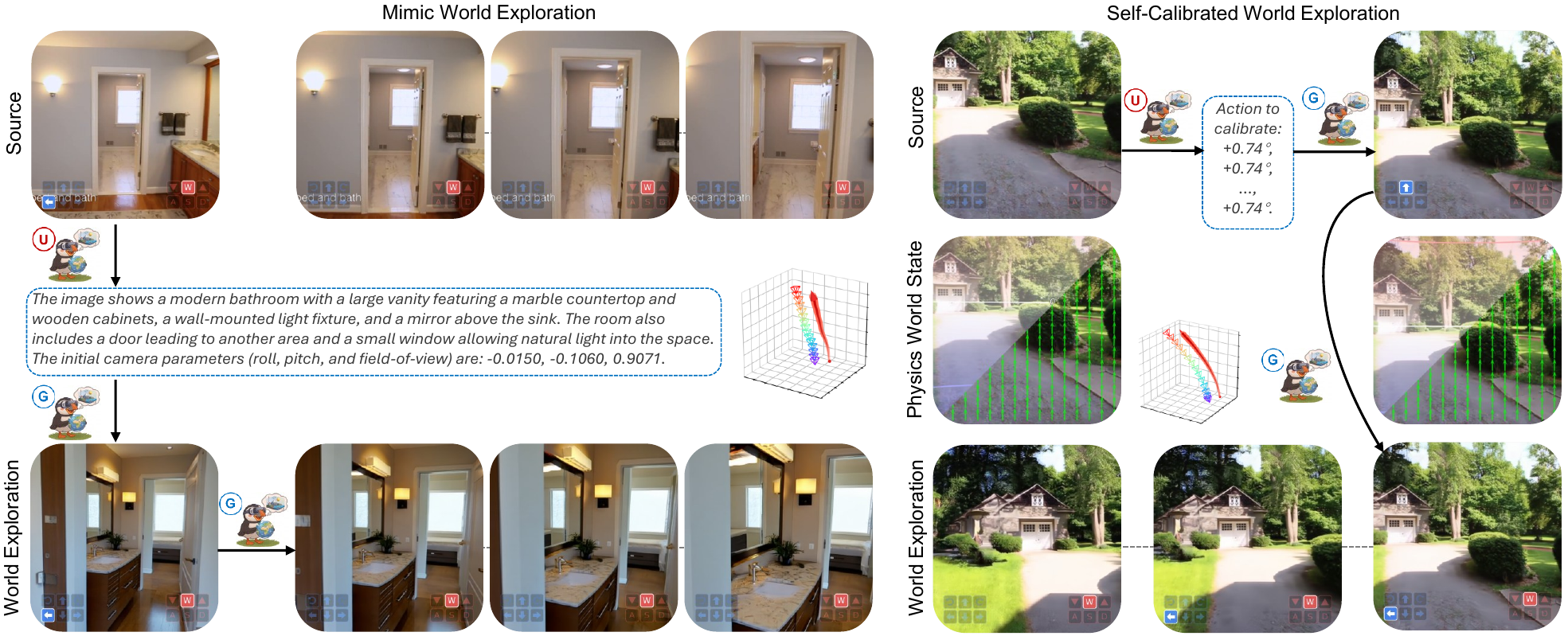}

    \caption{
    \textbf{Closed-loop applications of Puffin-World:} Mimic world exploration (left) and self-calibrated world exploration, both of which require multi-task synergy involving understanding and generation. The type of each sub-task is marked with circled U (understanding) and G (generation).
    }
    \label{fig:applications}
\end{figure*}

\subsection{Applications}
\label{sec:application}
As mentioned in Section~\ref{sec:multi_task_synergy}, Puffin-World enables multi-task synergy for complex, closed-loop applications that jointly involve perception, reasoning, and generation. Two representative closed-loop capabilities further illustrate the benefits of this unified formulation as shown in Figure~\ref{fig:applications}. In \textit{mimic world exploration}, Puffin-World reproduces a target exploration process by expanding a 3D world from a common starting view under a prescribed camera trajectory. In \textit{self-calibrated world exploration}, the model instead reasons about its current physical state, predicts corrective camera actions, and imagines the resulting observations to progressively resolve gravity misalignment, forming an interaction loop reminiscent of World-Action Models (WAMs). Both behaviors emerge from the same multimodal model, without auxiliary perception, calibration, or generation modules, demonstrating the potential of Puffin-World as a general foundation for interactive 3D environments, virtual reality, and embodied agents.
\section{Conclusion}
We present Puffin-World, a unified multimodal model that perceives, simulates, and generates the physical world through three native 3D world states: physics, geometry, and appearance. In particular, the Omni-Camera representation integrates gravity-aware absolute orientation with ray-based relative motion into a unified camera condition, supporting single-view control, multi-view synthesis, and challenging camera trajectories. Building on this representation, physics propagation transfers the natively perceived absolute state of the reference view across future frames, maintaining a coherent physical frame for stable and self-calibrated world generation. Within a single framework, Puffin-World jointly supports physical-world perception, free-viewpoint spatial simulation, and 3D world generation and reconstruction, while Puffin-16M provides large-scale vision--language--camera supervision and diverse challenging motions for scaling these capabilities. Extensive experiments demonstrate strong and consistent performance across these tasks and further highlight the benefits of multi-task synergy for closed-loop applications.

We believe that grounding multimodal tasks in a shared, physically anchored world representation provides a promising path toward versatile spatial intelligence and physical AI. Future directions include extending the framework to dynamic scenes and longer horizons, and modeling richer physical states beyond gravity and latitude.

\section*{Acknowledgment}
This research is supported by cash and in-kind funding from NTU S-Lab and industry partner(s). It is also supported by Singapore MOE AcRF Tier 2 (MOE-T2EP20224-0003) and NRF Investigatorship (NRF-NRFI11-2026-0008).

{
    \small
    \bibliographystyle{ieeenat_fullname}
    \bibliography{main}
}

\newpage
\appendix
\section{Appendix}

\begin{figure*}[htbp]
    \centering
    \includegraphics[width=\linewidth]{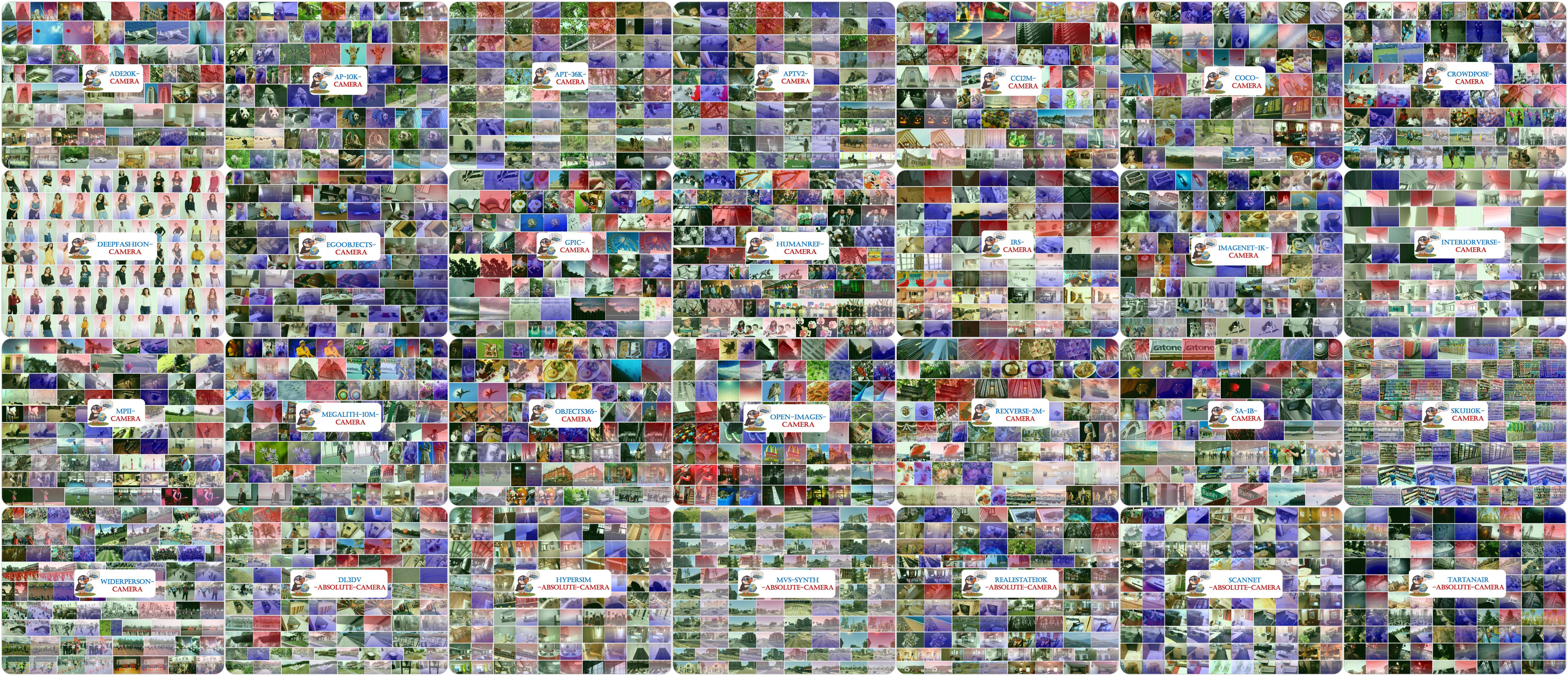}
    \caption{
    \textbf{The overview of the public datasets annotated by Puffin-World, with a total of 28 datasets covering approximately 44.5 million images across diverse scene distributions.}
    }
    \label{fig:camera_dataset}
\end{figure*}

\begin{figure*}[htbp]
    \centering
    \includegraphics[width=\linewidth]{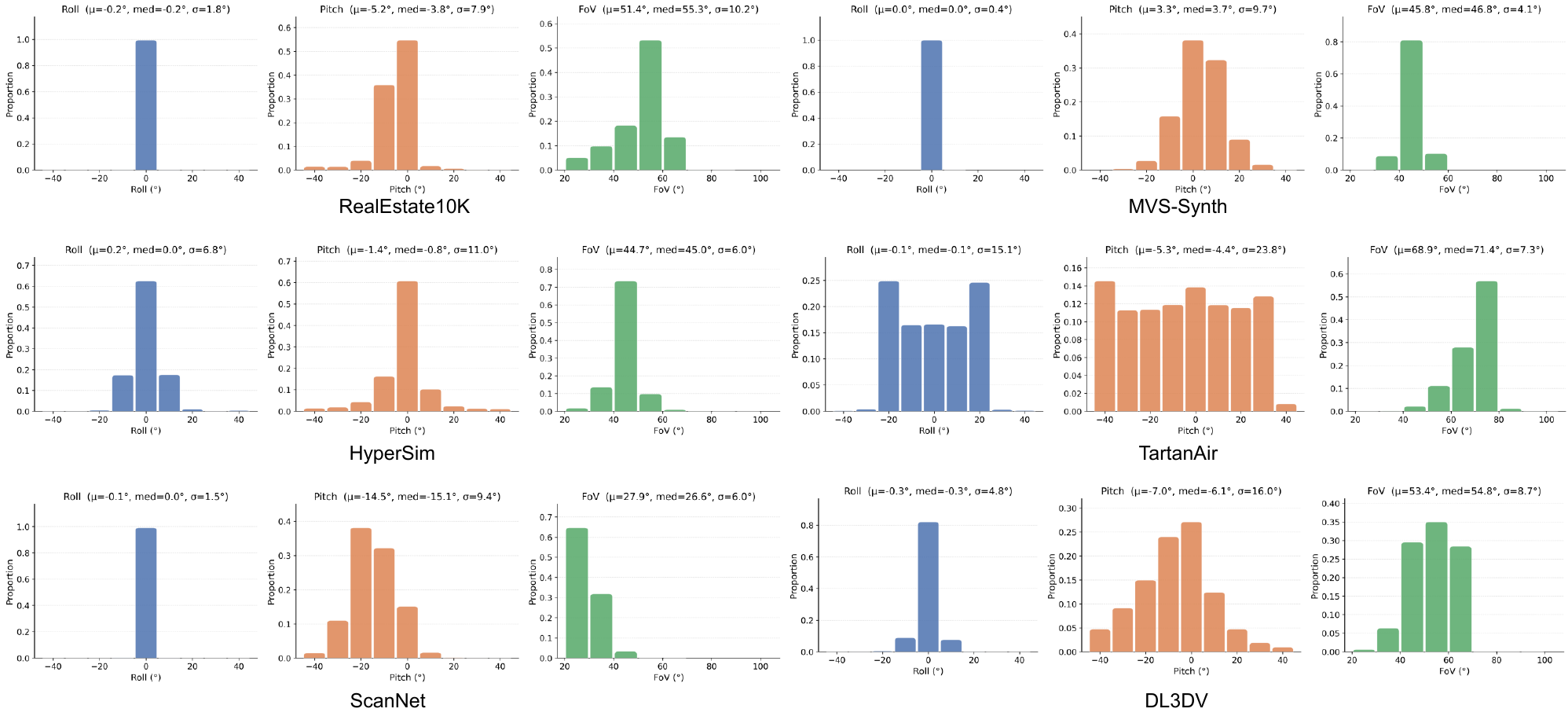}
    \caption{
    \textbf{The distributions of the absolute camera parameters (roll and pitch) and FoV of existing public 3D or sequential datasets}~\cite{ling2024dl3dv, RealEstate10K, dai2017scannet, MVS-Synth, Tartanair, roberts2021hypersim}.
    }
    \label{fig:dataset_cam_dis}
\end{figure*}


\begin{table*}[t]
  \centering
  \caption{\textbf{Camera parameter statistics of 28 annotated datasets using Puffin-World.} For every source dataset we report its original
  task, the number of validly annotated images and scenes, and the mean/median/standard
  deviation of the predicted roll, pitch, and vertical FoV (in degrees).  $^\dagger$computed on a ${\sim}4.9$M-image sample;
  $^\ddagger$pooled from the released train+val statistics.}
  \label{tab:camera_datasets}
  \footnotesize
  \setlength{\tabcolsep}{5pt}
  \renewcommand{\arraystretch}{1.12}
  \arrayrulecolor{rulegray}
  \resizebox{0.96\textwidth}{!}{%
  \begin{tabular}{@{}llrrrrr@{}}
    \toprule[1.1pt]
    & & & & \multicolumn{3}{c}{\textbf{Camera Parameters}
      ($\mu$/med/$\sigma$, $^\circ$)} \\
    \cmidrule(l){5-7}
    \textbf{Dataset} & \textbf{Original task} & \textbf{\#Images} &
    \textbf{\#Scenes} & \textbf{Roll} & \textbf{Pitch} & \textbf{FoV} \\
    \midrule[0.65pt]
    \multicolumn{7}{@{}l}{\emph{Single-image datasets}} \\
    \cmidrule(r){1-7}
    ADE20K \cite{zhou2019semantic}        & Semantic segmentation                & 27.6K  & --  & $0.0/0.0/1.9$    & $-1.3/{-1.0}/8.0$    & $32.7/31.2/11.5$ \\
    \rowcolor{rowblue}
    AP-10K \cite{yu2021ap}        & Animal pose estimation               & 10.0K  & --  & $0.4/0.2/8.1$    & $-9.2/{-6.1}/13.9$   & $26.5/23.2/8.3$  \\
    APT-36K \cite{yang2022apt}      & Animal pose tracking                 & 35.7K  & --  & $0.2/0.0/3.5$    & $-4.9/{-3.7}/11.3$   & $24.1/20.3/6.0$  \\
    \rowcolor{rowblue}
    APTv2 \cite{yang2023aptv2}        & Animal pose tracking                 & 41.3K  & --  & $0.2/0.0/3.5$    & $-5.1/{-3.7}/11.2$   & $24.2/22.9/6.0$  \\
    CC12M \cite{changpinyo2021conceptual}        & Vision--language pre-training        & 10.90M & --  & $0.2/0.0/5.7$    & $-2.8/{-0.3}/11.8$   & $32.3/29.0/10.6$ \\
    \rowcolor{rowblue}
    COCO  \cite{lin2014microsoft}        & Detection \& segmentation            & 122K   & --  & $0.0/0.0/6.2$    & $-7.3/{-4.0}/15.0$   & $34.9/32.5/11.7$ \\
    CrowdPose \cite{li2019crowdpose}    & Human pose estimation                & 20.0K  & --  & $0.1/0.0/3.5$    & $-5.1/{-3.7}/8.5$    & $31.6/28.7/10.7$ \\
    \rowcolor{rowblue}
    DeepFashion \cite{liu2016deepfashion}   & Fashion recognition \& retrieval     & 13.7K  & --  & $0.1/0.3/3.5$    & $1.1/0.4/3.6$      & $32.6/32.1/3.4$  \\
    EgoObjects \cite{zhu2023egoobjects}    & Egocentric object detection          & 241.6K & --  & $-0.1/0.0/8.0$   & $-24.5/{-25.6}/15.3$ & $39.3/36.7/10.2$ \\
    \rowcolor{rowblue}
    GPIC$^\dagger$ \cite{chandrasegaran2026gpic} & Image generation corpus             & 4.9M   & --  & $0.1/0.0/5.6$    & $-2.6/{-1.3}/13.0$   & $33.8/31.2/11.6$ \\
    HumanRef \cite{jiang2025referring}      & Referring expression comprehension   & 45.3K  & --  & $0.1/0.0/3.6$    & $-2.5/{-1.5}/8.4$    & $32.5/31.2/9.9$  \\
    \rowcolor{rowblue}
    IRS \cite{wang2021irs}           & Stereo matching / depth (syn.)       & 188.3K & --  & $0.2/0.0/4.9$    & $-8.3/{-5.0}/14.8$   & $54.8/56.7/6.1$  \\
    ImageNet-1K$^\ddagger$ \cite{deng2009imagenet} & Image classification        & 1.33M  & --  & $0.1/0.0/7.8$    & $-7.8/{-3.7}/15.7$   & $29.0/25.6/8.9$  \\
    \rowcolor{rowblue}
    InteriorVerse \cite{li2018interiornet} & Inverse rendering (syn.\ indoor)     & 61.2K  & --  & $0.8/0.0/6.3$    & $-1.3/{-0.8}/16.9$   & $71.6/68.6/14.9$ \\
    MPII \cite{andriluka14cvpr}          & Human pose estimation                & 25.0K  & --  & $0.0/0.0/2.7$    & $-5.0/{-3.8}/7.8$    & $27.6/26.6/6.2$  \\
    \rowcolor{rowblue}
    Megalith-10M \cite{bohan2024megalith} & Image generation corpus              & 9.38M  & --  & $0.0/0.0/6.7$    & $-3.1/{-2.3}/12.1$   & $34.6/31.3/12.0$ \\
    Objects365 \cite{shao2019objects365}    & Object detection                     & 1.72M  & --  & $0.1/0.0/4.0$    & $-4.4/{-2.7}/11.2$   & $33.4/31.3/11.0$ \\
    \rowcolor{rowblue}
    Open Images \cite{kuznetsova2020open}   & Detection \& multi-label cls.        & 8.32M  & --  & $0.1/0.0/6.2$    & $-3.1/{-1.1}/14.8$   & $31.8/31.2/8.2$  \\
    Rexverse-2M \cite{jiang2024chatrex}   & Multimodal grounding \& captioning   & 414.6K & --  & $0.2/0.0/6.8$    & $-4.0/{-0.9}/12.9$   & $32.2/28.9/11.1$ \\
    \rowcolor{rowblue}
    SA-1B \cite{kirillov2023segment}         & Promptable segmentation              & 897.6K & --  & $0.2/0.0/4.9$    & $2.0/1.5/13.1$     & $36.3/34.4/13.2$ \\
    SKU110K \cite{goldman2019precise}       & Dense object detection               & 19.1K  & --  & $-0.1/0.0/1.9$   & $-13.9/{-14.1}/6.6$  & $49.6/51.9/12.5$ \\
    \rowcolor{rowblue}
    WiderPerson \cite{zhang2019widerperson}   & Pedestrian detection                 & 13.3K  & --  & $0.1/0.0/1.4$    & $-1.8/{-1.4}/5.8$    & $29.1/27.2/8.3$  \\
    \midrule[0.55pt]
    \multicolumn{7}{@{}l}{\emph{Multi-view datasets (absolute-pose annotations)}} \\
    \cmidrule(r){1-7}
    DL3DV~\cite{ling2024dl3dv}         & Novel view synthesis / 3D recon.     & 2.16M  & 6,377 & $-0.3/{-0.3}/4.9$  & $-7.0/{-6.1}/16.0$   & $53.4/54.8/8.7$  \\
    \rowcolor{rowblue}
    HyperSim~\cite{roberts2021hypersim}      & 3D scene understanding (syn.)        & 73.6K  & 448   & $0.2/0.0/6.8$    & $-1.4/{-0.8}/11.0$   & $44.7/45.0/6.0$  \\
    MVS-Synth~\cite{MVS-Synth}     & Multi-view stereo (syn.)             & 12.0K  & 120   & $0.0/0.0/0.4$    & $3.3/3.7/9.7$      & $45.8/46.8/4.1$  \\
    \rowcolor{rowblue}
    RealEstate10K~\cite{RealEstate10K} & Novel view synthesis / camera traj.  & 595.4K & 5,134 & $-0.2/{-0.2}/1.8$  & $-5.2/{-3.8}/7.9$    & $51.4/55.3/10.2$ \\
    ScanNet~\cite{dai2017scannet}       & RGB-D reconstruction         & 2.32M  & 1,468 & $-0.1/0.0/1.5$   & $-14.5/{-15.1}/9.4$  & $27.8/26.6/6.0$  \\
    \rowcolor{rowblue}
    TartanAir~\cite{Tartanair}     & SLAM / visual odometry (syn.)        & 306.6K & 18    & $-0.1/{-0.1}/15.1$ & $-5.3/{-4.4}/23.8$   & $68.9/71.4/7.3$  \\
    \bottomrule[1.1pt]
  \end{tabular}}
\end{table*}

This appendix provides additional details and qualitative results to complement the main paper. We first elaborate on the datasets used and released in this work, including camera annotations for public datasets, additional physical labels in Puffin-16M, and their camera parameter statistics. We then provide further training details, including the construction of dense geometric supervision for real-world datasets. Finally, we present extensive qualitative results for camera-to-world understanding, camera-controllable generation, and 3D world modeling, covering diverse scenes, challenging camera configurations, long trajectories, compound motions, native world-state prediction, and 3D reconstruction.

\subsection{Dataset Details}

\subsubsection{Annotating Public Datasets}
Publicly used datasets~\citep{ling2024dl3dv, chandrasegaran2026gpic, dai2017scannet} show versatile abilities to various tasks; however, most of them lack accurate absolute camera parameters anchored to the real world. To this end, we annotate 28 datasets using Puffin-World's powerful world perception capability, of which the understanding branch treats camera properties as a language modality and predicts three parameters from each image: roll, pitch, and vertical field-of-view (FoV). The overview of the annotated datasets is shown in Figure~\ref{fig:camera_dataset}. As listed in Tab.~\ref{tab:camera_datasets}, the collection comprises 22 single-image datasets spanning recognition, detection, segmentation, pose estimation, and generative pre-training, together with six sequential/3D \emph{Absolute-Camera} datasets. Overall, the release contains approximately $44.5$M camera annotations. Across these datasets, several consistent biases emerge: roll is strongly concentrated around $0^\circ$, indicating a dominant level-camera bias; pitch is generally negative, reflecting a tendency toward downward-looking viewpoints; and FoV varies substantially with capture regime, with web photographs typically exhibiting narrower views than synthetic, scanned, or video-based datasets. Together, these statistics reveal that conventional vision corpora substantially under-sample strongly tilted, rolled, and wide-angle viewpoints.

Several datasets exhibit distinctive camera distributions. For example, \emph{EgoObjects} shows the strongest downward-looking bias, consistent with head-mounted capture of hands and nearby objects, while \emph{DeepFashion} exhibits exceptionally low variation in roll, pitch, and FoV, reflecting tightly controlled studio photography. In contrast, \emph{TartanAir} presents the largest rotational diversity, making it particularly valuable for camera orientations rarely observed in natural photographs; \emph{ScanNet} and \emph{SKU110K} also show pronounced downward pitch due to room-scanning and shelf-facing capture, respectively. These annotations transform widely used vision datasets into camera-grounded resources that support viewpoint-bias analysis, camera-conditioned generation, camera-diverse data curation, and large-scale monocular physics perception. For the \emph{Absolute-Camera} subsets, as analyzed in Figure~\ref{fig:dataset_cam_dis}, aligning per-frame predictions with camera trajectories further enables consistency analysis and gravity-grounded world modeling.

\begin{figure*}[htbp]
    \centering

    \includegraphics[width=\linewidth]{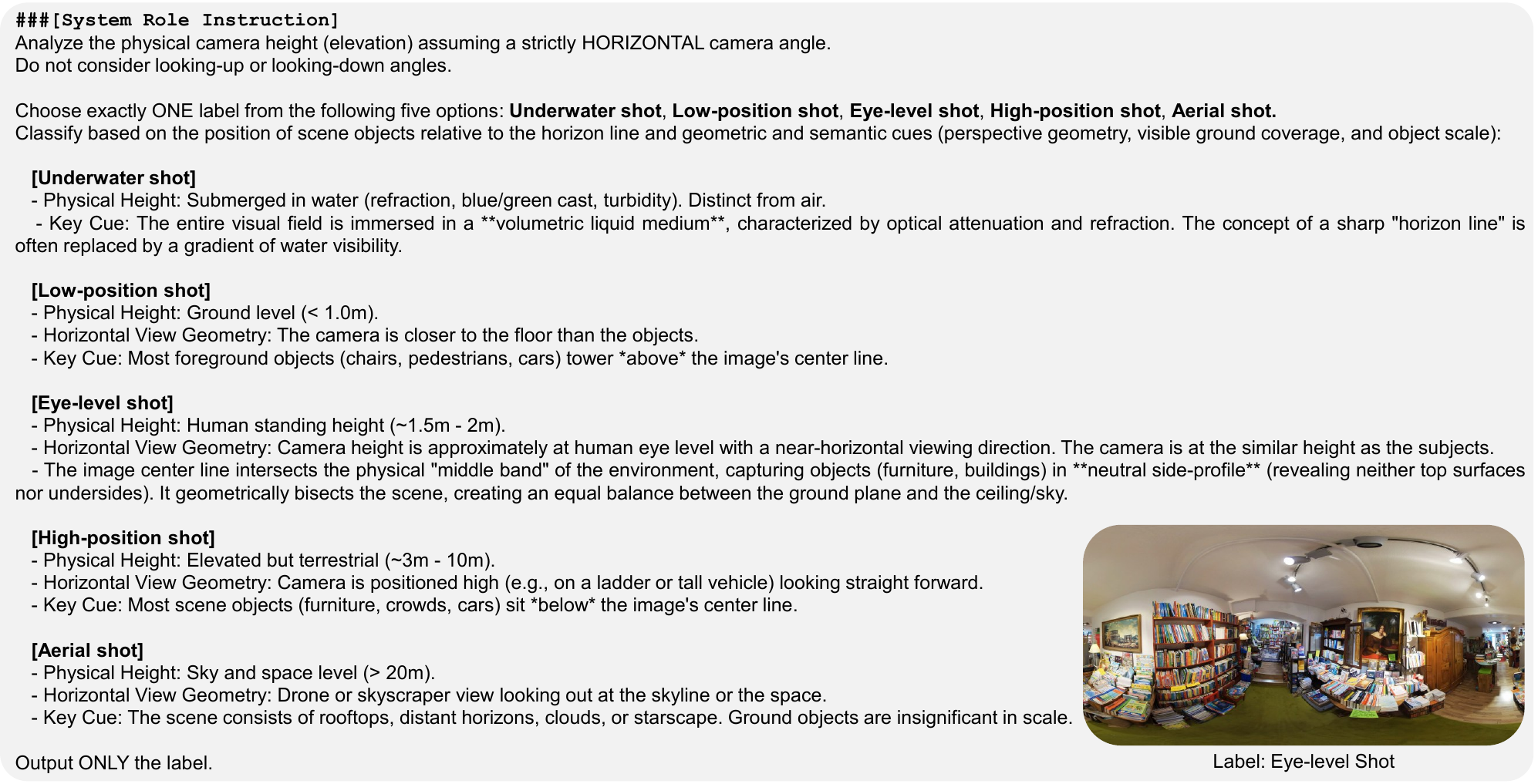}

    \caption{
    \textbf{The designed prompt for captioning the camera height for the panoramic images.} We categorize it into five levels: underwater shot, low-position shot, eye-level shot, high-position shot, and aerial shot. A representative example, consisting of the input panoramic image and the corresponding output label, is shown in the bottom-right corner.
    }
    \label{fig:camera_height_caption}
\end{figure*}

\begin{figure*}[htbp]
    \centering

    \includegraphics[width=\linewidth]{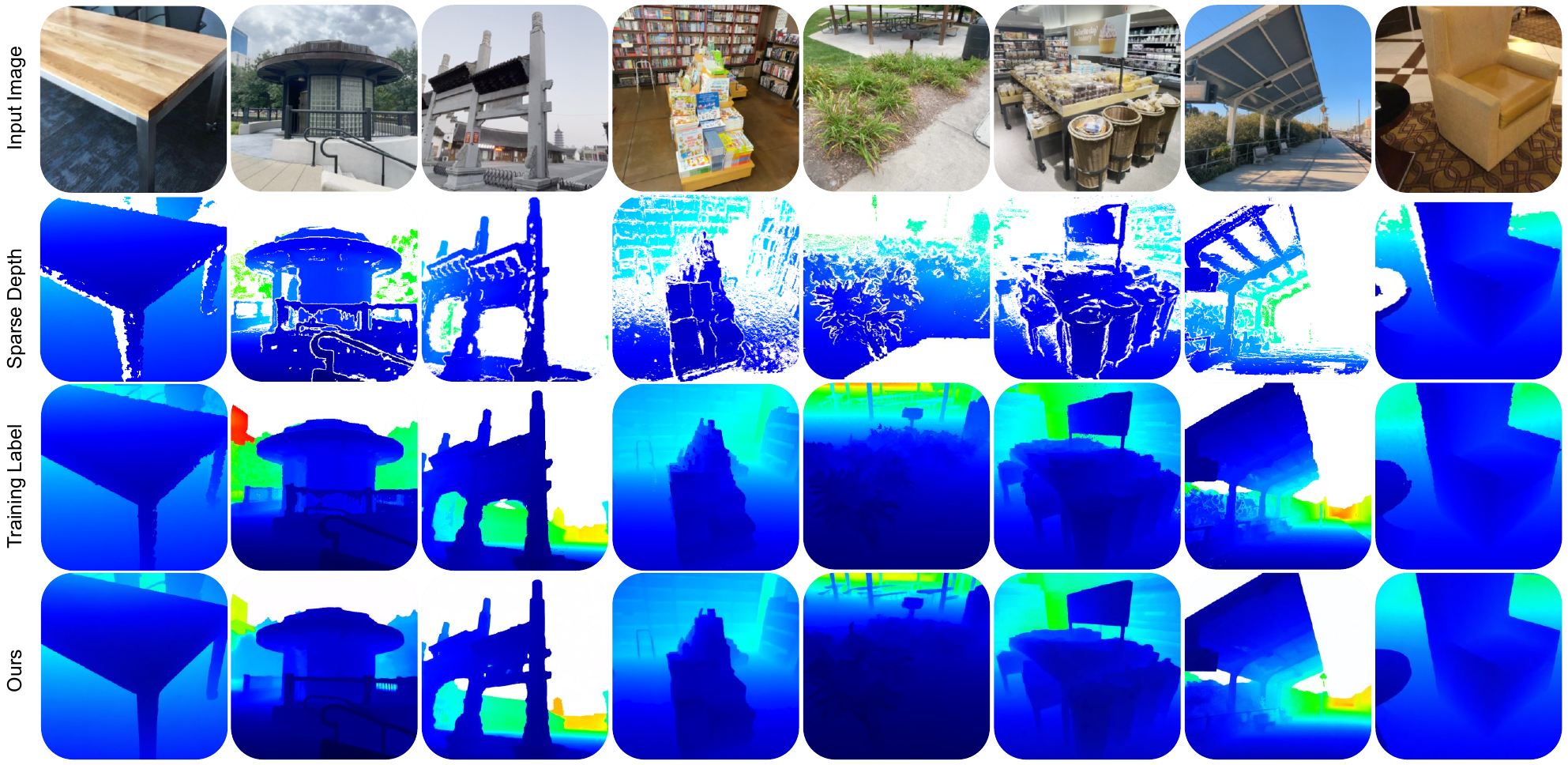}

    \caption{
    \textbf{Depth map visualization on the DL3DV dataset~\cite{ling2024dl3dv}.} From top to bottom, we show the input image, the sparse label provided in the original dataset, our annotated and aligned dense label (training used), and the predicted result by Puffin-World.
    }
    \label{fig:depth_cp}
\end{figure*}

\subsubsection{Camera Height of Puffin-16M}
Beyond the precise camera parameters and camera trajectories provided in Puffin-16M, we further annotate the camera height of each original panoramic image, prior to perspective rendering, using Qwen3-VL-32B~\citep{bai2025qwen3}. We categorize camera height into five levels---underwater shot, low-position shot, eye-level shot, high-position shot, and aerial shot---to provide a coarse yet physically meaningful description of the observer's vertical position in the scene. The resulting height label is subsequently inherited by all perspective views rendered from the corresponding panorama, thereby complementing their precise camera parameters with an additional high-level physical cue. The detailed annotation prompts are provided in Figure~\ref{fig:camera_height_caption}. We expect these complementary camera annotations to facilitate richer spatial reasoning, viewpoint-aware generation, and future studies of camera-grounded physical world understanding and generation.

\begin{table*}[t]
\centering
\footnotesize
\setlength{\tabcolsep}{7pt}
\renewcommand{\arraystretch}{1.18}
\arrayrulecolor{rulegray}

\caption{\textbf{Training datasets for our model.}
We summarize the source category, number of scenes and frames, depth availability,
temporal sampling interval, test-set usage, and the corresponding training stages
for each dataset.}
\label{tab:3d_training_data}

\resizebox{\textwidth}{!}{%
\begin{tabular}{lcccccc}
\toprule[1.1pt]

\textbf{Dataset}
&
\textbf{Category}
&
\textbf{Scenes / Frames}
&
\textbf{Depth}
&
\textbf{Sample Interval}
&
\textbf{Testset Split}
&
\textbf{Training Stage}
\\

\midrule[0.65pt]

DL3DV~\cite{ling2024dl3dv}
&
Public
&
6377 / 2M
&
Yes$^\ast$
&
$[1,8]$
&
No
&
III, IV
\\

\rowcolor{rowblue}
RealEstate10K~\cite{RealEstate10K}
&
Public
&
5134 / 600K
&
No
&
$[48,144]$
&
Yes
&
III, IV
\\

HyperSim~\cite{roberts2021hypersim}
&
Public
&
756 / 70K
&
Yes
&
$[1,1]$
&
No
&
IV
\\

\rowcolor{rowblue}
MVS-Synth~\cite{MVS-Synth}
&
{Public}
&
120 / 12K
&
Yes
&
$[1,2]$
&
No
&
IV
\\

TartanAir~\cite{Tartanair}
&
\makecell{Public}
&
369 / 300K
&
Yes
&
$[1,10]$
&
No
&
IV
\\

\rowcolor{rowblue}
ScanNet~\cite{dai2017scannet}
&
{Public}
&
1468 / 2M
&
{Yes$^\ast$}
&
$[1,8]$
&
No
&
IV
\\

\midrule[0.55pt]

Puffin-Traj-1M
&
Self-Constructed
&
1M / 100M
&
No
&
$[1,12]$
&
Yes
&
III, IV
\\

\rowcolor{rowblue}
Puffin-Cam-15M
&
Self-Constructed
&
-- / 15M
&
No
&
--
&
Yes
&
I, II, III, IV
\\

\bottomrule[1.1pt]
\end{tabular}%
}

\vspace{1pt}
\begin{flushleft}
\scriptsize
$^\ast$ We refined the sparse depth label into the dense one.
\end{flushleft}

\end{table*}
\subsection{Training Details}
Table~\ref{tab:3d_training_data} summarizes the datasets used for Puffin-World across different training stages. We combine large-scale public sequential or 3D datasets with our self-constructed Puffin-16M (Puffin-Cam-15M and Puffin-Traj-1M) to provide complementary supervision in native 3D world states. The public datasets cover diverse real-world and synthetic environments with varying trajectory lengths, motion patterns, and depth availability, thereby providing broad spatial and geometric supervision. When depth annotations are available, they are incorporated to strengthen geometry learning, while datasets without depth primarily contribute to multi-view appearance consistency and camera-conditioned generation. Puffin-Traj-1M complements these sources with substantially richer and more challenging rotational trajectories, extending the motion distribution beyond that of conventional capture datasets. Puffin-Cam-15M further supplies large-scale language-image-camera triplets and is reused across multiple stages to preserve strong camera perception and generation capabilities. Together, these datasets form a balanced training mixture that progressively supports single-view understanding/generation, cross-view generation, trajectory modeling, and 3D world reconstruction within the same framework.

Although ScanNet~\cite{dai2017scannet} and DL3DV~\cite{ling2024dl3dv} provide camera poses, their depth annotations remain incomplete due to sensor holes in ScanNet and sparse MVS reconstruction~\cite{schonberger2016colmap} in DL3DV. As described in Section~\ref{sec:datasets_add_captions}, we use DA3 to predict dense depth for each frame and align it with the available measurements, preserving reliable metric scale while filling missing regions. Figure~\ref{fig:depth_cp} compares the sparse depth provided by the original datasets, our aligned dense annotations used for training, and the depth predicted by Puffin-World. Compared with the original sparse measurements, our annotations provide substantially denser and more complete geometric supervision, reducing holes and missing structures and thereby facilitating unified RGB--depth training without specialized tokenizer designs. Moreover, the depth predicted by Puffin-World exhibits sharper object boundaries and fewer artifacts, while maintaining a spatial layout that is highly consistent with the corresponding input image.

\begin{figure}[!htbp]
    \centering

    \includegraphics[width=\linewidth]{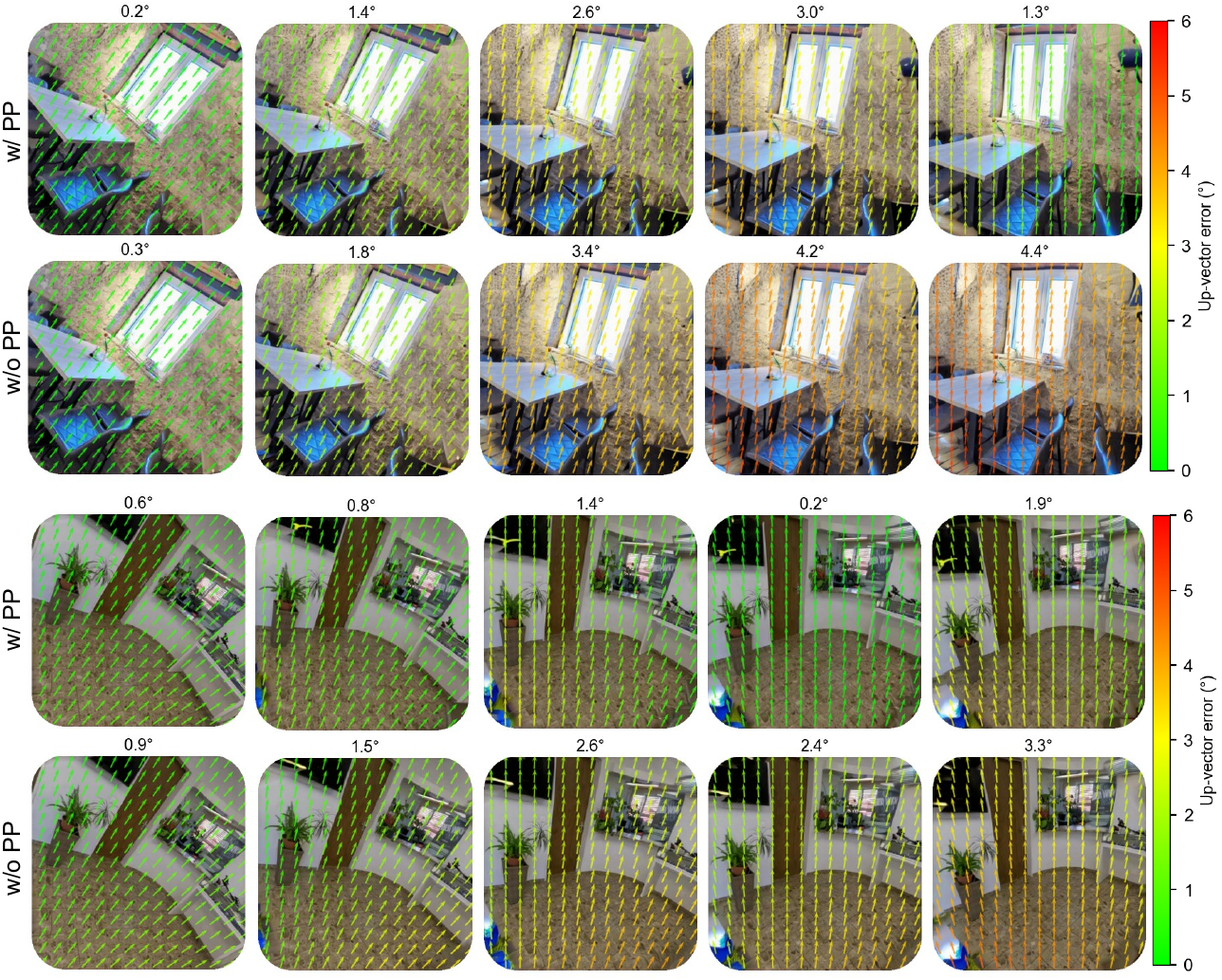}

    \caption{
    \textbf{Ablation study of the proposed physics propagation (PP) strategy.} For each sample, we show the generation results marked by the error up-vector and the error value with and without PP.
    }
    \label{fig:ablation_pp_vis}
\end{figure}

\subsection{More Results}
\subsubsection{Camera-to-World Understanding}
\begin{figure*}[!htbp]
    \centering

    \includegraphics[width=\linewidth]{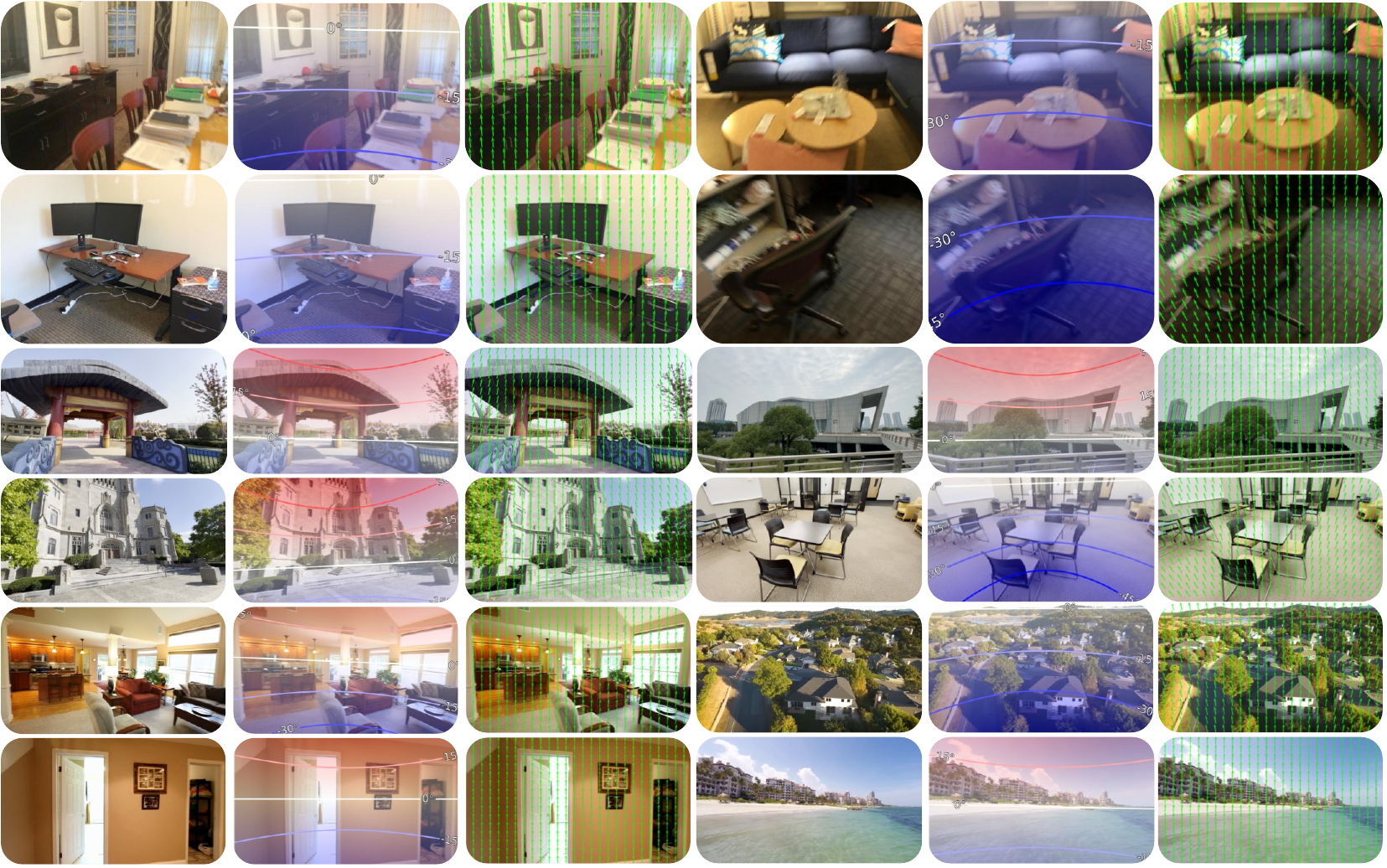}


    \includegraphics[width=\linewidth]{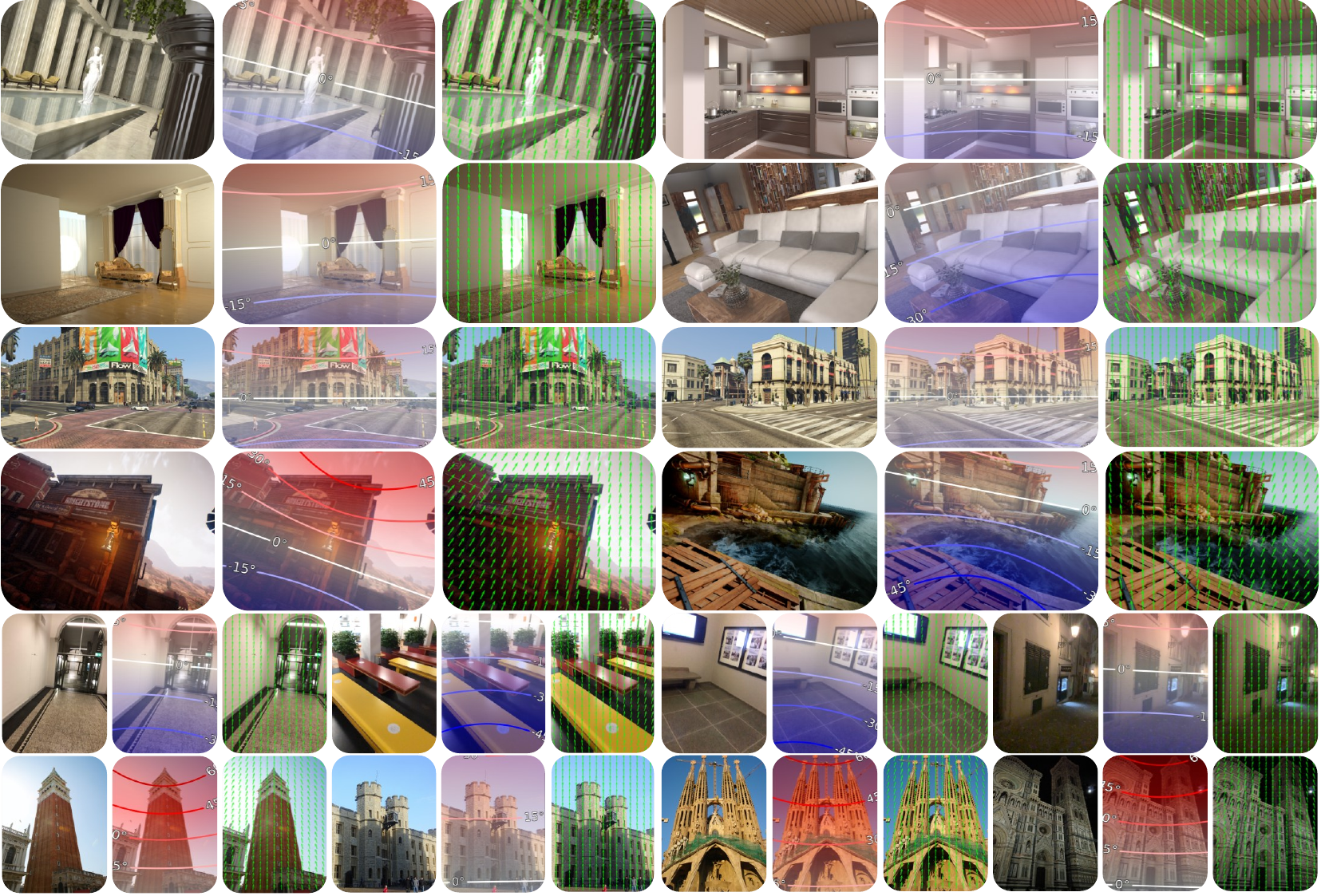}

    \caption{
    \textbf{Visualizations of the estimated physics world state (gravity field and latitude map) on different datasets~\cite{dai2017scannet, ling2024dl3dv, RealEstate10K, roberts2021hypersim, MVS-Synth, Tartanair, sarlin2022lamar, Megadepth}, in which the absolute camera parameters are predicted by Puffin-World.}
    }
    \label{fig:cam_vis_sp}
\end{figure*}

We provide additional qualitative results across diverse datasets~\cite{dai2017scannet, ling2024dl3dv, RealEstate10K, roberts2021hypersim, MVS-Synth, Tartanair, sarlin2022lamar, Megadepth} in Figure~\ref{fig:cam_vis_sp}. For each input image, we first estimate its absolute camera parameters using the understanding branch of Puffin-World, and then convert them into the corresponding latitude map and gravity field via Eq.~\ref{eq:perspective_field} for intuitive visualization. The results demonstrate that Puffin-World generalizes well across diverse scenes and remains robust under challenging camera configurations.

\subsubsection{Camera-Controllable Generation}
\begin{figure*}[!htbp]
    \centering

    \includegraphics[width=\linewidth]{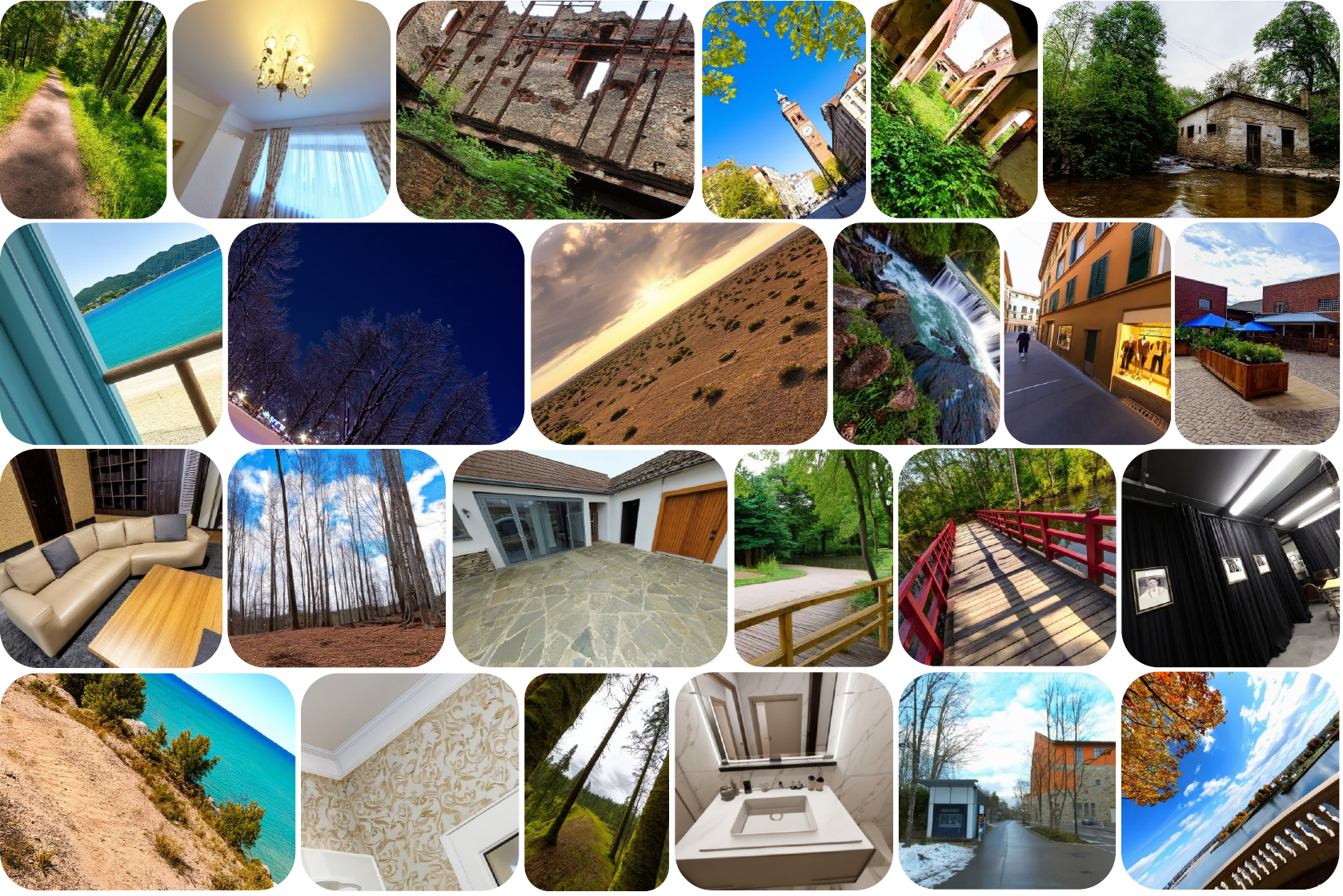}


    \includegraphics[width=\linewidth]{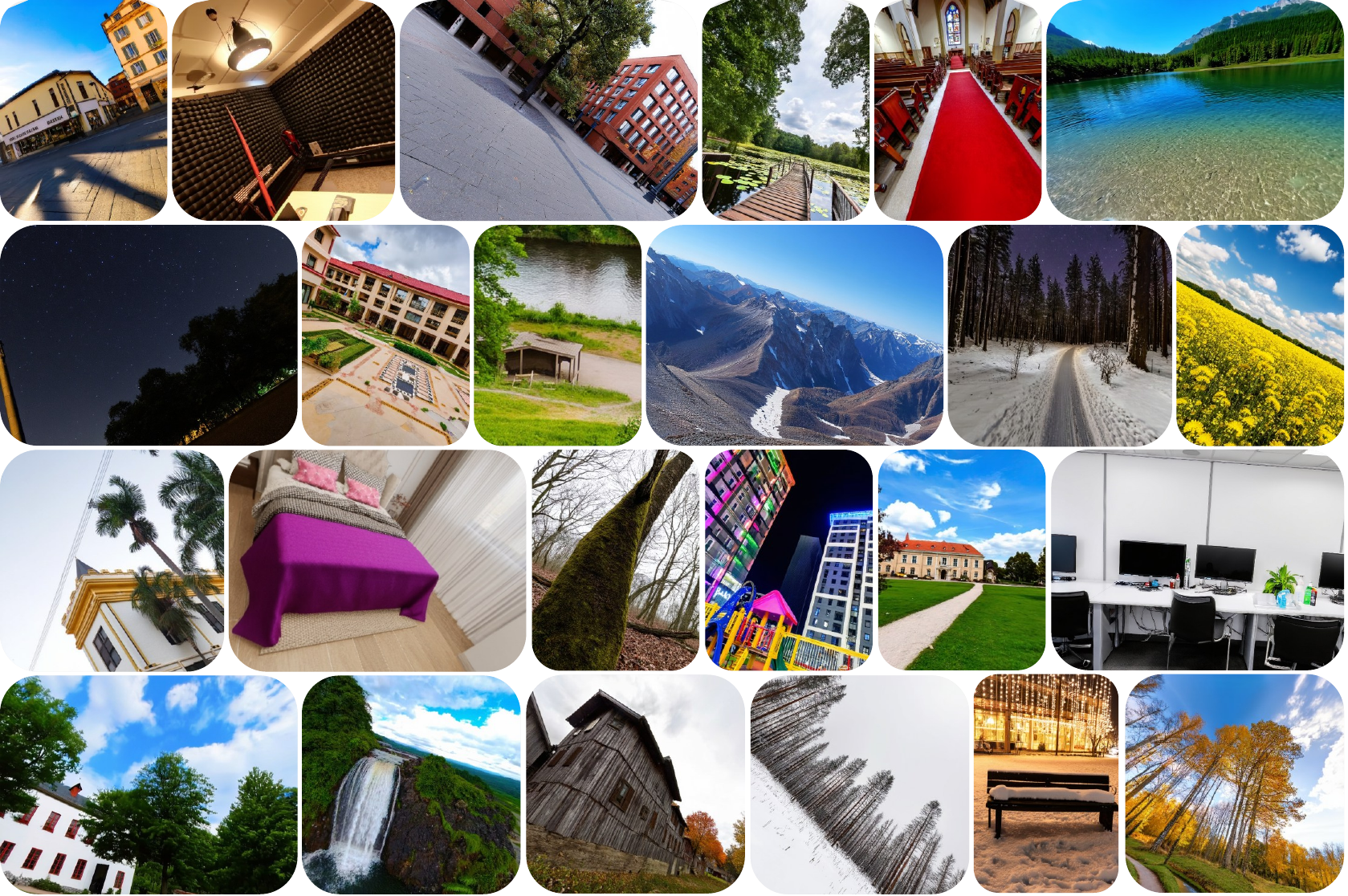}

    \caption{
    \textbf{Camera-controllable text-to-image generation results for free-viewpoint spatial simulation.}
    }
    \label{fig:cam_gen_sp}
\end{figure*}

Additional qualitative results are presented in Figure~\ref{fig:cam_gen_sp}, covering diverse spatial configurations across both indoor and outdoor environments. Puffin-World enables flexible free-viewpoint spatial simulation under a wide range of camera parameters, while preserving realistic appearance, coherent scene layouts, and semantic consistency with the input condition. In particular, the generated views faithfully follow the specified camera orientation and field-of-view, yielding noticeable yet physically plausible changes in perspective, horizon, and spatial composition. These results further demonstrate the robustness and generalization of Puffin-World for camera-controllable generation across diverse scenes and challenging viewpoints.

\subsubsection{3D World Modeling}
\begin{figure*}[!htbp]
    \centering

    \includegraphics[width=\linewidth]{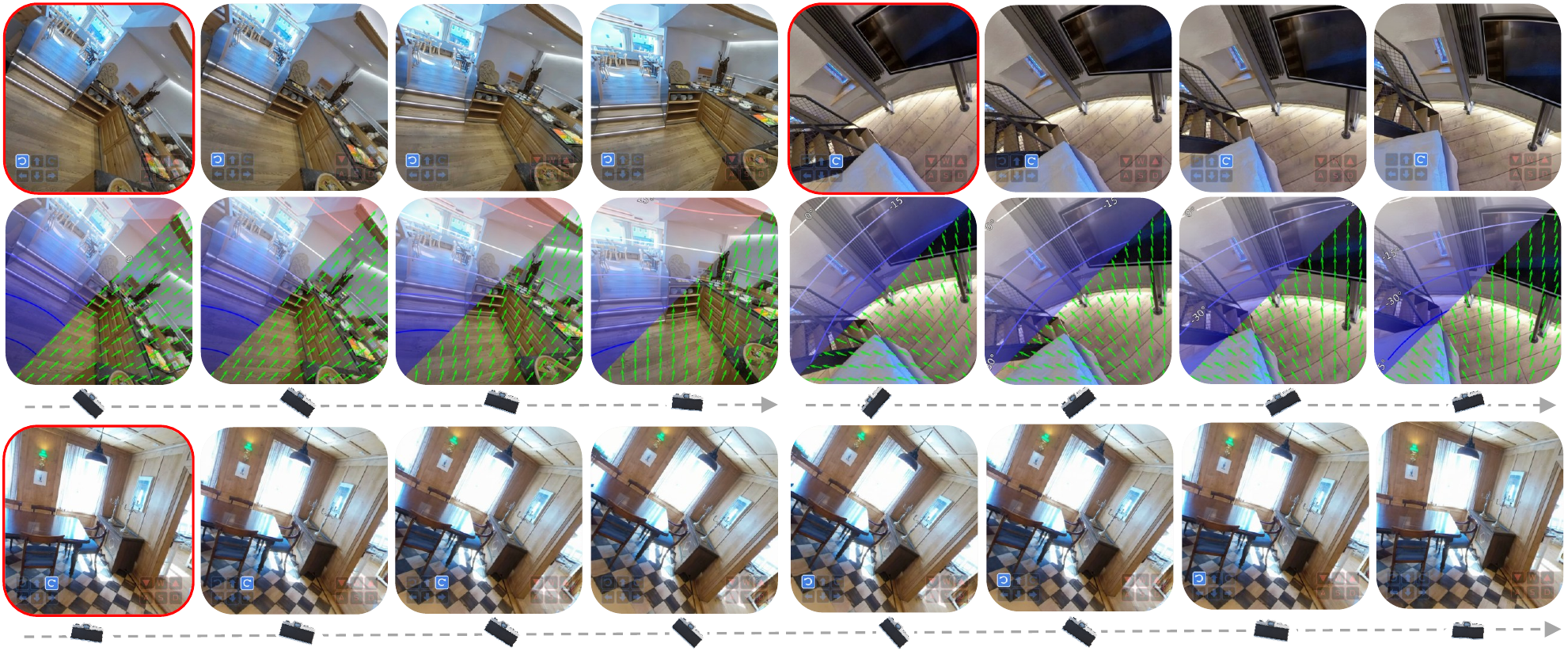}


    \includegraphics[width=\linewidth]{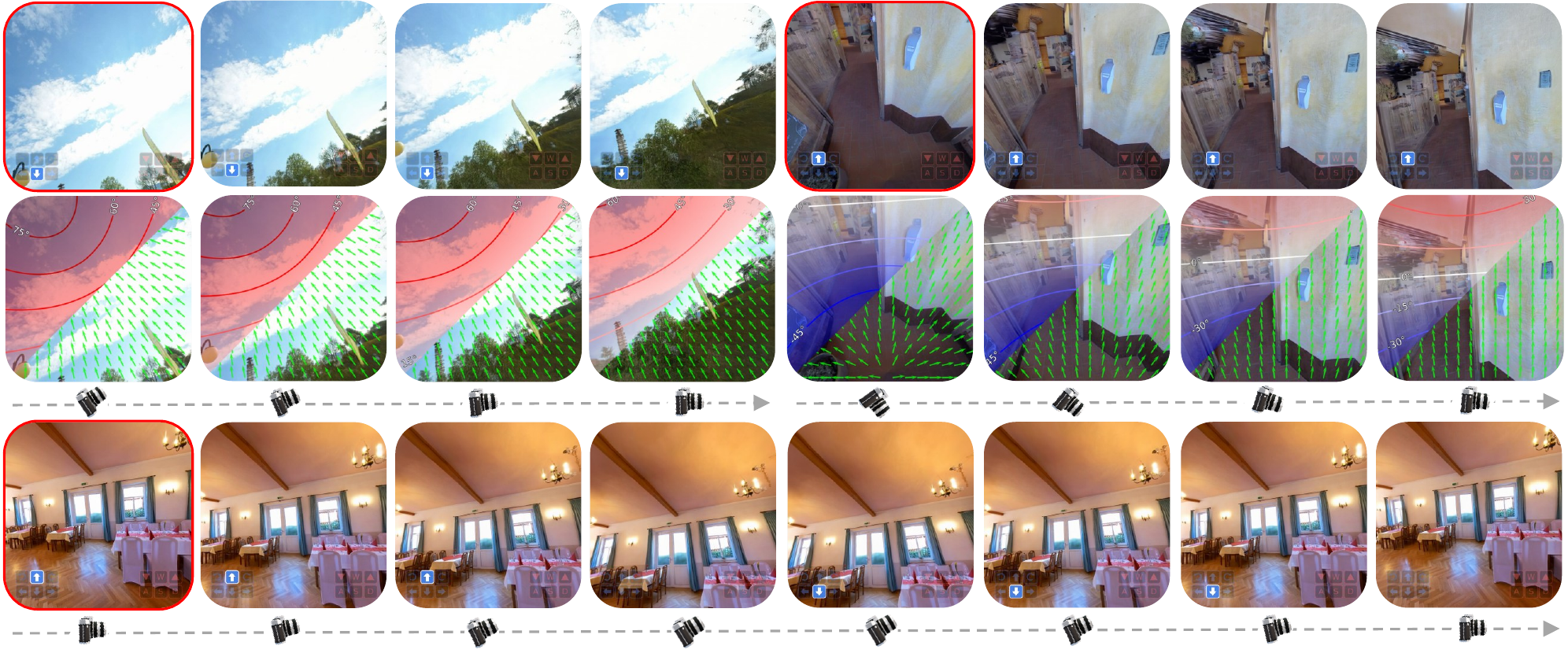}


    \includegraphics[width=\linewidth]{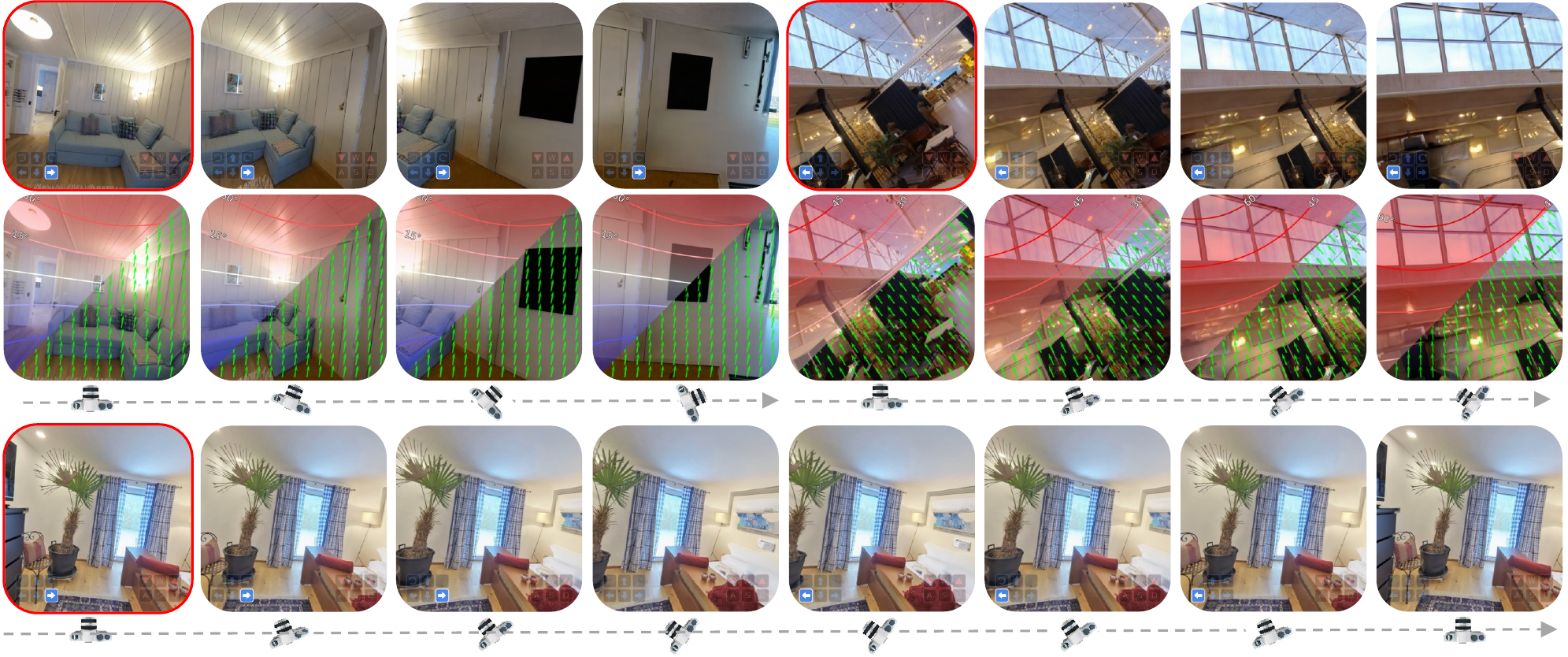}

    \caption{
    \textbf{3D world generation with diverse rotational controls (top: roll, middle: pitch, bottom: yaw).} For each control, we show the generated results from the single-pass path (with visualized physics world state) and recursive path.
    }
    \label{fig:rotation_traj_sp}
\end{figure*}

\begin{figure*}[!htbp]
    \centering

    \includegraphics[width=\linewidth]{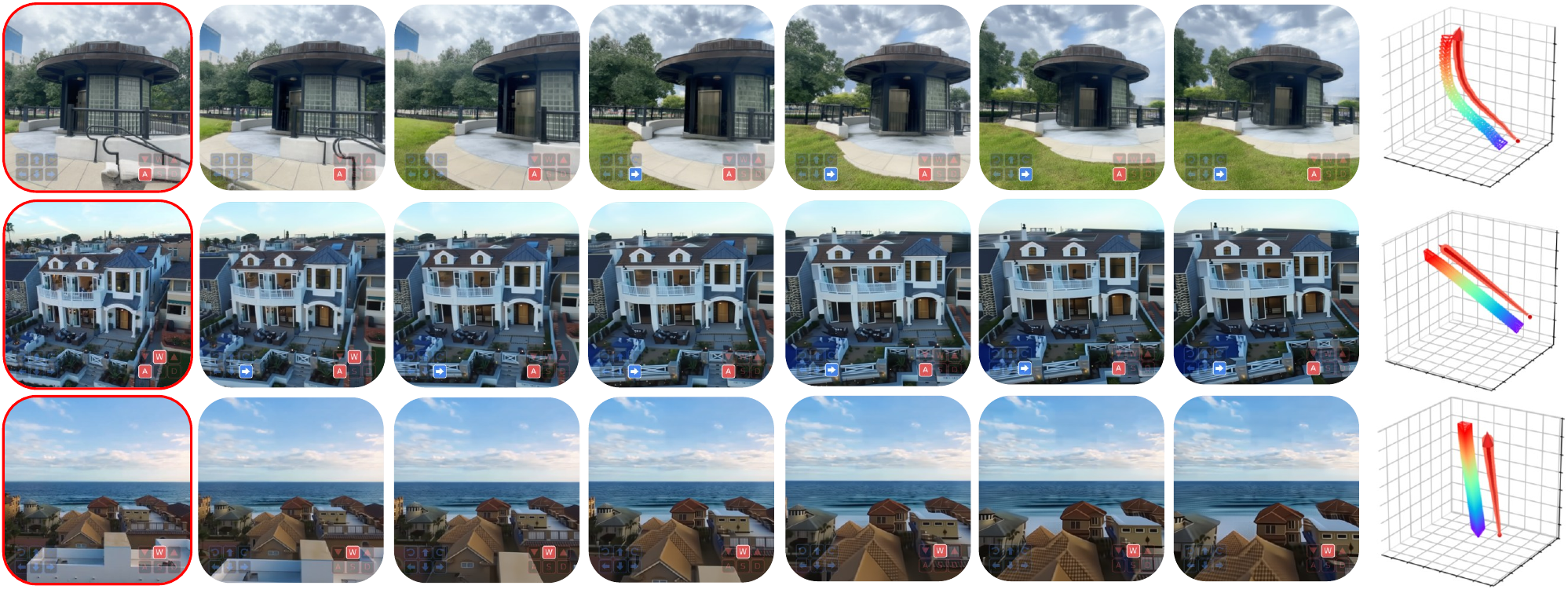}


    \includegraphics[width=\linewidth]{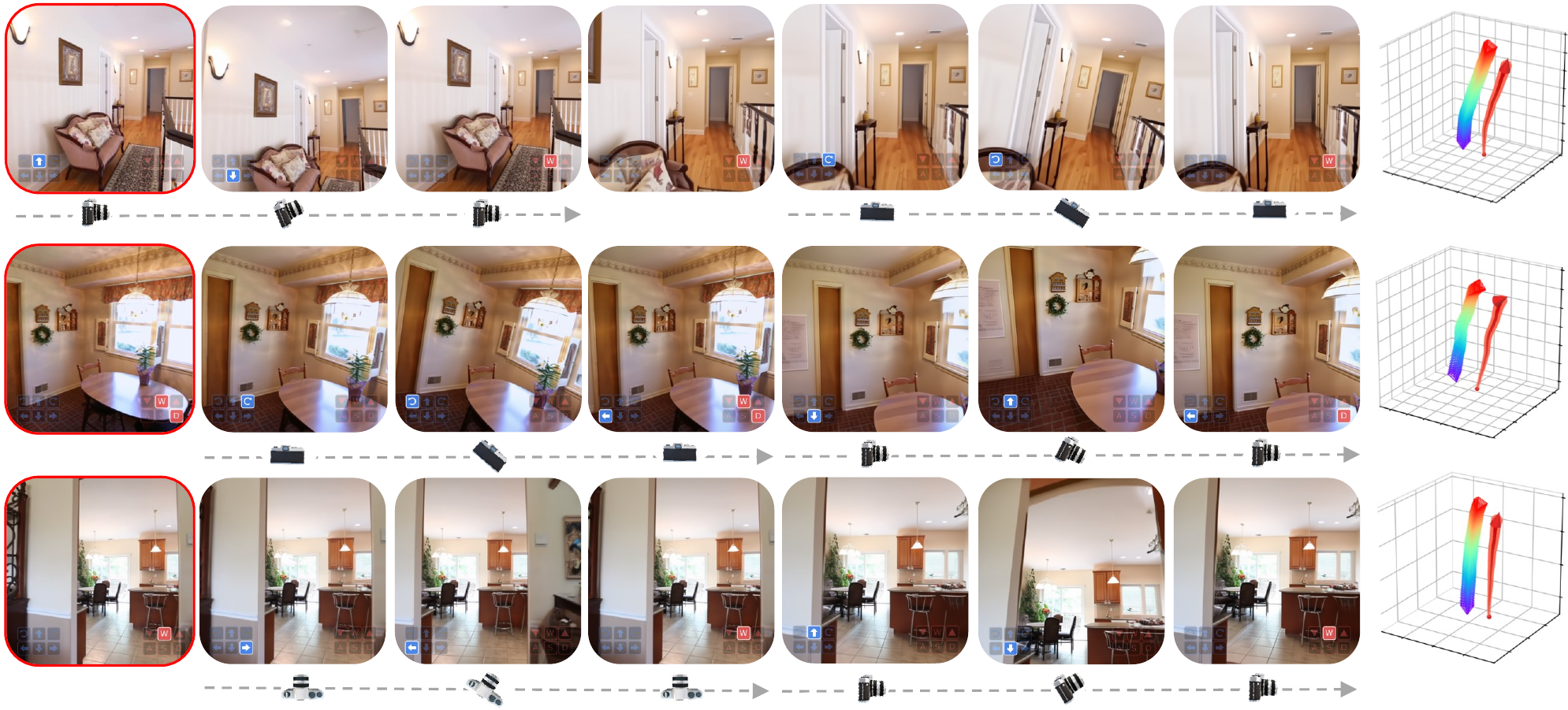}


    \includegraphics[width=\linewidth]{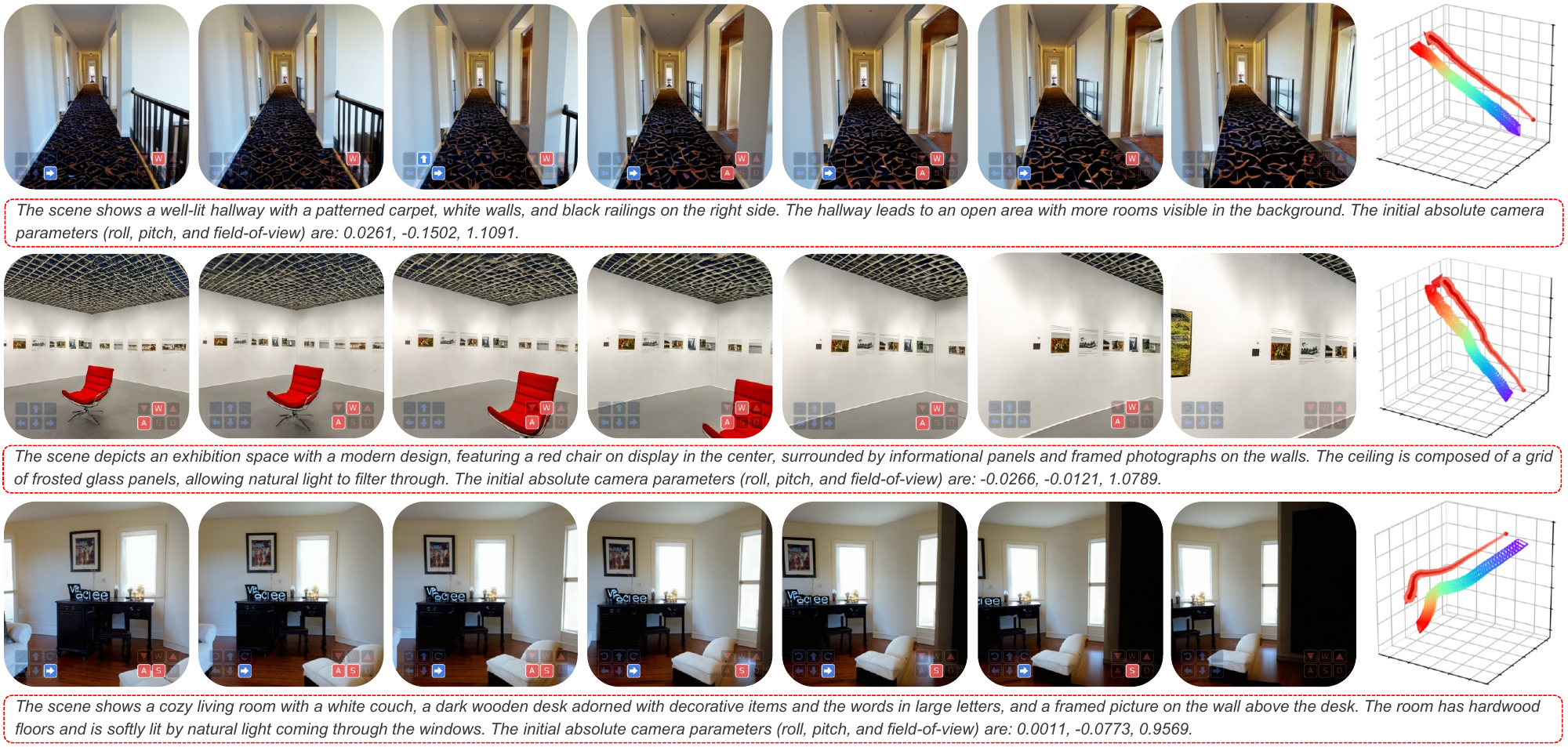}

    \caption{
    \textbf{3D world generation with diverse camera actions and conditions, including long trajectories, compound motions, and text-driven generation.}
    }
    \label{fig:3D_res_sp}
\end{figure*}

\begin{figure*}[!htbp]
    \centering

    \includegraphics[width=\linewidth]{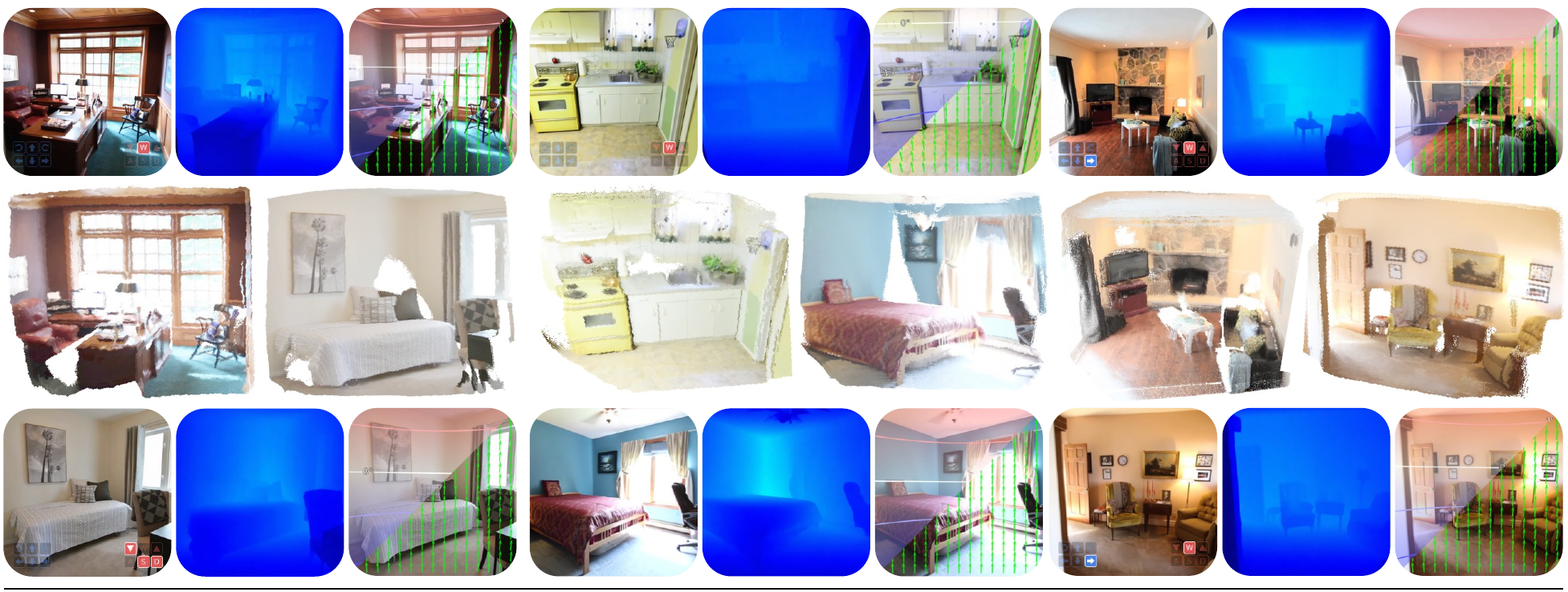}


    \includegraphics[width=\linewidth]{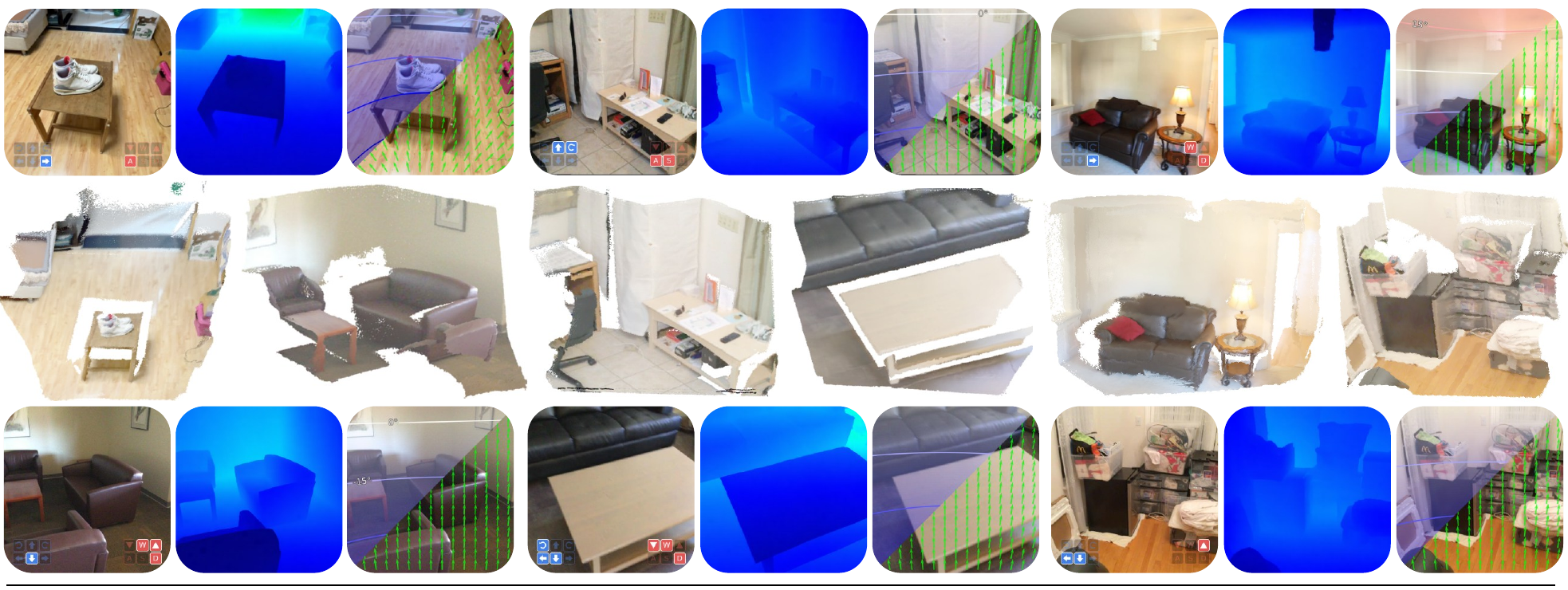}


    \includegraphics[width=\linewidth]{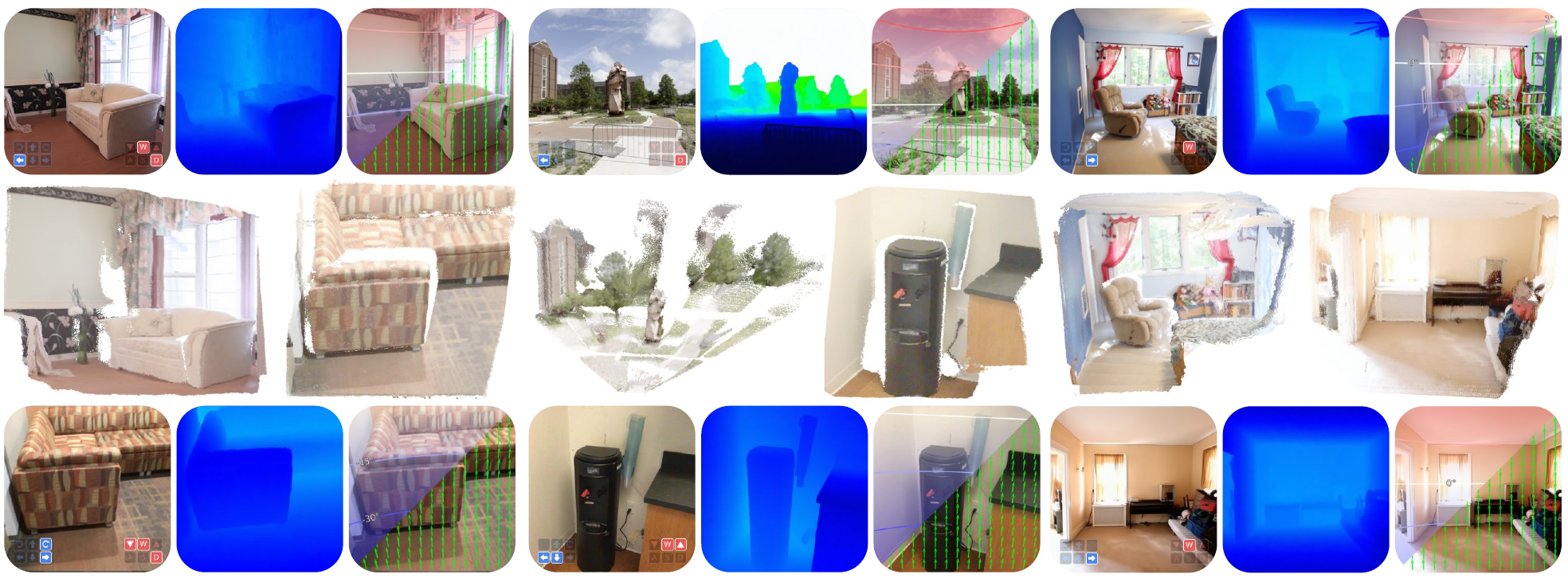}

    \caption{
    \textbf{Additional qualitative results of 3D world modeling.} For each combination, we show the generated native 3D world states: appearance (image), geometry (depth), and physics (gravity field and latitude map), and the 3D reconstructed result. Note that only one group of the generated world states is visualized in each sample for clarity.
    }
    \label{fig:3D_res_recon_sp}
\end{figure*}

As illustrated in Figures~\ref{fig:rotation_traj_sp}, \ref{fig:3D_res_sp}, and \ref{fig:3D_res_recon_sp}, we provide additional qualitative results of 3D world generation under diverse settings, including explicit rotational control, long-horizon trajectories, compound camera actions, text-driven generation, native 3D world-state prediction, and subsequent 3D reconstruction. Across these scenarios, Puffin-World consistently follows the prescribed camera motions while preserving semantic content and coherent spatial structure over extended viewpoint changes. The model remains robust under large rotations and complex combinations of translation and rotation, where maintaining both appearance consistency and physical plausibility is particularly challenging. Moreover, text-driven generation produces geometrically coherent multi-view worlds that remain faithful to the input prompt, while the predicted physics, geometry, and appearance states exhibit strong mutual consistency. The resulting 3D reconstructions further recover clear scene structure and stable spatial layouts, demonstrating that Puffin-World can consolidate its generated multi-view observations into a coherent 3D representation.

\newpage

\end{document}